\documentclass[10pt,twocolumn,letterpaper]{article}

\usepackage{cvpr}              

\usepackage{multirow}
\usepackage{graphicx}
\usepackage{booktabs}
\usepackage{subcaption}
\usepackage{makecell}
\usepackage{enumitem}
\usepackage{wrapfig}

\usepackage{color}
\usepackage{colortbl}
\definecolor{graybg}{gray}{0.9} 

\usepackage{wrapfig}  
\usepackage{caption}  
\usepackage{booktabs} 

\usepackage{caption}
\usepackage{graphicx}
\usepackage{pgfplots}
\usepackage{tikz}
\usetikzlibrary{patterns} 
\pgfplotsset{compat=1.17} 

\usepackage{titletoc}

\usepackage{tocloft}
\usepackage{xcolor}
\usepackage{multirow}

\usepackage{graphicx}
\usepackage{caption}
\usepackage{subcaption}

\makeatletter

\newlength{\appendixtopspace}
\newlength{\appendixtitlespace}
\newlength{\appendixsectionspace}
\newlength{\appendixsubsectionspace}

\newcommand{\appendixtocentry}[5]{%

    \ifnum#1=0

        \vspace{\appendixsectionspace}%

        \@dottedtocline{1}{0em}{2.8em}%
            {\hyperlink{#5}{\bfseries #2\quad #3}}%
            {\hyperlink{#5}{\bfseries #4}}%

    \else\ifnum#1=1

        \vspace{\appendixsubsectionspace}%

        \@dottedtocline{2}{1.8em}{3.8em}%
            {\hyperlink{#5}{#2\quad #3}}%
            {\hyperlink{#5}{#4}}%

    \else

        \vspace{0.15em}%

        \@dottedtocline{3}{3.5em}{4.5em}%
            {\hyperlink{#5}{#2\quad #3}}%
            {\hyperlink{#5}{#4}}%

    \fi\fi
}

\newcommand{\appsection}[1]{%

    \section{#1}%

    \hypertarget{appsec:\thesection}{}%

    \addtocontents{atoc}{%
        \protect\appendixtocentry
        {0}{\thesection}{#1}{\thepage}{appsec:\thesection}%
    }%
}

\newcommand{\appsubsection}[1]{%

    \subsection{#1}%

    \hypertarget{appsubsec:\thesubsection}{}%

    \addtocontents{atoc}{%
        \protect\appendixtocentry
        {1}{\thesubsection}{#1}{\thepage}{appsubsec:\thesubsection}%
    }%
}

\newcommand{\makeappendixcontents}{%

    \twocolumn[{%

        {\centering
            \Large
            \textbf{\@title}\par
            \vspace{0.5em}
            Supplementary Material\par
        }

        \vspace{1.0em}

        \begingroup

            \setlength{\parindent}{0pt}%
            \setlength{\parskip}{0pt}%

            \renewcommand{\baselinestretch}{1.15}%

            \normalsize

            \@starttoc{atoc}%

        \endgroup

        \vspace{0.5em}

    }]%

    \setcounter{page}{1}%

}

\makeatother

\definecolor{cvprblue}{rgb}{0.21,0.49,0.74}
\usepackage[pagebackref,breaklinks,colorlinks,allcolors=cvprblue]{hyperref}

\def\confName{CVPR}
\def\confYear{2027}

\title{From Predictive Geometry to Action: Learning Complementary Predictive Representations for Embodied Manipulation}

\author{
Xiaoxu Xu$^{1,2}$ \quad
Hao Li$^{2}$ \quad
Hongtao Zhang$^{2}$ \quad
Yili Tang$^{2}$ \quad
Jiafei Cao$^{2}$ \quad
Junhao Cai$^{2}$  \\
Haoxiang Ma$^{2,\dagger}$ \quad
Weixin Li$^{1}$ \quad
Weinan Zhang$^{2,\dagger}$
 \\
\\
$^{1}$Beihang University \qquad
$^{2}$Shanghai Artificial Intelligence Laboratory
}

\begin{document}
\maketitle

\begin{abstract}
Effective manipulation requires policies to capture not only current scene geometry but also its evolution under interaction. Existing approaches either augment policies with geometric information from current observations or use future prediction as auxiliary supervision, leaving how to represent predictive geometry for action generation largely unexplored. To address this, we propose a two-stage framework that learns complementary predictive representations from multi-view video clips. During pretraining, a frozen 3D foundation model provides dense geometric features, which are compressed into predictive geometric representations and optimized to preserve scene structure and temporal evolution. Action supervision is then applied to geometric transitions to extract a complementary motor representation focused on action-aware changes. During downstream post-training, predictive geometry is transferred via geometry queries, while the motor representation guides intermediate features along the action-generation pathway. Future observations serve only as privileged training supervision and are not required at inference. Experiments on LIBERO, LIBERO-Plus, SimplerEnv, and real-world bimanual and single-arm tasks demonstrate consistent improvements, including a 10.4-percentage-point absolute gain on the WidowX evaluation. We further validate transfer across multiple VLA architectures and WAMs, improving both future-world and action prediction.
\end{abstract}    
\section{Introduction}
\label{sec:introduction}


\begin{figure}[!t]
\centering
\includegraphics[width=1.0\linewidth]{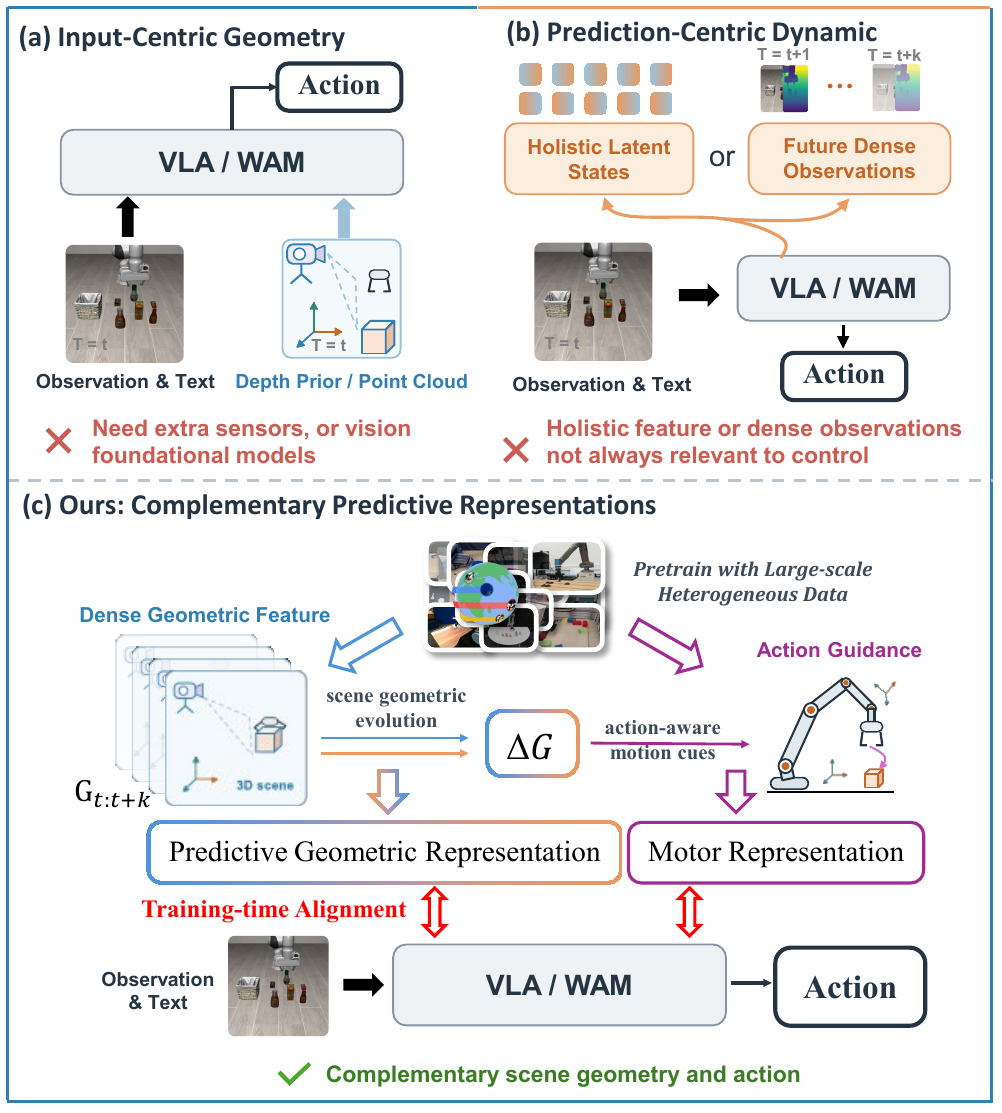}
\caption{\small Comparison of geometry-guidance paradigms for policy learning.
(a) Input-centric methods augment policies with spatial information from current observations.
(b) Prediction-centric methods use future-world predictions as auxiliary supervision.
(c) Our approach learns predictive geometric representations from dense geometric features and derives complementary motor representations from scene geometry evolution. Both are transferred through role-specific guidance pathways during training and are not required at inference time.}
\vspace{-0.35cm}
\label{introduction}
\end{figure}


Anticipating how the physical world evolves under interaction\citep{assran2025v,assran2023self} is increasingly important for vision-language-action (VLA) models\citep{ma2026internvla,luo2026being} and world action models (WAM)\citep{li2026causal,ye2026world}. This capability enables agents to reason about interaction consequences rather than merely mapping current observations to immediate actions. However, effective manipulation requires not only temporal foresight but also precise spatial understanding. Geometry describes object layouts, spatial relationships, and robot-scene configurations, while dynamics characterize how these structures evolve during interaction. These aspects play complementary roles in action generation: geometry establishes the spatial structure for executing actions, while its evolution provides cues about future scene changes. This raises a central question: how should geometric structure and its predicted evolution be represented to better support action generation?


Recent efforts to incorporate geometry into predictive policy learning follow two directions. As shown in Fig.~\ref{introduction}, input-centric approaches\citep{sun2025geovla,spatialvla,li2026wam4d,zhang2025spatial} augment observations with depth, point clouds or pretrained geometric features. While improving spatial grounding, these methods primarily use geometry as additional context for the currently observed scene and may depend on dedicated sensing or geometry-estimation pipelines. In contrast, prediction-centric approaches\citep{yang20264d,he2026elan4d,li2025spatial,li2025qdepth,su2026world} supervise models with future RGB-D observations, geometric states, or other world-prediction targets, explicitly exposing policies to future scene structures. However, these targets are typically represented as dense observations or holistic latent states, requiring representations to preserve geometric information whose relevance to control may vary substantially. Consequently, existing approaches primarily organize geometry to describe the scene and its evolution, leaving less explicit how action-aware information should be distinguished from that required for coherent world representation. This creates a gap between predictive geometric modeling and action-oriented motor representation learning.

A natural way to bridge this gap is to recognize that predictive geometry and action generation place different preferences on learned representations. Predictive geometric modeling benefits from preserving a sufficiently complete and coherent account of scene structure. Action generation, in contrast, benefits from emphasizing geometric variations that are informative for motor commands while reducing the influence of static or action-irrelevant scene content. Rather than forcing these roles to be served in exactly the same representation, we introduce two complementary representations with distinct but related purposes: a \emph{predictive geometric representation} that provides a compact and structured description of scene geometry and its evolution, and a \emph{motor representation} that emphasizes action-aware information from geometric transitions. The former provides structured predictive context, while the latter focuses this predictive information toward control.

Building on this insight, we propose CPR, a two-stage framework that learns complementary predictive representations from large-scale heterogeneous interaction videos\citep{open_x_embodiment,jin2026umi,Libero,goyal2017something,chen2025robotwin} and transfers them to downstream policies. As illustrated in Fig.~\ref{introduction}, CPR first constructs predictive geometric representations from dense geometric features and then extracts action-aware information from their temporal evolution. Specifically, as shown in Fig.~\ref{fig2}, during representation pretraining, a frozen 3D foundation model\citep{wang2025vggt} extracts dense geometric features from multi-view video observations. A hierarchical query-based encoder progressively compresses these features into compact temporal states. Geometry reconstruction, masked representation prediction, and geometric transition prediction jointly encourage the predictive geometric representation to preserve scene structure and capture its temporal evolution. We further apply action supervision to transitions between consecutive predictive geometric states, extracting a motor representation that emphasizes action-aware geometric transitions. Thus, the two representations capture complementary aspects of scene evolution: structured geometric information and control-relevant transitions. During post-training, as shown in Fig.~\ref{fig3}, they are transferred through role-specific supervision pathways. Dedicated geometry queries learn to infer predictive geometric representations from current-time inputs, while intermediate features along the action-generation pathway are aligned with the motor representation. Importantly, future observations and the pretrained representation model are used only to construct privileged supervision during training and are not required at inference. We evaluate CPR across multiple VLA architectures and WAMs on diverse simulated and real-world manipulation benchmarks, demonstrating consistent improvements across robot embodiments and action-generation paradigms. Our contributions are summarized as follows:





\begin{itemize}[leftmargin=1.2em, itemindent=0pt]
\item We identify complementary roles for predictive geometric and motor representations in predictive policy learning: preserving structured scene geometry and emphasizing control-relevant geometric transitions, respectively.

\item We develop a two-stage framework to learn and transfer these complementary representations. Hierarchical geometric tokenization and predictive objectives capture evolving geometric states, while action supervision extracts control-relevant information from their transitions. The resulting representations are transferred through geometry-query and motor-alignment pathways.

\item Extensive experiments on LIBERO, LIBERO-Plus, SimplerEnv, and real-world bimanual and single-arm manipulation tasks demonstrate consistent improvements across policy architectures and robot embodiments. CPR achieves a 10.4-percentage-point absolute gain on the WidowX evaluation and improves both future-world and action prediction in WAM.
\end{itemize}

\section{Related Work}

\noindent\textbf{Predictive Representations for VLA and WAM.}
Predictive modeling provides VLA models and WAMs with foresight beyond the current observation. Existing approaches introduce future information either through observation prediction or compact predictive representations. Observation-level methods predict future RGB observations, videos, or world states to support action generation~\citep{zhang2025dreamvla,lv2025f1,yuan2026fastwam,agarwal2026cosmos}, while latent-space approaches compress future information into compact conditions or visuomotor representations~\citep{xu2026futurevla,su2026world,sun2026vla}. Recent methods~\citep{lee2026mu_0,yan2026robust,lee2026act,wang2026robointer1,li2025cronusvla,chen2025internvla,instructvla,ding2026skillplug}, also explore structured predictive signals such as dynamic, spatial, or semantic information to connect VLMs or world modeling with control. However, existing methods vary in how geometric structure and action-relevant temporal transitions are organized. Our work focuses on predictive geometry: we learn a structured representation that preserves scene geometry and its evolution, and derive a complementary motor representation from geometric transitions.

\noindent\textbf{Geometric Learning for Robotic Manipulation.}
Geometric information has been increasingly incorporated into robotic policies to improve spatial grounding. Input-centric approaches \citep{spatialvla,sun2025geovla,bhat20253d,zhang2025spatial,huang2025roboground,li2026robointer,yang2026robo3r} augment current observations with depth, point clouds, 2D or 3D coordinates, or spatial features, providing direct access to the geometry of the observed scene. More recent prediction-centric approaches \citep{li2025qdepth,he2026elan4d,yang20264d,su2026world} instead use geometric or spatiotemporal prediction as auxiliary supervision, including depth prediction, robot trajectories, and 3D motion representations. These approaches demonstrate the value of geometry for predictive policy learning, but geometric structure and action relevance may favor different types of information: preserving scene geometry requires broad structural fidelity, whereas action generation benefits from transitions that are informative for motor behavior. Our work addresses this distinction by learning complementary predictive geometric and motor representations, and transferring them separately to downstream policies through dedicated interfaces.
\section{Method}
\label{sec:method}

\begin{figure*}[!t]
\centering
\includegraphics[width=1.0\linewidth]{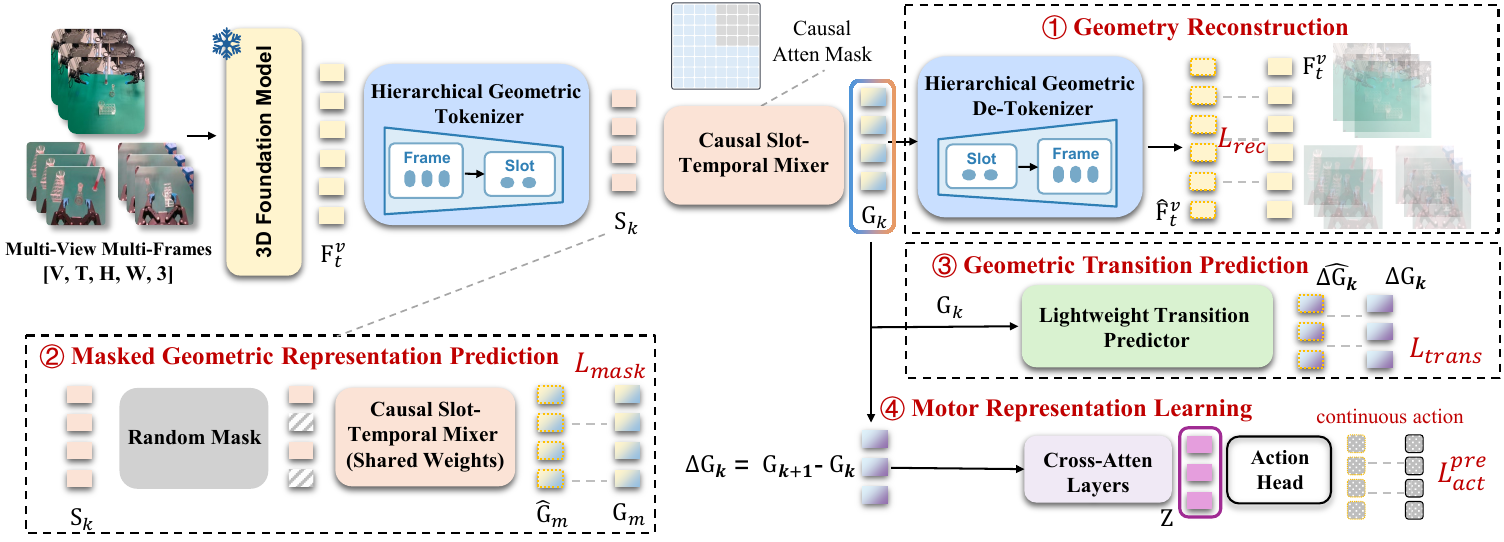}
\caption{\small Overview of complementary predictive representation pretraining. A frozen 3D foundation model extracts dense geometric features from multi-view video observations, which are compressed by the Hierarchical Geometric Tokenizer and temporally modeled by a causal Slot-Temporal Mixer to produce the predictive geometric representation. Geometry reconstruction, masked geometric representation prediction, and geometric transition prediction jointly shape this representation to preserve scene structure and capture its temporal evolution. The resulting geometric transitions are used with action supervision to extract an action-aware motor representation, providing complementary predictive and motor information for policy learning.
}
\label{fig2}
\vspace{-0.45cm}
\end{figure*}

\begin{figure}[!t]
\centering
\includegraphics[width=1.0\linewidth]{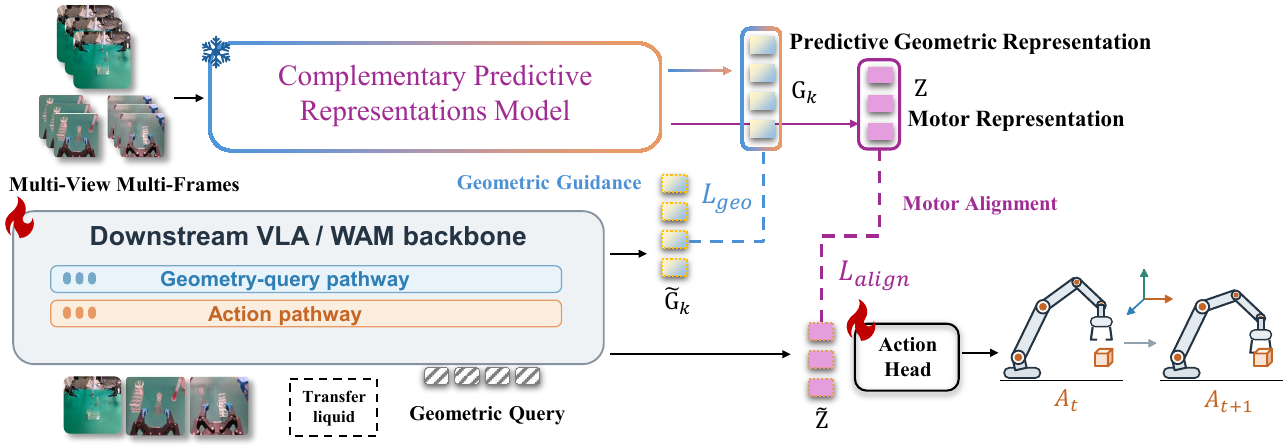}
\caption{\small Overview of representation-guided policy post-training. The frozen pretrained model provides the predictive geometric representation $\mathbf{G}$ and motor representation $\mathbf{Z}$ as privileged supervision. The two representations are transferred through role-specific pathways: $\mathbf{G}$ supervises dedicated geometry queries, while $\mathbf{Z}$ aligns with intermediate features along the action-generation pathway.}
\label{fig3}
\vspace{-0.35cm}
\end{figure}

We present a two-stage framework for learning complementary predictive representations and transferring them to downstream policies. As illustrated in Fig.~\ref{fig2}, during \emph{(i) Complementary Predictive Representation Pretraining}, continuous multi-view video clips are first converted into compact predictive geometric representations that preserve the spatial structure of the scene while capturing its temporal evolution. We then use the transitions between these predictive geometric representations, together with temporally aligned action supervision, to extract action-aware information into a complementary motor representation. During \emph{(ii) Representation-Guided Policy Post-training}, as shown in Fig.\ref{fig3}, the predictive geometric representations and motor representations are transferred to the policy through two dedicated supervision pathways. Geometry queries are trained to infer structured predictive geometric information from current-time inputs, while intermediate features along the action-generation pathway are aligned with the motor representation.

\subsection{Complementary Predictive Representation Pretraining}
\label{sec:pretraining}
\noindent\textbf{Hierarchical Geometric Tokenization.}
\label{sec:hierarchical_encoding}
As shown in Fig.\ref{fig2}, we first transform dense multi-view observations into a compact and temporally organized sequence of geometric states. A frozen pretrained 3D foundation model~\citep{wang2025vggt} is applied independently to each frame-view pair $O_t^v$, where $t\in T$ and $v\in V$ denote the number of frames and camera views, respectively, producing dense geometric features $\mathbf{F}_t^v$. These features provide rich spatial information, but directly modeling all geometric tokens across views and time incurs substantial computational overhead. We therefore introduce a hierarchical query-based encoder that progressively compresses the representation along the view and temporal dimensions.

At each time step $t$, geometric features from all camera views are concatenated along the token dimension and augmented with learnable view embeddings to preserve view-specific information. A set of learnable geometric queries then cross-attends to the resulting multi-view tokens, producing a compact frame-level geometric state $\mathbf{H}_t$. We subsequently aggregate the frame-level states along the temporal dimension. Inspired by \citep{wan2025wan}, the first state $\mathbf{H}_0$ is retained without temporal aggregation and serves as a temporal anchor, while each complete group of four subsequent frame-level states is compressed into one temporal slot. This produces $K$ temporal slots, where $K = 1+\left\lfloor\frac{T-1}{4}\right\rfloor$. Temporal positional embeddings are introduced before temporal aggregation to preserve the ordering of the underlying frames. We denote the tokenized geometric sequence as $\mathbf{S}=\{\mathbf{S}_k\}_{k=1}^{K}$. This hierarchical design substantially reduces the spatiotemporal token budget while maintaining an organized representation of scene geometry over time.

\noindent\textbf{Predictive Geometric Representation Learning.}
\label{sec}
Although the tokenized geometric sequence $\mathbf{S}$ provides compact descriptions of individual temporal segments, each state alone does not explicitly encode how the interaction evolves over time. As shown in Fig.\ref{fig2}, we therefore process the sequence using a causal Slot-Temporal Mixer, implemented as a stack of causal attention blocks. At temporal slot $k$, the representation attends only to states at or before $k$, such that each output summarizes the geometric history available up to that point. The resulting sequence forms the predictive geometric representations $\mathbf{G}=\{\mathbf{G}_k\}_{k=1}^{K}$.

We shape $\mathbf{G}$ with three complementary objectives, each emphasizing a different property of predictive geometric information. Geometry reconstruction preserves the spatial information inherited from the frozen 3D foundation model; masked geometric representation prediction encourages temporal consistency and contextual inference across geometric states; and geometric transition prediction explicitly models local transitions between consecutive states. Together, these objectives encourage $\mathbf{G}$ to remain structurally informative while capturing predictable patterns in the evolution of scene geometry.

\textbf{1) Geometry Reconstruction.}
Hierarchical tokenization substantially compresses the dense geometric features and may consequently discard information about the underlying scene structure. To preserve such information, we attach a reconstruction decoder to the compact geometric pathway. The decoder mirrors the encoding hierarchy by first expanding the temporal slots into frame-level representations and subsequently reconstructing the geometric features associated with each camera view, yielding $\widehat{\mathbf{F}}_t^v$. Let $\Omega$ denote the set of valid frame-view pairs. We optimize
\begin{equation}
\mathcal{L}_{\mathrm{rec}}
=
\frac{1}{|\Omega|}
\sum_{(t,v)\in\Omega}
\operatorname{SmoothL1}
\left(
\widehat{\mathbf{F}}_t^v,
\operatorname{sg}(\mathbf{F}_t^v)
\right),
\label{eq:geometry_reconstruction_loss}
\end{equation}
where $\operatorname{sg}(\cdot)$ denotes the stop-gradient operator. By reconstructing features from the frozen 3D encoder rather than raw observations, this objective encourages the compact representation to retain geometrically informative scene structure despite the substantial token compression.

\textbf{2) Masked Geometric Representation Prediction.}
To further capture temporal dependencies among geometric states, we introduce a slot-level masked prediction objective that requires missing geometric states to be inferred from the available temporal context. Specifically, we randomly sample a subset of temporal slots $\mathcal{M}\subseteq\{2,\ldots,K\}$
for masking. A full-sequence target branch processes all temporal slots through the Slot-Temporal Mixer to produce the target representation $\mathbf{G}$. In parallel, a context branch processes the masked temporal slots using the same weight-shared Slot-Temporal Mixer. A lightweight predictor then uses the resulting contextual representations to estimate the target states at the masked positions, producing $\widehat{\mathbf{G}}_m$ for each $m\in\mathcal{M}$. The objective is
\begin{equation}
\mathcal{L}_{\mathrm{mask}}
=
\frac{1}{|\mathcal{M}|}
\sum_{m\in\mathcal{M}}
\operatorname{SmoothL1}
\left(
\widehat{\mathbf{G}}_m,
\operatorname{sg}(\mathbf{G}_m)
\right)
+
\lambda_{\mathrm{vc}}\mathcal{R}_{\mathrm{vc}}(\mathbf{G}),
\label{eq:masked_state_prediction}
\end{equation}
where $\mathcal{R}_{\mathrm{vc}}$ denotes a variance-covariance regularizer. Its variance term maintains sufficient dispersion across feature dimensions, while its covariance term discourages redundant correlations between dimensions. Together, they prevent degenerate latent solutions and promote an informative predictive state space.

\textbf{3) Geometric Transition Prediction.}
Masked representation prediction encourages contextual inference among temporal states, but does not directly constrain the direction in which the latent geometry evolves. We therefore explicitly model local geometric transitions. For two consecutive predictive states, we define the latent transition as
\begin{equation}
\Delta\mathbf{G}_k
=
\mathbf{G}_{k+1}-\mathbf{G}_k,
\qquad
k=1,\ldots,K-1,
\label{eq:geometric_transition}
\end{equation}
and denote the transition sequence by
$\Delta\mathbf{G}=\{\Delta\mathbf{G}_k\}_{k=1}^{K-1}$. A lightweight transition predictor takes the current predictive state $\mathbf{G}_k$ and estimates its subsequent latent change $\widehat{\Delta\mathbf{G}}_k$. We optimize
\begin{equation}
\mathcal{L}_{\mathrm{trans}}
=
\frac{1}{K-1}
\sum_{k=1}^{K-1}
\operatorname{SmoothL1}
\left(
\widehat{\Delta\mathbf{G}}_k,
\operatorname{sg}(\Delta\mathbf{G}_k)
\right).
\label{eq:geometric_transition_loss}
\end{equation}
Predicting the residual transition rather than reconstructing another complete geometric state focuses this objective on how the latent scene configuration changes over time. Together with geometry reconstruction and masked representation prediction, transition prediction encourages $\mathbf{G}$ to capture both the spatial structure of the scene and its predictable temporal evolution.

\noindent\textbf{Motor Representation Learning.}
\label{sec:action_relevant_transition}
The predictive geometric representation is designed to retain a broad and structured account of scene geometry and its evolution. While such information is useful for predictive modeling, not every geometric variation is equally informative for control. In particular, $\Delta\mathbf{G}$ may contain transitions induced by robot motion and object interaction together with variations that are only weakly related to the corresponding motor commands. We therefore use the geometric transitions as a motion-centric source and introduce action supervision to emphasize the component of these transitions that is informative for action generation.  Specifically, a set of learnable motor queries cross-attends to the transition tokens in $\Delta\mathbf{G}$, which serve as keys and values. The resulting features form the motor representation $\mathbf{Z}$. An action prediction head maps $\mathbf{Z}$ to a predicted action chunk $\widehat{\mathbf{A}}$, which is supervised by the ground-truth action chunk $\mathbf{A}$:
\begin{equation}
\mathcal{L}_{\mathrm{act}}^{\mathrm{pre}}
=
\left\|
\widehat{\mathbf{A}}-\mathbf{A}
\right\|_1.
\label{eq:pretraining_action_loss}
\end{equation}

The complete representation-pretraining objective is
\begin{equation}
\mathcal{L}_{\mathrm{pre}}
=
\mathcal{L}_{\mathrm{rec}}
+
\lambda_{\mathrm{mask}}\mathcal{L}_{\mathrm{mask}}
+
\lambda_{\mathrm{trans}}\mathcal{L}_{\mathrm{trans}}
+\mathcal{L}_{\mathrm{act}}^{\mathrm{pre}},
\label{eq:pretraining_objective}
\end{equation}
where $\lambda_{\mathrm{mask}}$ and
$\lambda_{\mathrm{trans}}$
balance the corresponding objectives.

\subsection{Representation-Guided Policy Post-training}
\label{sec:posttraining}
We next transfer the complementary information encoded by the predictive geometric and motor representations to a downstream policy. As shown in Fig.\ref{fig3}, for each post-training example, the frozen pretrained model processes a multi-view training clip containing the current and future observations and extracts the target predictive geometric representation $\mathbf{G}$ and motor representation $\mathbf{Z}$. These future-informed representations serve exclusively as privileged supervision, while the downstream policy continues to receive only its standard current-time inputs. Let $O_t$ denote the current observation, $l$ the language instruction, and $s_t$ the robot state. The policy backbone processes $(O_t,l,s_t)$ together with a set of learnable geometry queries. In addition to the intermediate features used by the policy for action generation, the geometry queries produce a sequence of predicted geometric representation $\widetilde{\mathbf{G}}
= \{\widetilde{\mathbf{G}}_k\}_{k=1}^{K}$. We align these representations with the frozen targets $\mathbf{G}$ using
\begin{equation}
\mathcal{L}_{\mathrm{geo}}
=
\frac{1}{K}
\sum_{k=1}^{K}
\operatorname{SmoothL1}
\left(
\widetilde{\mathbf{G}}_k,
\operatorname{sg}(\mathbf{G}_k)
\right),
\label{eq:posttraining_geometry_loss}
\end{equation}
which encourages the model to capture the current scene configuration together with its anticipated geometric evolution. The motor representation is transferred through a different interface, rather than requiring the policy to explicitly recover $\mathbf{Z}$ through an additional prediction output, we use $\mathbf{Z}$ as a guidance target for intermediate features along the action-generation pathway. Let $\widetilde{\mathbf{Z}}$ denote the intermediate policy features, we align these features with the frozen motor representation target using a cosine-distance objective:
\begin{equation}
\mathcal{L}_{\mathrm{align}}
=
1-
\cos
\left(
\widetilde{\mathbf{Z}},
\operatorname{sg}(\mathbf{Z})
\right).
\label{eq:posttraining_action_alignment_loss}
\end{equation}


Finally, we combine the two representation-guidance objectives with the native action-generation objective of the downstream policy:
\begin{equation}
\mathcal{L}_{\mathrm{post}}
=
\mathcal{L}_{\mathrm{act}}^{\mathrm{post}}
+
\lambda_{\mathrm{geo}}\mathcal{L}_{\mathrm{geo}}
+
\lambda_{\mathrm{align}}\mathcal{L}_{\mathrm{align}},
\label{eq:posttraining_total_objective}
\end{equation}
where $\mathcal{L}_{\mathrm{act}}^{\mathrm{post}}$ denotes the action-generation objective of the policy backbone. The representation guidance therefore does not replace the backbone's native action modeling objective, but complements it with predictive geometric and motor supervision. At inference time, neither future observations nor the pretrained representation model is required.

\section{Experiment}
We design our experiments to answer three questions:
\begin{enumerate}[label=\textbf{(Q\arabic*)}, leftmargin=*]
    \item Does CPR guidance improve downstream manipulation performance across diverse settings?
    \item Does CPR generalize across policy paradigms while improving predictive modeling?
    \item How should CPR be organized and transferred for effective policy learning?
\end{enumerate}

\noindent\textbf{Experimental Setup.}
We evaluate CPR guidance across simulated and real-world manipulation settings, covering task suites, robot embodiments, and downstream policies. Unless otherwise specified, we use QwenOFT~\citep{yang2025qwen3,openvla-oft} as the downstream policy baseline and denote this baseline as \emph{Ours w/o CPR}. Our evaluation includes LIBERO~\citep{Libero}, LIBERO-Plus~\citep{fei2025libero}, SimplerEnv~\citep{simpleenv}, and four real-world manipulation tasks. Our real-world evaluation consists of three bimanual tasks, \textit{remove a stopper}, \textit{pick up tubes into box}, and \textit{transfer liquid into an erlenmeyer}, together with one single-arm task, \textit{make a burger}. These tasks cover bimanual coordination, object transfer, liquid manipulation, and multi-stage manipulation, providing challenges for evaluating perception and action generation in physical environments. We compare against policy baselines without predictive-representation guidance, including OpenVLA-OFT~\citep{openvla-oft} and $\pi_0$~\citep{black2024pi_0}. We further compare with input-centric approaches that incorporate geometric information into the policy input, including GeoVLA~\citep{sun2025geovla}, 3D-CAVLA~\citep{bhat20253d}, SpatialVLA~\citep{spatialvla}, and FALCON~\citep{zhang2025spatial}. In addition, we include prediction-centric approaches that provide future or predictive information to policy learning, including 4D-WAM~\citep{yang20264d}, ELAN4D~\citep{he2026elan4d}, QDepth-VLA~\citep{li2025qdepth}, and WoG~\citep{su2026world}. Implementation details and training configurations are provided in the Appendix.F and B.

\begin{table}[t]
\centering
\caption{\small Comparison of different policies on LIBERO and LIBERO-Plus.}
\label{tab:libero_and_plus}
\tiny
\setlength{\tabcolsep}{3pt}
\renewcommand{\arraystretch}{0.9}
\resizebox{\linewidth}{!}{%
\begin{tabular}{cccccc|c}
\toprule
\multirow{2}{*}[-0.4ex]{Methods}
& \multicolumn{5}{c|}{\tiny LIBERO}
& \multicolumn{1}{c}{\tiny LIBERO-Plus} \\
\cmidrule(lr){2-6}
\cmidrule(lr){7-7}
& Object & Spatial & Goal & Long & Avg.
& Total \\
\midrule
OpenVLA-OFT~\citep{openvla-oft} & 98.4 & 97.6 & 97.9 & 94.5 & 97.1 & 69.6 \\
$\pi_0$~\citep{black2024pi_0} & 98.8 & 96.8 & 95.8 & 85.2 & 94.2 & 53.6 \\
GeoVLA\citep{sun2025geovla} & 99.0 & 98.4 & 96.6 & 96.6 & 97.7 & - \\
3D-CAVLA \citep{bhat20253d} & 99.8 & 98.2 & 98.2 & 96.1 & 98.1 & - \\
SpatialForcing~\citep{li2026spatial} & 99.6 & \textbf{99.4} & \textbf{98.8} & 96.0 & 98.5 & - \\
4D-WAM\citep{yang20264d} & 99.5 & 98.8 & 98.6 & \textbf{97.4} & \textbf{98.6} & 71.0 \\
ELAN4D\citep{he2026elan4d} & 98.8 & 98.2 & 96.8 & 94.2 & 97.0 & 78.2 \\
\rowcolor{graybg}
Ours w/o CPR & 98.2 & 98.2 & 96.2 & 93.6 & 96.6 & 75.0 \\
\rowcolor{graybg}
Ours & \textbf{100.0} & 99.0 & 98.2 & 97.0 & \textbf{98.6} & \textbf{81.7} \\
\bottomrule
\end{tabular}%
}
\vspace{-0.3cm}
\end{table}

\begin{table}[t]
\centering
\caption{\small Comparison of different policies on WidowX Robot.}
\label{tab:widowx_robot}
\tiny
\setlength{\tabcolsep}{3pt}
\renewcommand{\arraystretch}{0.9}
\resizebox{\linewidth}{!}{%
\begin{tabular}{cccccc}
\toprule
Methods
& \makecell{Put\\Spoon}
& \makecell{Put\\Carrot}
& \makecell{Stack\\Cube}
& \makecell{Put\\Eggplant}
& Avg. \\
\midrule
OpenVLA-OFT\citep{openvla-oft} & 12.5 & 4.2 & 4.2 & \textbf{100.0} & 30.2 \\
$\pi_0$\citep{black2024pi_0} & 25.0 & 16.7 & 12.5 & 29.2 & 20.9 \\
SpatialVLA\citep{spatialvla} & 16.7 & 25.0 & 29.2 & \textbf{100.0} & 42.7 \\
FALCON\citep{zhang2025spatial} & 62.5 & 41.7 & 20.8 & \textbf{100.0} & 56.3 \\
QDepth-VLA\citep{li2025qdepth} & 82.0 & 57.5 & 39.6 & 95.0 & 68.5 \\
WoG\citep{su2026world} & 79.2 & 50.0 & 33.0 & 91.7 & 63.5 \\
\rowcolor{graybg}
Ours w/o CPR & 83.3 & 58.3 & 54.2 & 50.0 & 61.5 \\
\rowcolor{graybg}
Ours & \textbf{95.8} & \textbf{70.8} & \textbf{70.8} & 50.0 & \textbf{71.9} \\
\bottomrule
\end{tabular}%
}
\vspace{-0.8cm}
\end{table}

\subsection{Simulation and Real-World Results}
\noindent\textbf{Simulation Results on LIBERO and LIBERO-Plus.}
We first evaluate CPR on manipulation and under distribution shifts using LIBERO and LIBERO-Plus. As shown in Tab.~\ref{tab:libero_and_plus}, CPR improves the average LIBERO success rate from 96.6\% to 98.6\%, an absolute gain of 2.0 percentage points, with notable improvements on the \textit{Spatial} and \textit{Long} suites. We further evaluate the policy zero-shot on LIBERO-Plus without adaptation, where CPR increases the overall task-count-weighted success rate from 75.0\% to 81.7\%. The gains on both benchmarks indicate that complementary predictive guidance benefits manipulation under both nominal and shifted environments, providing structured geometric information together with action-aware cues for downstream control.

\noindent\textbf{Simulation Results on WidowX.}
We next evaluate the method in SimplerEnv using the WidowX robot. As reported in Tab.~\ref{tab:widowx_robot}, CPR guidance raises the average success rate from 61.5\% to 71.9\%, corresponding to an absolute gain of 10.4 percentage points. The improvement on \textit{Stack Cube} is especially informative because successful stacking requires accurate reasoning about the relative configuration of the manipulated cube and its target, as well as coordinated actions as the scene evolves. The gain on this task is consistent with the intended roles of the two representations: predictive geometric guidance provides structured information about scene configuration and its evolution, while motor guidance emphasizes action-aware transitions associated with manipulation.


\begin{figure}[t]
    \centering
    \includegraphics[width=0.9\linewidth]{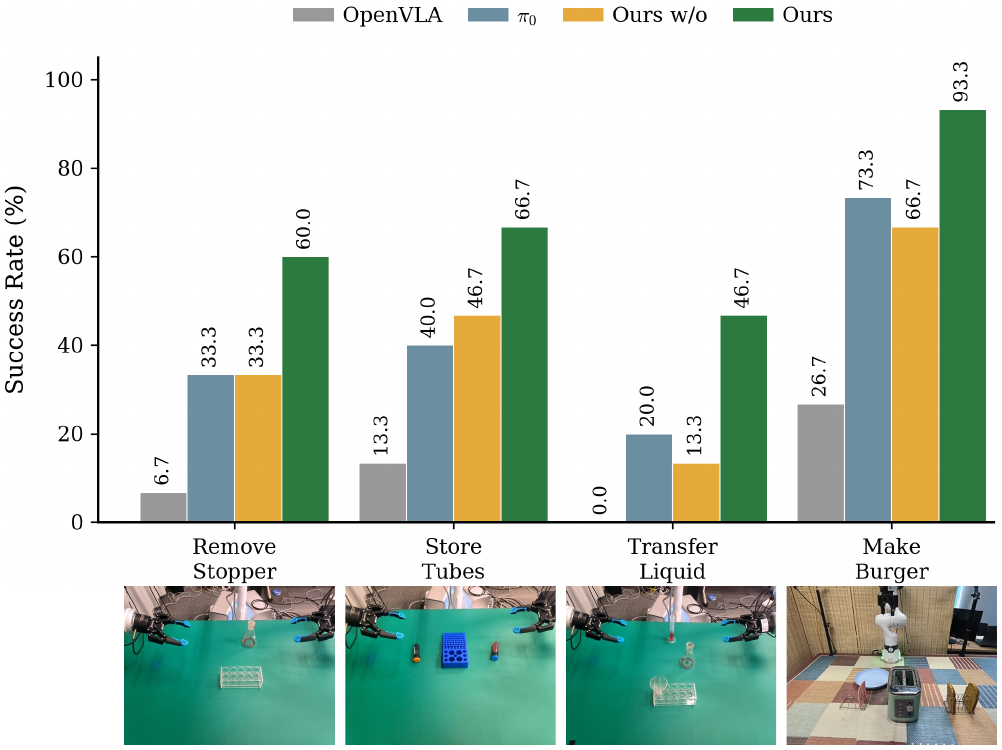}
    \caption{\small Performance comparison on real-robot manipulation tasks.}
    \label{real}
    \vspace{-0.9cm}
\end{figure}
\noindent\textbf{Real-world Evaluation.}
We evaluate CPR on four real-world manipulation tasks shown in Fig.~\ref{real}. The CPR-guided model consistently outperforms the baselines across tasks. The improvements are particularly pronounced on fine-grained manipulation tasks, such as \textit{Remove Stopper} and \textit{Transfer Liquid}, where precise spatial reasoning and action execution are required. By providing fine-grained geometric perception together with predictive information about future states, CPR enables the policy to better anticipate object configurations and execute more precise interactions. These results demonstrate that the complementary predictive guidance learned by CPR translates from simulation to real-world manipulation, particularly in tasks requiring accurate and fine-grained control.

\vspace{-0.25cm}
\subsection{Transfer Across Diverse Policy Paradigms} 

\begin{figure}[t]
    \centering
    \includegraphics[width=0.95\linewidth]{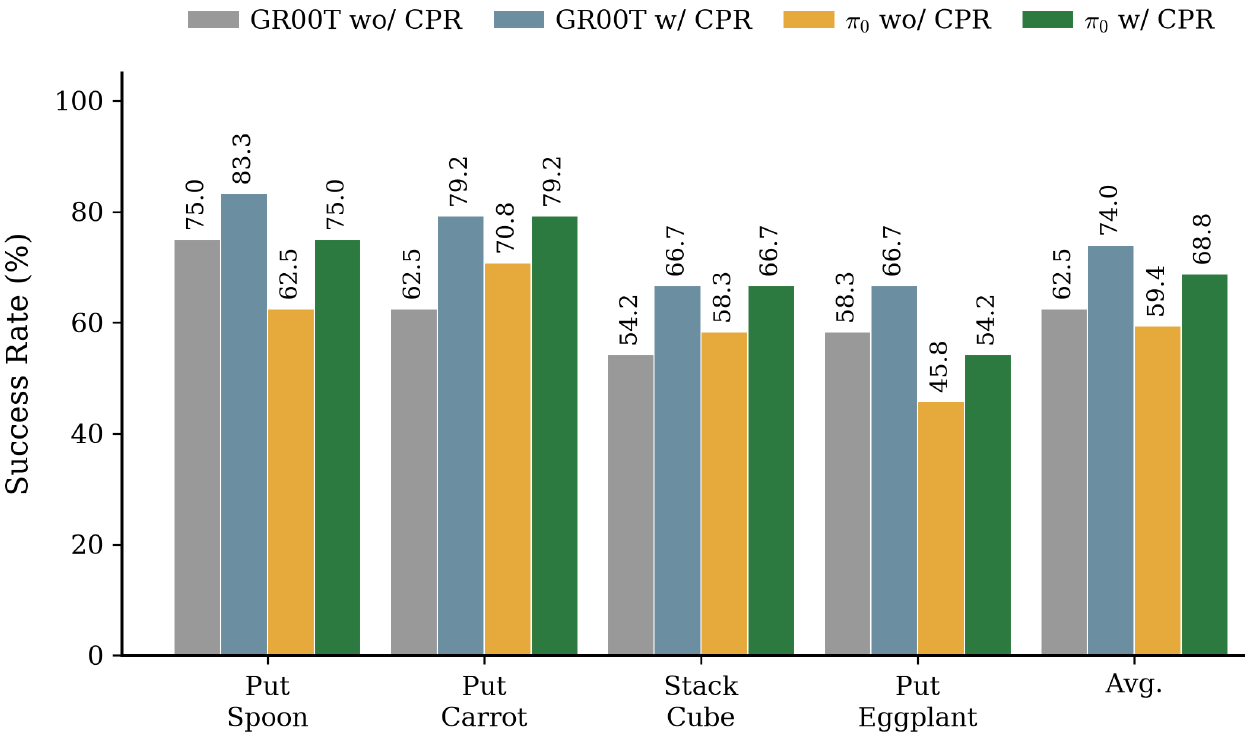}
    \caption{\small Transfer of CPR to different downstream VLA models on the WidowX robot.}
    \label{tab:model_performance}
    \vspace{-0.3cm}
\end{figure}

\noindent\textbf{Transfer to Diverse VLA Models.}
To examine whether CPR depends on a particular downstream policy architecture, we integrate the same representation guidance into two additional VLA models, GR00T-style~\citep{bjorck2025gr00t} and $\pi_0$-style models~\citep{black2024pi_0}. The representation pretraining procedure and the roles of the two representations remain unchanged, while their transfer interfaces are adapted to the corresponding downstream architectures. As shown in Fig.~\ref{tab:model_performance}, adding the same alignment objective improves the average success rate of GR00T from 62.5\% to 74.0\%. Similarly, $\pi_0$ improves from 59.4\% to 68.8\%, corresponding to a gain of 9.4 percentage points. Together with the gains observed for OFT, these results indicate that the proposed guidance is compatible with substantially different VLA architectures and is not tied to a single action-generation mechanism.

\begin{table}[t]
  \centering
  \captionsetup{justification=centering}
  \caption{\small Transfer of CPR to FastWAM on the LIBERO.}
  \label{tab:fastwam_performance}

  \resizebox{1.0\linewidth}{!}{%
    \begin{tabular}{cccccc}
      \toprule
      Methods & Object & Spatial & Goal & Long & Avg. \\
      \midrule
      FastWAM wo/ CPR & \textbf{100.0} & 98.2 & 97.2 & 95.2 & 97.6 \\
      FastWAM w/ CPR  & \textbf{100.0} & \textbf{99.0} & \textbf{97.4} & \textbf{97.8} & \textbf{98.5} \\
      \bottomrule
    \end{tabular}%
  }
  \vspace{-0.8cm}
\end{table}

\noindent\textbf{Transfer to World Action Models.}
\label{fastwam}
We further investigate whether complementary predictive representations remain useful for the WAM paradigm. We integrate CPR into FastWAM \citep{yuan2026fastwam}, which jointly models future world evolution and robot actions. To preserve the respective roles of the two representations, we introduce an additional geometry DiT branch in parallel with the original world and action modeling branches for predictive geometric modeling, while using the motor representation to provide action-aware guidance to the action branch. As shown in Tab.~\ref{tab:fastwam_performance}, incorporating CPR improves the average LIBERO success rate of FastWAM from 97.6\% to 98.5\%. Although FastWAM already explicitly models future world evolution, these results suggest that structured predictive geometry and action-aware guidance can provide complementary information beyond the original world-action modeling objective. Combined with the results across three VLA architectures, this experiment further demonstrates the broad compatibility of CPR across substantially different policies. More details are provided in the Appendix.C.5.

\subsection{Analysis of CPR Design and Transfer}


\begin{table}[!t]
\centering
\caption{\small Ablation of CPR guidance on the WidowX evaluation. PGR and MR denote the predictive geometric and motor representations, respectively.}
\label{tab:complementary_rep_ablation}
\scriptsize
\setlength{\tabcolsep}{3pt}
\renewcommand{\arraystretch}{1.0}
\resizebox{\linewidth}{!}{%
\begin{tabular}{ccc|ccccc}
    \toprule
        & PGR & MR
        & \makecell{Put\\Spoon}
        & \makecell{Put\\Carrot}
        & \makecell{Stack\\Cube}
        & \makecell{Put\\Eggplant}
        & Avg. \\
    \midrule
    (a) &            &            & 83.3          & 58.3          & 54.2          & 50.0          & 61.5 \\
    (b) & \checkmark &            & 79.2          & 66.7          & 66.7          & \textbf{54.2} & 66.7 \\
    (c) &            & \checkmark & 91.7          & 62.5          & 62.5          & \textbf{54.2} & 67.7 \\
    (d) & \checkmark & \checkmark & \textbf{95.8} & \textbf{70.8} & \textbf{70.8} & 50.0          & \textbf{71.9} \\
    \bottomrule
\end{tabular}%
}
\vspace{-0.3cm}
\end{table}

\begin{table}[!t]
\centering
\caption{\small Ablation of representation organization on the WidowX evaluation.}
\label{tab:performance_comparison}
\scriptsize
\setlength{\tabcolsep}{3.5pt}
\renewcommand{\arraystretch}{1.0}
\resizebox{\linewidth}{!}{%
\begin{tabular}{cccccc}
    \toprule
    Methods
    & \makecell{Put\\Spoon}
    & \makecell{Put\\Carrot}
    & \makecell{Stack\\Cube}
    & \makecell{Put\\Eggplant}
    & Avg. \\
    \midrule
    Baseline
    & 83.3 & 58.3 & 54.2 & \textbf{50.0} & 61.5 \\
    Unified
    & 91.7 & \textbf{70.8} & 62.5 & 45.8 & 67.7 \\
    Complementary
    & \textbf{95.8} & \textbf{70.8} & \textbf{70.8} & \textbf{50.0} & \textbf{71.9} \\
    \bottomrule
\end{tabular}%
}
\vspace{-0.3cm}
\end{table}

\noindent\textbf{Effect of Complementary Predictive Representation Guidance.}
To examine complementary contributions of the two predictive representations, we conduct an ablation study on the WidowX evaluation, with results reported in Tab.~\ref{tab:complementary_rep_ablation}. Variant (a) corresponds to the VLA baseline without representation guidance. Introducing the predictive geometric representation in variant (b) improves the average success rate from 61.5\% to 66.7\%, yielding a gain of 5.2\%. Variant (c) instead transfers the motor representation by aligning it with the intermediate features used for action generation. This variant improves the average success rate to 67.7\%, outperforming the baseline by 6.2\%. This result supports motor representation in providing action-aware predictive guidance to the action-generation pathway. Finally, combining both representations in variant (d) achieves the best average success rate of 71.9\%, corresponding to a 10.4\% absolute improvement over the baseline. Combining predictive geometric and motor representation allows the downstream policy to benefit from predictive geometric information and action-aware dynamics, leading to the strongest performance.

\begin{figure}[t]
    \centering
    \includegraphics[width=1.0\linewidth]{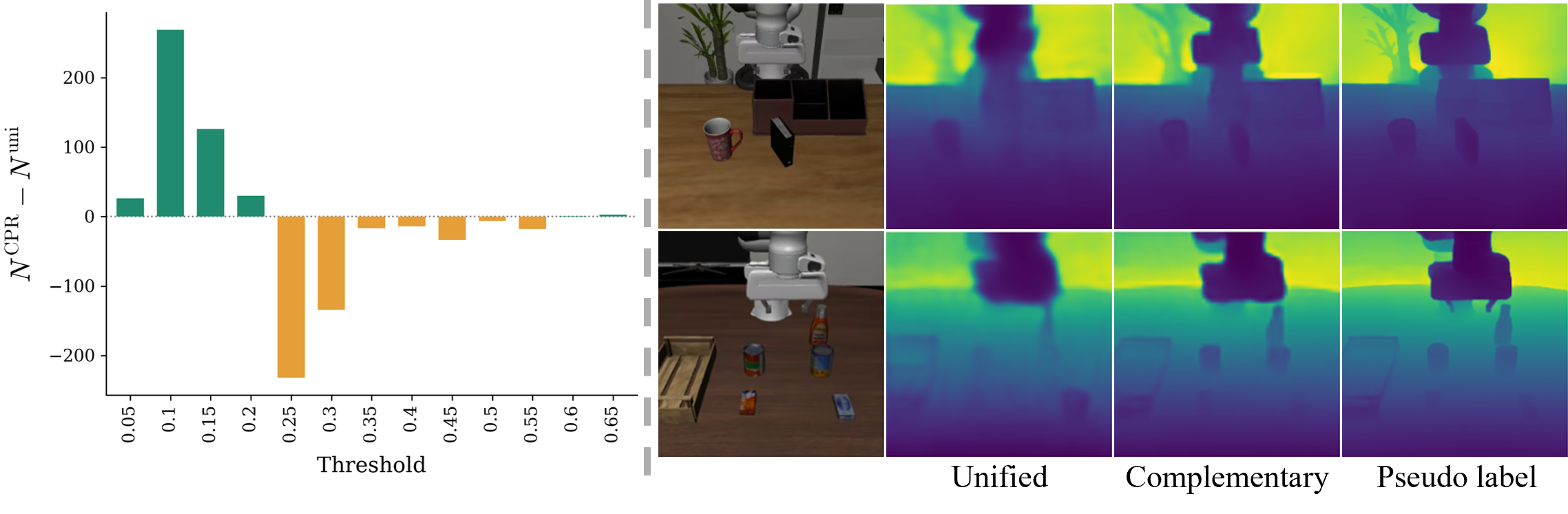}
    \caption{\small Probing analysis of representation organization on LIBERO-90. The top compares action-error distributions between the CPR motor and unified representations, while the bottom shows depth-probing results.}
    \label{unified_probe}
    \vspace{-0.3cm}
\end{figure}

\noindent\textbf{Effect of Representation Organization.}
\label{action_probe}
We examine whether assigning predictive geometry and action-aware information to complementary representations is beneficial. In the unified variant, the motor branch is removed, and the predictive geometric representation is jointly supervised by the action objective and aligned with intermediate features of the downstream VLA. As shown in Tab.~\ref{tab:performance_comparison}, the unified representation improves the average success rate from 61.5\% to 67.7\%, but remains below our complementary design at 71.9\%, which indicates that providing complementary representations for structured predictive geometry and action-aware information is more effective than asking a single representation to serve both roles under training setup. To further probe encoded information, we evaluate the pretrained representations zero-shot on LIBERO-90\citep{Libero}. As shown in Fig.~\ref{unified_probe}, the motor representation concentrates more samples in the low action-error region, while the predictive geometric representation better preserves scene structure in depth probing, where the pseudo labels are generated by DepthAnything\citep{lin2025depth}. The latter recovers more faithful object geometry and relative depth relationships compared with the unified representation. Together, these results suggest that assigning distinct roles to predictive geometric and motor representations helps preserve both geometric fidelity and action-relevant information, supporting the complementary design of CPR. More details are provided in Appendix.A.3.

\begin{table}[t]
\centering
\caption{\small Comparison with different predictive representation learning methods under a unified downstream policy framework on the WidowX evaluation.}
\label{tab:predictive_rep_comparison}
\scriptsize
\setlength{\tabcolsep}{3.5pt}
\renewcommand{\arraystretch}{1.0}
\resizebox{\linewidth}{!}{%
\begin{tabular}{cccccc}
    \toprule
    Methods
    & \makecell{Put\\Spoon}
    & \makecell{Put\\Carrot}
    & \makecell{Stack\\Cube}
    & \makecell{Put\\Eggplant}
    & Avg. \\
    \midrule
    WoG~\citep{su2026world}
    & 83.3 & 62.5 & 37.5 & 75.0 & 64.6 \\
    FutureVLA~\citep{xu2026futurevla}
    & 79.2 & 70.8 & 41.7 & \textbf{79.2} & 67.7 \\
    RepWAM~\citep{wang2026repwam}
    & 87.5 & \textbf{75.0} & 54.2 & 58.3 & 68.8 \\
    Ours
    & \textbf{95.8} & 70.8 & \textbf{70.8} & 50.0 & \textbf{71.9} \\
    \bottomrule
\end{tabular}%
}
\vspace{-0.8cm}
\end{table}

\noindent\textbf{Comparison with Predictive Representation Learning Methods.}
Existing predictive representation learning methods differ in both their learned representations and downstream policy architectures, making direct comparisons potentially confounded by differences in the overall training pipeline. To isolate the effect of predictive representation learning, we therefore integrate several representative approaches into the same downstream VLA framework and evaluate them under a unified setting on the WidowX benchmark. Specifically, we match the pretraining data scale and number of optimization steps to those of CPR, followed by the same downstream post-training procedure. This ensures consistency in both pretraining budgets and downstream adaptation. As shown in Tab.~\ref{tab:predictive_rep_comparison}, our method achieves the highest average success rate of 71.9\%, compared with 64.6\% for WoG~\citep{su2026world}, 67.7\% for FutureVLA~\citep{xu2026futurevla}, and 68.8\% for RepWAM~\citep{wang2026repwam}. Since these methods are evaluated with the same downstream policy framework, the comparison provides a more controlled assessment of the predictive representations used to guide policy learning. The result suggests that the benefit of predictive guidance depends not only on exposing the policy to future information, but also on how that information is organized for downstream control. Our method provides structured predictive geometric information together with action-aware information extracted from geometric transitions, allowing these complementary signals to be transferred through dedicated supervision pathways.

\begin{figure}[!t]
\centering
\includegraphics[width=1.0\linewidth]{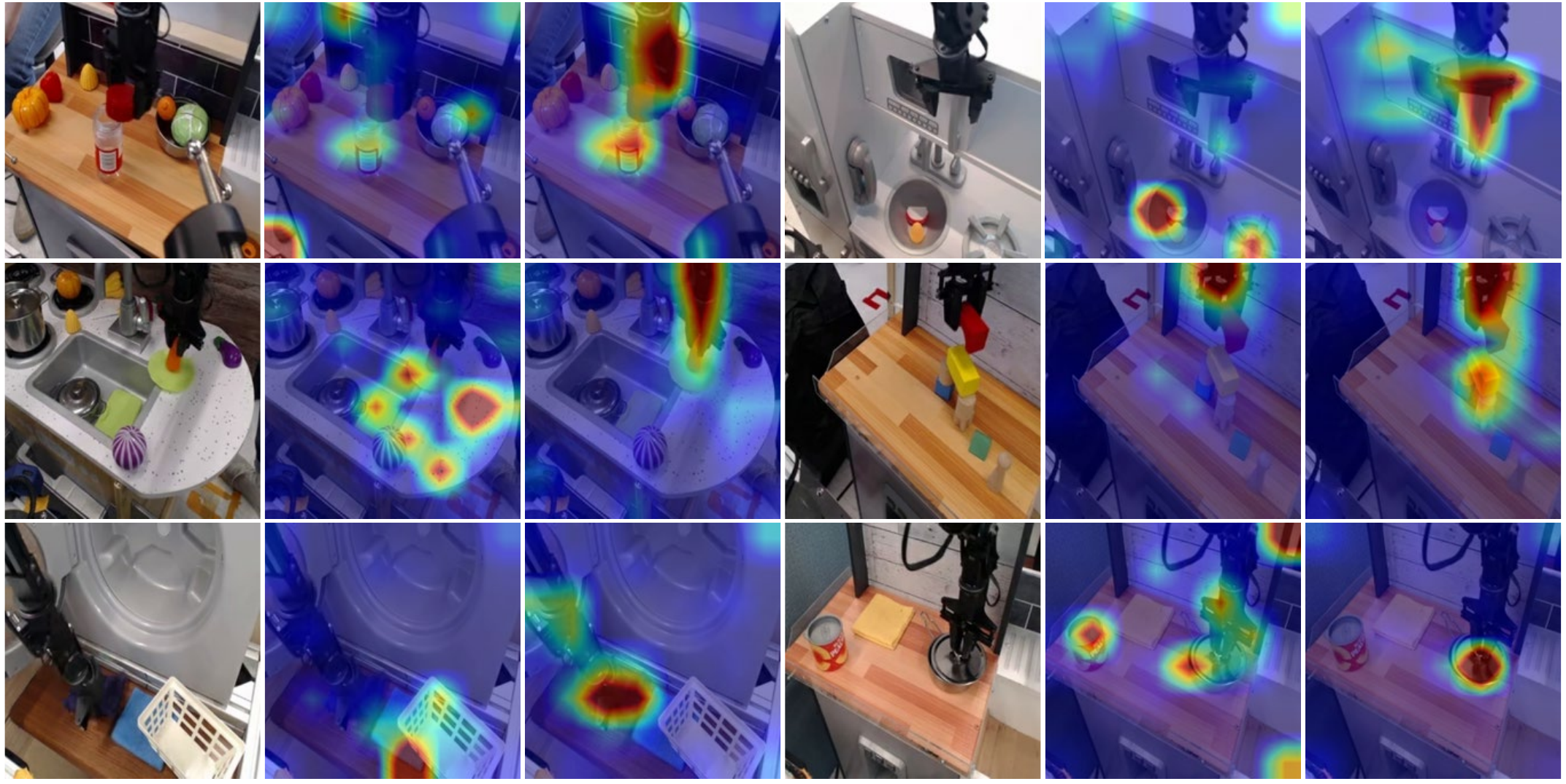}
\caption{\small Attention-map comparison between the unguided baseline (middle) and the CPR-guided policy (right). CPR guidance
produces more concentrated responses around task-relevant objects
and interaction regions.}
\label{attention_map}
\vspace{-0.5cm}
\end{figure}

\noindent\textbf{Qualitative Analysis of Visual Attention.}
We visualize the attention maps of the baseline and CPR-guided policies under the same observations and language instructions. As shown in Fig.~\ref{attention_map}, the baseline exhibits diffuse attention, whereas the CPR-guided model more consistently concentrates on task-relevant objects, target regions, and surrounding spatial context. In particular, stronger localization is observed around manipulated objects and regions involved in the ongoing interaction. These qualitative observations are consistent with the intended roles of the two guidance signals: predictive geometric supervision encourages structured scene understanding, while motor guidance emphasizes information associated with action-aware transitions. Although attention maps do not directly measure policy reasoning, they provide qualitative evidence that CPR can encourage more task-focused visual representations.

\vspace{-0.2cm}
\section{Conclusion}
\vspace{-0.2cm}
We study complementary predictive representation learning to connect evolving scene geometry with action generation. A predictive geometric representation captures structured scene evolution, while a motor representation emphasizes action-relevant geometric transitions; the two are transferred to downstream policies through dedicated guidance pathways without requiring future observations at inference time. Experiments across simulated and real-world manipulation settings demonstrate consistent gains across VLA architectures and WAMs, including improvements in both manipulation performance and future world/action prediction. These results highlight the importance of organizing predictive geometric information according to its role in downstream control.

\section{LLM Usage Statement}
In this work, we used LLM-based assistants solely to polish the wording of the text. We did not use generative AI tools to develop research ideas or design experiments. All AI-assisted work was carefully reviewed by the authors. Specifically, all AI-assisted text was checked, corrected, and approved by the authors, and all AI-assisted code was verified for correctness. Every reported result was generated through the authors' own runs. The authors take full responsibility for the final content of this work, including all text, claims, and artifacts produced with the assistance of generative AI.


\appendix

\makeappendixcontents

\clearpage

\appsection{Additional Experiments}

\appsubsection{\textbf{(Q1):} Simulation and Real-World Results}

\noindent\textbf{Full LIBERO-Plus Results.}
We provide the detailed results on the seven perturbation types of LIBERO-Plus in Tab.~\ref{tab:libero_plus}, which provide a more detailed view of the robustness gains reported in the main experiments.


\begin{figure}[htbp]
  \centering
  \includegraphics[width=1.0\linewidth]{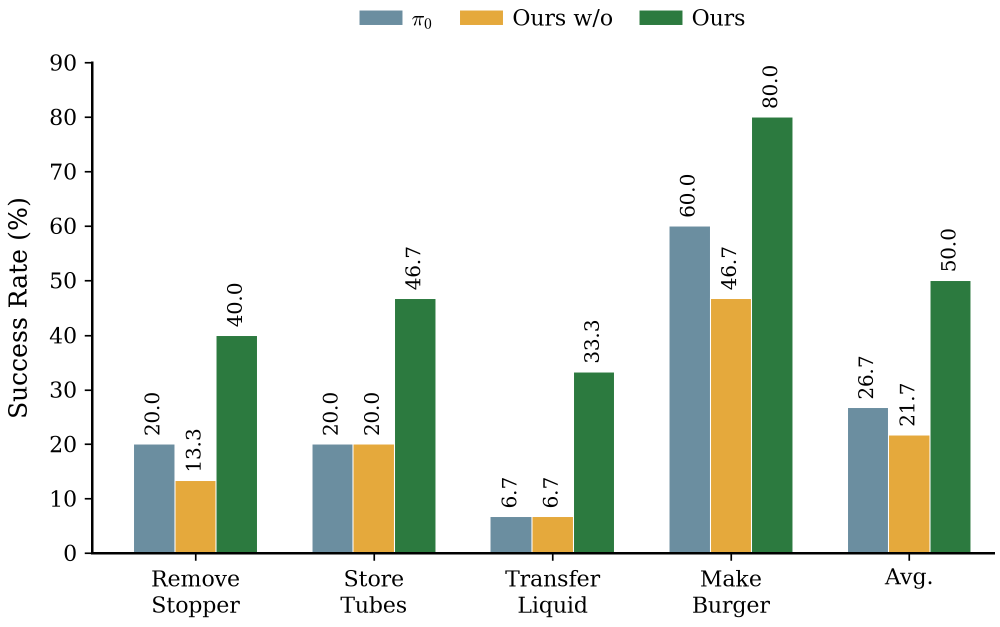}
  \caption{\small Out-of-Distribution success rates on real-world tasks under visual and spatial perturbations.}
  \label{ood}
\end{figure}
\begin{figure*}[!t]
\centering
\includegraphics[width=1.0\linewidth]{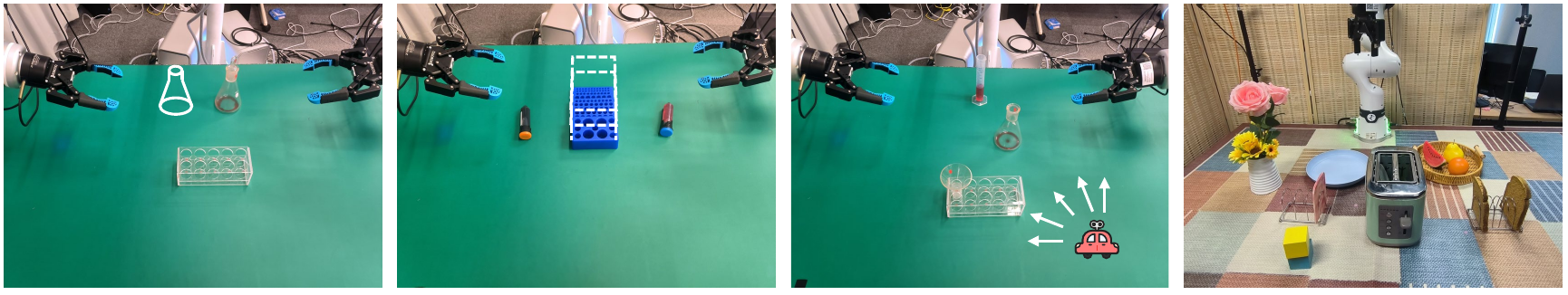}
\caption{\small Visualization of the real-world out-of-distribution perturbations. From left to right: \textit{Remove Stopper}, with randomly shifted flask positions; \textit{Store Tubes}, with randomly varied target-box heights; \textit{Transfer Liquid}, with a randomly moving toy car as a dynamic visual distractor; and \textit{Make a Burger}, with additional irrelevant objects added to the workspace. These perturbations introduce changes in object placement, target geometry, dynamic visual context, and scene clutter while preserving the underlying manipulation tasks.}
\label{real_pic}
\end{figure*}

\noindent\textbf{Real-World Out-of-Distribution Generalization.}
We further evaluate the ability of CPR to generalize to unseen visual and spatial conditions without additional training. Starting from the same real-world tasks described above, we introduce task-specific perturbations that change the observation distribution while preserving the underlying manipulation objective. Specifically, as shown in Fig.\ref{real_pic}, for \textit{Remove Stopper}, we shift the position of the erlenmeyer flask to alter the relative geometry between the two arms and the manipulated object. For \textit{Store Tubes}, we randomly vary the height of the target box, requiring the policy to adapt to changes in the target geometry. For \textit{Transfer Liquid}, we introduce a randomly moving toy car as a dynamic visual distractor during execution, increasing the difficulty of separating task-relevant objects from irrelevant scene motion. For \textit{Make a Burger}, we add multiple irrelevant objects to the workspace, introducing additional visual clutter.

These perturbations respectively introduce changes in object placement, target geometry, dynamic visual context, and scene clutter, while keeping the underlying task semantics unchanged. As shown in Fig.~\ref{ood}, CPR achieves the highest success rate on all four tasks and improves the overall average success rate from 21.7\% for the baseline without CPR to 50.0\%. The consistent improvement under these unseen conditions suggests that complementary predictive representation guidance can provide useful geometric and action-aware information beyond the nominal training distribution, helping the downstream policy maintain effective manipulation behavior under changes in visual and spatial context.

\begin{table*}[!t]
\centering
\caption{\small Comparison of different VLA models on LIBERO-Plus. 
Success rates (\%) are reported under the seven perturbation types.
Total denotes the official task-count-weighted mean over all 10,030 perturbed tasks,
rather than an unweighted mean of the seven categories.}
\label{tab:libero_plus}
\scriptsize
\setlength{\tabcolsep}{2pt}
\renewcommand{\arraystretch}{0.95}
\resizebox{0.75\linewidth}{!}{%
\begin{tabular}{ccccccccc}
\toprule
Methods & Cam. & Rob. & Lang. & Light & Back. & Noise & Layout & Total \\
\midrule
OpenVLA-OFT\citep{openvla-oft}    & 56.4 & 31.9 & 79.5 & 88.7 & 93.3 & 75.8 & 74.2 & 69.6 \\
$\pi_0$~\citep{black2024pi_0}        & 13.8 & 6.0  & 58.8 & 85.0 & 81.4 & 79.0 & 68.9 & 53.6 \\
WorldVLA\citep{worldvla}           & 0.1  & 27.9 & 41.6 & 43.7 & 17.1 & 10.9 & 38.0 & 25.0 \\
UniVLA\citep{bu2025univla}         & 1.8  & 46.2 & 69.6 & 69.0 & 81.0 & 21.2 & 31.9 & 43.9 \\
4D-WAM\citep{yang20264d}           & 45.2 & 64.3 & \textbf{90.6} & \textbf{94.3} & 57.7 & 69.1 & 79.2 & 71.0 \\
ELAN4D\citep{he2026elan4d}                               & 63.7 & \textbf{70.7} & 77.8 & 89.8 & 91.4 & 79.9 & \textbf{81.4} & 78.2 \\
\rowcolor{graybg}
Ours w/o CPR                         & 47.6 & 59.2 & 86.5 & 94.2 & \textbf{94.9} & 76.1 & 78.8 & 75.0 \\
\rowcolor{graybg}
Ours                                 & \textbf{77.5} & 60.8 & 88.6 & 86.1 & 89.1 & \textbf{95.6} & 77.2 & \textbf{81.7} \\
\bottomrule
\end{tabular}%
}
\end{table*}

\appsubsection{\textbf{(Q2):} Transfer Across Diverse Policy Paradigms} 

\noindent\textbf{Quantitative Analysis of FastWAM.}
Beyond downstream task success in Sec.4.2 in main paper, we examine whether CPR improves the predictive capabilities of WAMs themselves. We freeze the FastWAM models trained on LIBERO and evaluate them on LIBERO-90 without additional adaptation. Future visual prediction is evaluated using PSNR, SSIM, and LPIPS, while action prediction is measured by the $\ell_1$ error between predicted and ground-truth actions. As shown in Tab.~\ref{tab:metrics_comparison}, CPR consistently improves both future visual prediction and action prediction. PSNR increases from 19.67 to 22.91, SSIM improves from 0.764 to 0.803, and LPIPS decreases from 0.229 to 0.186. The action $\ell_1$ error decreases from 0.124 to 0.075, corresponding to a relative reduction of approximately 39.5\%. These results provide evidence that the proposed representation guidance improves not only downstream manipulation performance but also the predictive quality of the underlying world-action model. The improvement in visual prediction is consistent with the structured geometric supervision provided by the predictive geometric representation, while the reduction in action error is consistent with the action-aware information emphasized by the motor representation. Together, they provide complementary predictive guidance for jointly modeling future world evolution and robot actions.

\begin{table*}[!t]
\centering
\caption{\small Zero-shot future-world and action prediction performance of FastWAM on LIBERO-90. PSNR and SSIM are higher-is-better, while LPIPS and action $\ell_1$ error are lower-is-better.}
\label{tab:metrics_comparison}
\begin{tabular}{ccccc}
\toprule
Model
& PSNR$\uparrow$
& SSIM$\uparrow$
& LPIPS$\downarrow$
& Action $\ell_1\downarrow$ \\
\midrule
FastWAM w/o CPR
& 19.67 & 0.764 & 0.229 & 0.124 \\
FastWAM w/ CPR
& \textbf{22.91} & \textbf{0.803} & \textbf{0.186} & \textbf{0.075} \\
\bottomrule
\end{tabular}
\end{table*}

\noindent\textbf{Qualitative Analysis of FastWAM.}
\label{sec:more_qualitative_fastwam}
We provide additional qualitative comparisons to further examine the improvements observed in the zero-shot world- and action-prediction evaluation on LIBERO-90. As illustrated in Fig.~\ref{fig:fast_libero_90}, the differences between FastWAM with and without CPR are evident in both geometric fidelity and action execution.

In Fig.~\ref{fig:fast_libero_90}(a), we consider a scene in which the robot is preparing to grasp a can. Without complementary predictive representation guidance, the predicted future frames exhibit a noticeable discrepancy in the estimated geometric configuration, particularly in the relative position between the gripper and the target object. As a consequence, the generated gripper trajectory does not reach the appropriate grasp location, resulting in an action prediction that is inconsistent with the target interaction. With CPR guidance, the predicted frames more accurately preserve the geometric configuration of the object and gripper, allowing the generated trajectory to approach the can at a more appropriate location. This qualitative example is consistent with the role of the predictive geometric representation in providing structured geometric information for future scene prediction. We observe similar differences in more temporally extended interactions, as shown in (b) and (c). Without CPR, the generated videos may exhibit missing or ambiguous object structures and progressively deviate from the expected scene evolution. At the same time, the predicted actions can become inconsistent with the intended manipulation sequence. For example, as shown in (c), in a drawer-opening interaction, the baseline predictions may fail to clearly represent the drawer's changing configuration and do not adequately capture the corresponding opening motion. In contrast, FastWAM with CPR maintains more coherent object configurations over time and produces predictions that more faithfully reflect the evolving manipulation process.

These observations provide a qualitative counterpart to the quantitative improvements reported in Tab.~\ref{tab:metrics_comparison}. The predictive geometric representation encourages the model to preserve structured geometric information during future-world modeling, which can improve the fidelity of object shape, relative spatial configuration, and scene evolution in the generated frames. Meanwhile, the motor representation emphasizes action-aware information from geometric transitions, providing guidance that is more directly associated with the execution of the corresponding manipulation. The combination therefore allows the model to better maintain both the evolving geometric state of the scene and the action sequence associated with that evolution. Overall, the qualitative results suggest that CPR improves the consistency between predicted future geometry and predicted robot actions. Rather than improving only the visual appearance of individual future frames, the complementary guidance helps the model maintain a more coherent correspondence between \emph{what the scene is expected to become} and \emph{what action should be taken as it evolves}. This observation is consistent with the complementary roles of the predictive geometric and motor representations in our framework.


\begin{figure*}[p]
    \centering
    \includegraphics[width=0.9\textwidth]{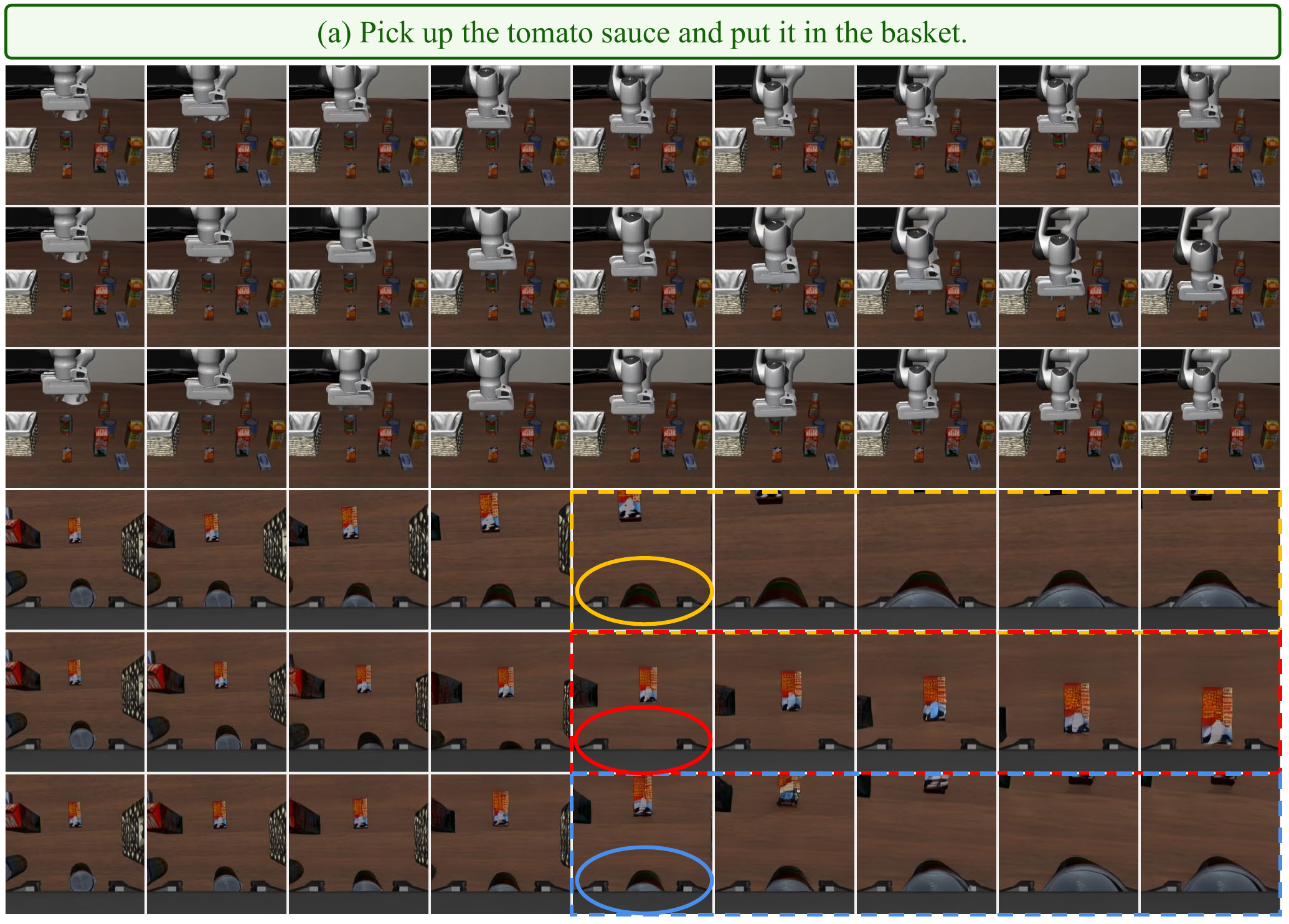}

    \captionsetup{labelformat=empty}
    \caption{}
\end{figure*}

\clearpage

\begin{figure*}[p]
    \ContinuedFloat
    \centering
    \includegraphics[width=0.85\textwidth]{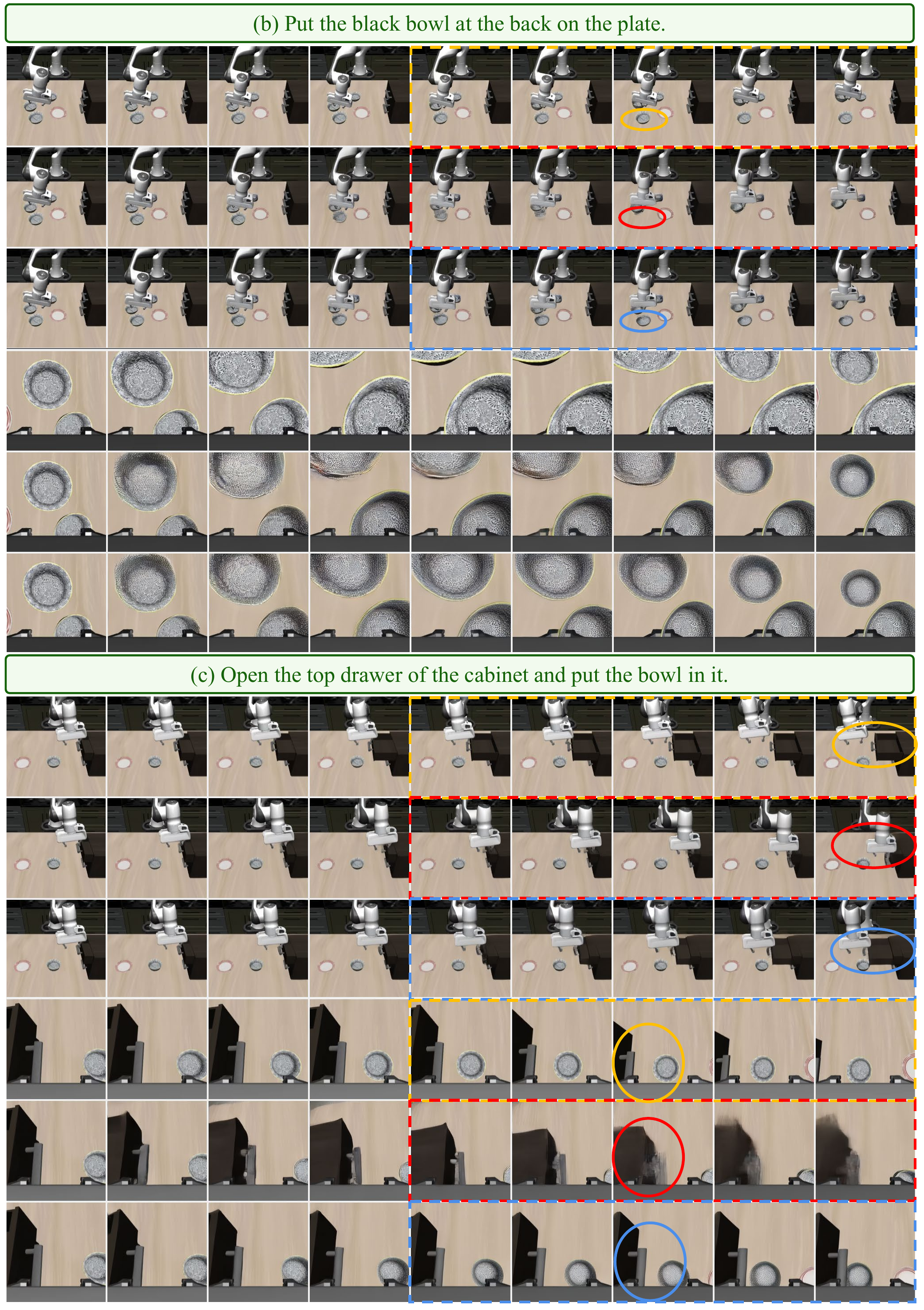}

    \caption{\small Qualitative comparison of zero-shot future world prediction on LIBERO-90. Each example contains a main camera view (upper half) and a wrist camera view (lower half), with columns progressing over nine predicted future frames. Within each view, the three rows from top to bottom correspond to ground-truth observations (GT), FastWAM without CPR guidance (w/o CPR), and FastWAM with CPR guidance (w/ CPR).}

    \label{fig:fast_libero_90}
\end{figure*}

\clearpage

\appsubsection{\textbf{(Q3):} Analysis of Representation Design and Transfer}

\begin{table}[!t]
\centering
\caption{\small Ablation of the pretraining objectives for the predictive geometric representation on the WidowX evaluation.}
\label{tab:pretraining_objective_ablation}

\setlength{\tabcolsep}{3pt}
\renewcommand{\arraystretch}{0.95}

\resizebox{\columnwidth}{!}{
\begin{tabular}{cccc|ccccc}
\toprule 
& $\mathcal{L}_{\text{rec}}$ & $\mathcal{L}_{\text{mask}}$ & $\mathcal{L}_{\text{trans}}$ 
& \makecell{Put\\Spoon} & \makecell{Put\\Carrot} & \makecell{Stack\\Cube} & \makecell{Put\\Eggplant} & Avg. \\ 
\midrule
(a) & & & & 83.3 & 58.3 & 54.2 & \textbf{50.0} & 61.5 \\ 
(b) & \checkmark & & & 87.5 & 66.7 & 58.3 & \textbf{50.0} & 65.6 \\ 
(c) & \checkmark & \checkmark & & 91.7 & \textbf{70.8} & 66.7 & 45.8 & 68.8 \\
(d) & \checkmark & \checkmark & \checkmark & \textbf{95.8} & \textbf{70.8} & \textbf{70.8} & \textbf{50.0} & \textbf{71.9} \\ 
\bottomrule
\end{tabular}
}
\end{table}

\noindent\textbf{Effect of Predictive Representation Pretraining Objectives.}
We further examine the contribution of the three objectives used to learn the predictive geometric representation, as shown in Tab.~\ref{tab:pretraining_objective_ablation}. Starting from the downstream baseline, introducing geometry reconstruction $\mathcal{L}_{\mathrm{rec}}$ together with the action-supervised motor branch improves the average success rate from 61.5\% to 65.6\%, indicating that preserving geometric structure already provides useful guidance for downstream control. Adding masked geometric state prediction $\mathcal{L}_{\mathrm{mask}}$ further increases the performance to 68.8\%, suggesting that contextual prediction across temporal states helps the representation capture richer temporal dependencies. Finally, incorporating geometric transition prediction $\mathcal{L}_{\mathrm{trans}}$ yields the best result of 71.9\%. This additional gain indicates that explicitly modeling transitions between consecutive geometric states provides further information about scene evolution that is useful for downstream action generation. Overall, the progressive improvements support the complementary roles of structural, contextual, and transition-level supervision in shaping the predictive geometric representation.

\begin{table}[!t]
\centering
\caption{\small Ablation study of the geometry-guidance loss weight $\lambda_{\mathrm{geo}}$ on the WidowX evaluation.}
\label{tab}
\begin{tabular}{cccccc}
\toprule
$\lambda_{\mathrm{geo}}$ & 0.1 & 0.3 & 0.5 & 0.7 & 0.9 \\
\midrule
SR (\%)
& 69.8 & \textbf{71.9} & 70.8 & 68.8 & 65.6 \\
\bottomrule
\end{tabular}
\end{table}

\noindent\textbf{Effect of the Geometry-Guidance Weight.}
We investigate the effect of the geometry-guidance loss weight $\lambda_{\mathrm{geo}}$ in the downstream post-training objective on the WidowX evaluation. As shown in Tab.~\ref{tab}, the average success rate initially improves as the weight increases from 0.1 to 0.3, reaching the best performance of 71.9\%. Further increasing $\lambda_{\mathrm{geo}}$ leads to a gradual performance degradation. These results suggest that the predictive geometric supervision is most effective when its contribution is appropriately balanced with the native action-generation objective. A relatively small geometry-guidance weight may provide insufficient supervision for transferring the structured predictive geometric information, whereas overly strong geometric supervision may place excessive emphasis on representation matching and reduce the relative influence of the action-generation objective.

\begin{figure}[!t]
  \centering
  \includegraphics[width=\linewidth]{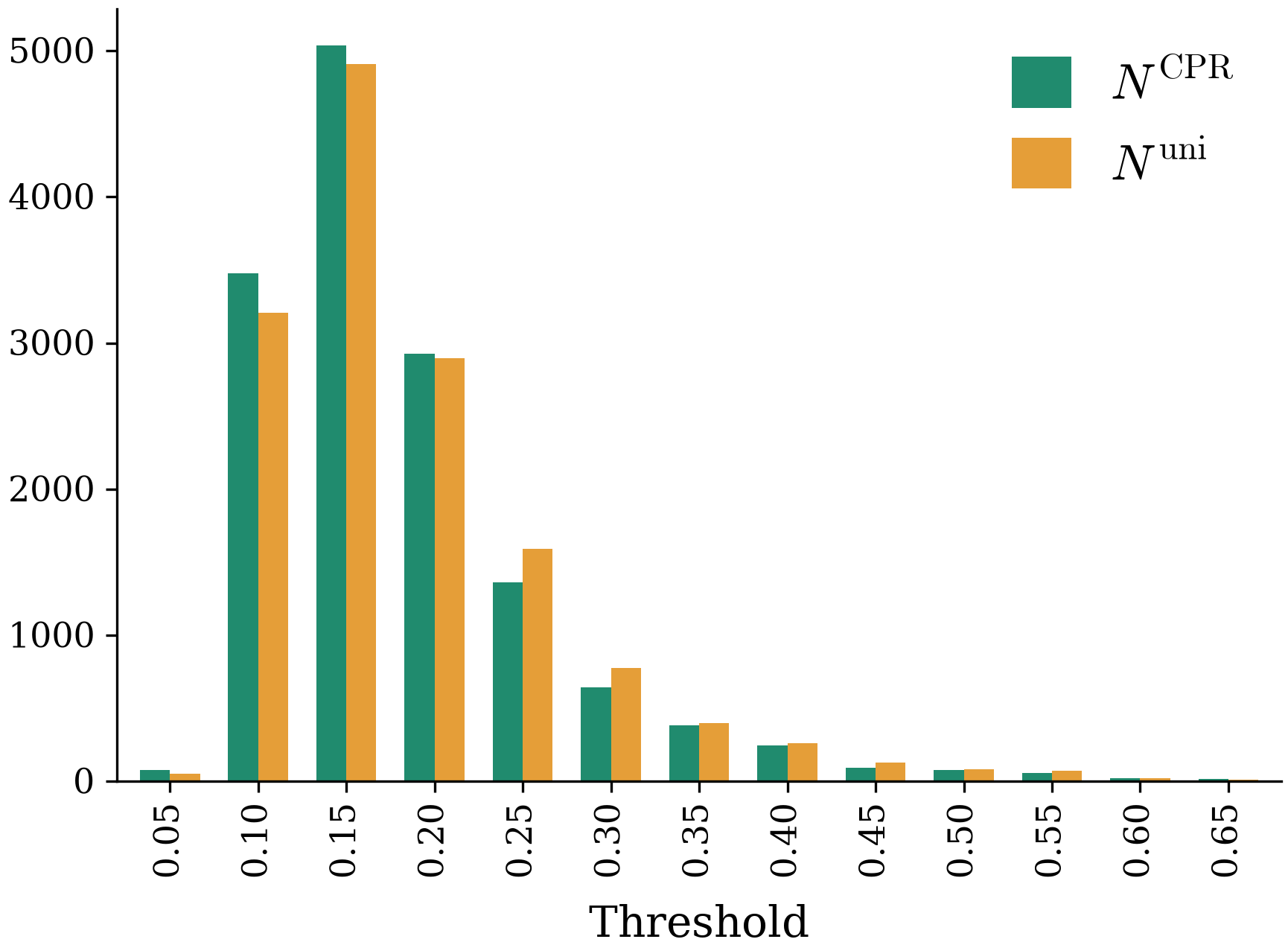}
  \caption{\small Distribution of action-probing experiment.}
  \label{total_number_action_prob}
\end{figure}

\noindent\textbf{More Details of Representation Organization.}
\label{RO}
We provide additional details of the representation-organization analysis reported in Sec.4.3 in main paper. For action probing, we visualize the number of samples falling into each $\ell_1$-loss bin for the different representation variants. As shown in Fig.~\ref{total_number_action_prob}, the overlapping bar charts provide the full error distributions underlying the action-probing analysis in the main paper, offering a more detailed view of how each representation distributes across low- and high-error regions.

We further provide qualitative results for the depth-probing experiment. As shown in Fig.~\ref{fig:depth_prob}, the depth predictions obtained from the predictive geometric representation more faithfully recover the spatial structure of the corresponding scenes than those obtained from the unified representation. In particular, the predicted maps better preserve object boundaries, relative depth relationships, and major scene structures across the future frames. These qualitative observations complement the quantitative probing results and provide additional evidence that the predictive geometric representation retains richer geometric information than the unified representation.

\begin{figure*}[p]
    \centering
    \includegraphics[width=\textwidth]{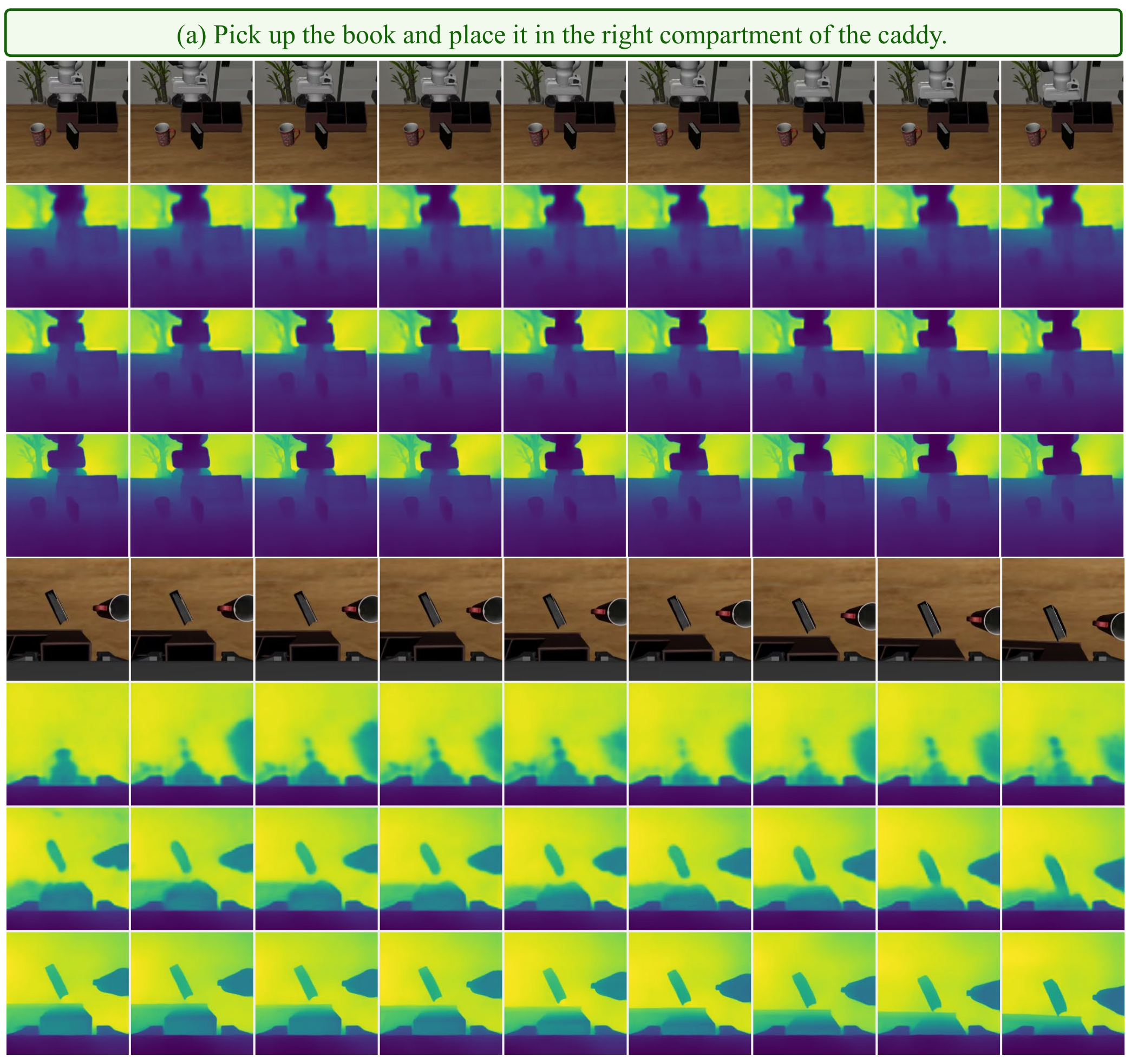}

    \captionsetup{labelformat=empty}
    \caption{}
\end{figure*}

\clearpage

\begin{figure*}[p]
    \ContinuedFloat
    \centering
    \includegraphics[width=\textwidth]{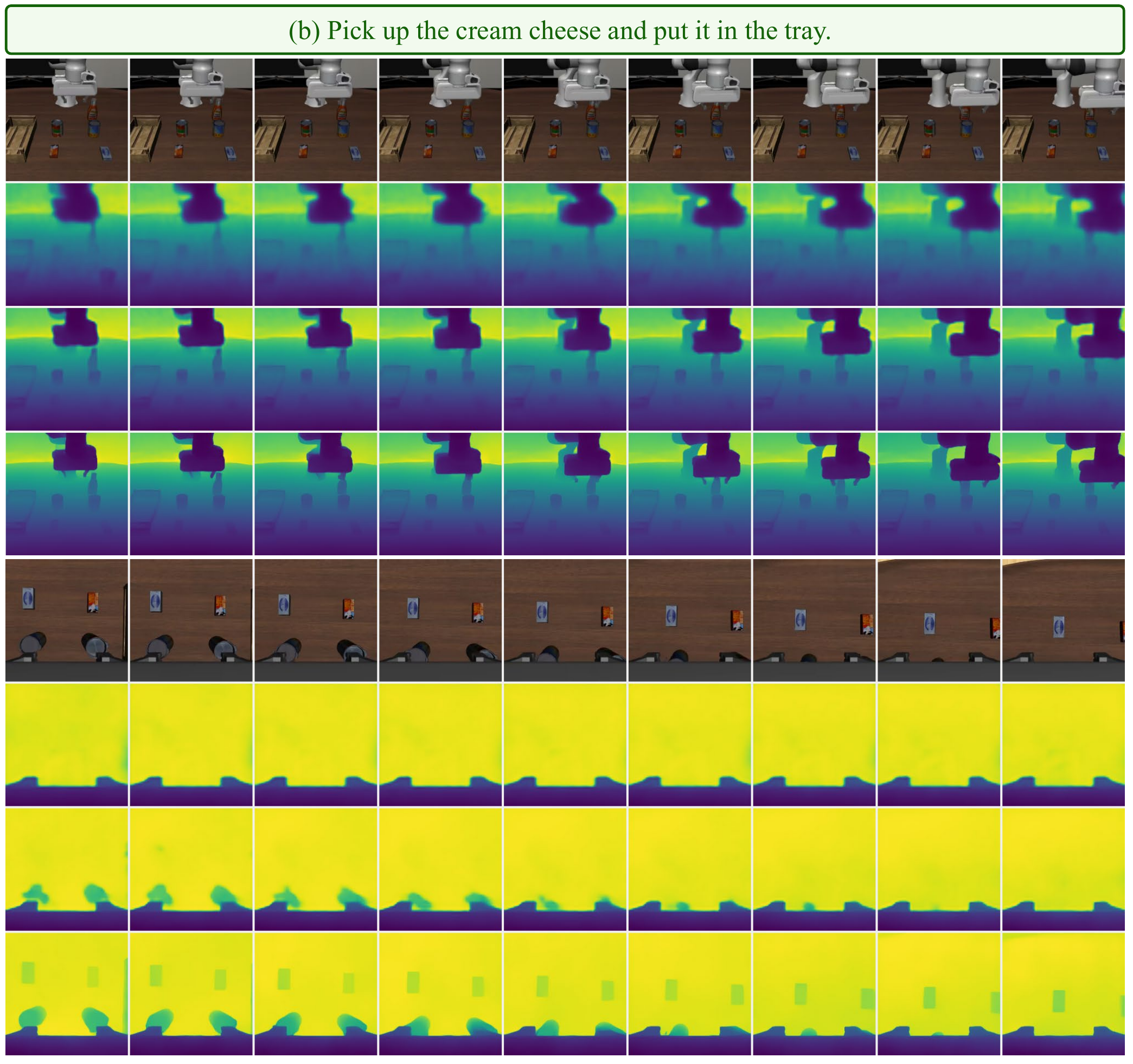}

    \caption{\small Qualitative visualization of the depth-probing analysis for representation organization. Each example contains a main camera view (upper half) and a wrist camera view (lower half), with columns progressing over nine future frames. Within each view, the rows from top to bottom show the corresponding RGB observations, depth predictions decoded from the unified representation, depth predictions decoded from the predictive geometric representation, and the pseudo-depth references extracted by Depth Anything~\citep{lin2025depth}. The predictive geometric representation produces depth maps that more faithfully recover object geometry, relative depth, and overall scene structure than the unified representation.}

    \label{fig:depth_prob}
\end{figure*}

\clearpage

\begin{table}[!t]
\centering
\setlength{\tabcolsep}{3pt}
\renewcommand{\arraystretch}{0.95}
\caption{More explore of representation organization on WidowX benchmark.}
    \label{tab:explore}
\resizebox{\columnwidth}{!}{
\begin{tabular}{cccccc}
\toprule
Methods
& \makecell{Put\\Spoon}
& \makecell{Put\\Carrot}
& \makecell{Stack\\Cube}
& \makecell{Put\\Eggplant}
& Avg. \\
\midrule
Dual-Unified
& \textbf{95.8} & 66.7 & 66.7 & 45.8 & 68.8 \\
Complementary
& \textbf{95.8} & \textbf{70.8} & \textbf{70.8} & \textbf{50.0} & \textbf{71.9} \\
\bottomrule
\end{tabular}
}
\end{table}

\begin{table}[!t]
    \centering
    \caption{Ablation on the effect of representation organization on LIBERO benchmark.}
    \label{tab:libero-org}
    \begin{tabular}{cccccc}
        \toprule
        Methods & Object & Spatial & Goal & Long & Avg. \\
        \midrule
        Unified & 99.0 & 97.8 & 97.0 & 96.8 & 97.6 \\
        Ours & \textbf{100.0} & \textbf{99.0} & 
        \textbf{98.2} & \textbf{97.0} & \textbf{98.6} \\
        \bottomrule
    \end{tabular}
\end{table}

\begin{table}[!t]
    \centering
    \caption{Ablation on the effect of representation organization on LIBERO-Plus benchmark.}
    \label{tab:libero-plus-org}
    \resizebox{\linewidth}{!}{
        \begin{tabular}{ccccccccc}
            \toprule
            Methods & Camera & Robot & Language & Light & Background & Noise & Layout & Total \\
            \midrule
            Unified & 69.8 & 56.6 & 85.9 & 84.2 & 88.3 & 89.5 & 74.4 & 77.7 \\
            Ours & \textbf{77.5} & \textbf{60.8} & 
            \textbf{88.6} & \textbf{86.1} & 
            \textbf{89.1} & \textbf{95.6} & 
            \textbf{77.2} & \textbf{81.7} \\
            \bottomrule
        \end{tabular}
    }
\end{table}

\noindent\textbf{Further Explore on Representation Organization in supervision pathways.} To disentangle representation organization from the number of supervision pathways, we conduct an additional controlled experiment, termed Dual-Unified, as shown in Tab.~\ref{tab:explore}. It retains Unified's pretraining setup but introduces two post-training supervision pathways: the shared representation supervises both the geometry-query pathway and the intermediate features of the action-generation pathway. Importantly, the representation targets remain identical to Unified. We can find that Dual-Unified still falls 3.1 percentage points short of CPR. This indicates that the complementary organization, which assigns distinct representations to preserve predictive geometric structure and capture action-aware transitions , provides additional benefits.

\noindent\textbf{Further Explore on Representation Organization with more benchmark.} 
To further investigate the complementary benefits of these representations, we extend the evaluation beyond the experiments on WidowX in main paper. We compare CPR with unified by training both models on LIBERO and evaluating their zero-shot generalization on LIBERO-Plus, as shown in Tab.\ref{tab:libero-org} and Tab.\ref{tab:libero-plus-org}. On LIBERO, CPR improves the average success rate from 97.6\% to 98.6\%. More importantly, on the more challenging LIBERO-Plus benchmark, CPR achieves an average success rate of 81.7\%, outperforming unified by 4.0 percentage points. In particular, CPR yields substantial improvements of 7.7 and 6.1 percentage points in camera and noise robustness, respectively. Moreover, CPR consistently outperforms unified across all seven evaluation dimensions, further demonstrating the generalization benefits of organizing predictive geometry and action-aware information into complementary representations.

\begin{table}[!t]
    \centering
    \caption{\small Ablation of motor representation construction on WidowX benchmark.}
    \label{tab:motor}

    \setlength{\tabcolsep}{3pt}
    \renewcommand{\arraystretch}{0.95}

    \resizebox{\columnwidth}{!}{
    \begin{tabular}{cccccc}
        \toprule
        Motor Input & \makecell{Put\\Spoon}
        & \makecell{Put\\Carrot}
        & \makecell{Stack\\Cube}
        & \makecell{Put\\Eggplant} & Avg. \\
        \midrule
        $\mathbf{G}$ & 87.5 & 66.7 & 66.7 & \textbf{54.2} & 68.8 \\
        $\mathbf{G}+\Delta\mathbf{G}$ & 87.5 & \textbf{70.8} & 66.7 & \textbf{54.2} & 69.8 \\
        $\Delta\mathbf{G}$  & \textbf{95.8} & \textbf{70.8}
        & \textbf{70.8} & 50.0 & \textbf{71.9} \\
        \bottomrule
    \end{tabular}
    }
\end{table}

\noindent\textbf{Effect of Geometric Transitions for Motor Representation Learning.}
We next examine why the motor representation is learned from geometric transitions rather than directly from predictive geometric states. We construct a variant in which the motion queries attend directly to $\mathbf{G}$ instead of $\Delta\mathbf{G}$ while keeping the remaining training procedure unchanged. As shown in Tab.~\ref{tab:motor}, directly using the predictive geometric states yields an average success rate of 68.8\%, whereas using geometric transitions increases the performance to 71.9\%, an absolute gain of 3.1 percentage points. This result supports the use of transitions as the input to motor representation learning. A possible interpretation is that consecutive predictive states share substantial information describing scene structure, whereas their difference emphasizes temporal variation. Using $\Delta\mathbf{G}$ therefore provides a more action-oriented signal from which action supervision can identify information associated with motor commands.

We additionally evaluate a $\mathbf{G} + \Delta\mathbf{G}$ variant that incorporates both absolute geometric states and transitions into motor representation learning. Incorporating both $\mathbf{G}$ and $\Delta\mathbf{G}$ improves over $\mathbf{G}$ alone, but remains below our transition-based design. This supports explicitly emphasizing geometric transitions when constructing MR, rather than directly combining absolute states and transitions within the motor representation pathway.
Importantly, MR is not intended to independently encode the complete geometric context required for action generation. CPR preserves absolute geometric information through PGR, which supervises dedicated geometry queries, while MR focuses on action-relevant geometric transitions and aligns with intermediate features along the action-generation pathway. Thus, the two representations provide complementary information through separate transfer pathways, retaining both scene geometry and temporal changes.

\begin{table}[!t]
  \centering
  \caption{\small Effect of the foundation encoder used during predictive representation pretraining 
on the WidowX evaluation.}
  \label{tab:dino_vggt_performance}
  \begin{tabular}{cccccc}
    \toprule
    Methods & \makecell{Put\\Spoon} 
            & \makecell{Put\\Carrot} 
            & \makecell{Stack\\Cube} 
            & \makecell{Put\\Eggplant} 
            & Avg. \\
    \midrule
    DINO & 87.5          & \textbf{75.0} & 54.2          & \textbf{62.5} & 69.8 \\
    VGGT & \textbf{95.8} & 70.8          & \textbf{70.8} & 50.0          & \textbf{71.9} \\
    \bottomrule
  \end{tabular}
\end{table}

\noindent\textbf{Effect of the Foundation Encoder.}
We further examine the dependence of representation pretraining on the chosen foundation encoder. We compare DINO\citep{oquab2023dinov2}, which provides generic visual representations, with VGGT\citep{wang2025vggt}, which provides stronger geometric structure. As shown in Tab.~\ref{tab:dino_vggt_performance}, both encoders lead to competitive downstream performance, achieving 69.8\% and 71.9\% average success rates, respectively. VGGT provides the stronger overall result and a notable advantage on \textit{Stack Cube}, which requires reasoning about relative object geometry. At the same time, DINO remains competitive and performs better on \textit{Put Carrot} and \textit{Put Eggplant}. These results suggest that the proposed representation learning framework can utilize different pretrained feature spaces while benefiting from a foundation model with stronger geometric structure.

\noindent\textbf{Further Ablation on the Effect of the Foundation Encoder.}
To further investigate this contribution, we conduct additional real-world evaluations comparing DINO-based and VGGT-based CPR under identical downstream training and evaluation protocols, as shown in Tab.\ref{tab:vggt_real}. Notably, the first three real-world tasks in Appendix.\ref{task_suits} involve substantial geometric perception and spatial reasoning, providing challenging physical settings to examine the practical benefits of explicit geometric priors. As shown in Tab.\ref{tab:vggt_real}, VGGT-based CPR consistently outperforms DINO-based CPR across all four tasks. Notably, VGGT-based CPR improves the average real-world success rate from 53.3\% to 66.7\%. The largest gain occurs on Transfer Liquid, which requires precise spatial alignment among multiple objects and tools, together with coordinated manipulation throughout the transfer process. The improvements on Remove Stopper and Store Tubes further support the practical benefits of explicit geometric information in spatially demanding manipulation tasks.

\begin{table}[!t]
    \centering
    \caption{Ablation of foundation encoder on real-world tasks.}
    \label{tab:vggt_real}
    \begin{tabular}{cccccc}
        \toprule
        Methods &  \makecell{Remove\\ Stopper} 
            & \makecell{Store\\Tubes} 
            & \makecell{Transfer\\Liquid} 
            & \makecell{Make\\Burger}
         & Avg. \\
        \midrule
        DINO & 48.3 & 60.0 & 15.0 & 90.0 & 53.3 \\
        VGGT & \textbf{60.0} & \textbf{66.7} & 
        \textbf{46.7} & \textbf{93.3} & \textbf{66.7} \\
        \bottomrule
    \end{tabular}
\end{table}

\begin{table}[!t]
\centering
\caption{\small Effect of the motor representation alignment depth on the WidowX evaluation. The alignment is applied to intermediate features from different layers.}
\label{tab:layer_performance}

\setlength{\tabcolsep}{4pt}
\renewcommand{\arraystretch}{1.0}

\resizebox{0.85\columnwidth}{!}{
\begin{tabular}{ccccccc}
\toprule
Alignment Layer & 6 & 12 & 18 & 24 & 30 & 36 \\
\midrule
SR (\%) & 69.8 & \textbf{71.9} & 67.7 & 64.6 & 65.6 & 61.5 \\
\bottomrule
\end{tabular}
}
\end{table}

\noindent\textbf{Effect of the Motor Representation Alignment Layer.}
We investigate where motor guidance should be introduced within the downstream VLA backbone. Specifically, we apply the same motor representation alignment objective to intermediate features from different transformer layers while keeping the remaining training configuration unchanged. As reported in Tab.~\ref{tab:layer_performance}, performance reaches its maximum at layer 12, with an average success rate of 71.9\%. The performance decreases when the alignment point is moved toward deeper layers. This suggests that motor guidance is most effective at an intermediate stage of the policy representation, where the features have incorporated sufficient contextual information while retaining flexibility to influence subsequent action generation.

\noindent\textbf{Qualitative Analysis of Predictive Geometry.}
\label{sec:qualitative_predictive_geometry}
We further qualitatively examine whether the geometry-query pathway transfers meaningful predictive geometric information into the downstream policy. Following the protocol described in the main paper, we freeze the post-trained policy and extract its predicted geometry-query features from current-time inputs. A lightweight DPT-based depth probe~\citep{ranftl2021vision} is then trained on these frozen features, while keeping the VLA backbone and the geometry-query representations fixed. Since ground-truth depth annotations are unavailable for the evaluated observations, we use pseudo-depth maps generated by Depth Anything~\citep{lin2025depth} as supervision for the probe.

As illustrated in Fig.~\ref{fig:real_depth_prob}, we provide qualitative results on both LIBERO and real-world manipulation scenes. For the LIBERO examples, the geometry-query features recover major spatial structures across both the main and wrist views, including object boundaries, relative depth relationships, and scene layout. Similar behavior is observed in the real-world examples across the main view and the two wrist views, suggesting that the learned geometry-query pathway preserves meaningful spatial structure under different embodiments and visual configurations. Importantly, these geometric features are inferred solely from current-time inputs, while their predictive supervision during post-training is constructed from future-informed representations. The qualitative results therefore provide evidence that the geometry-query pathway transfers predictive geometric information into the downstream policy, rather than merely encoding the current visual appearance.


\begin{figure*}[p]
    \centering
    \includegraphics[width=\textwidth]{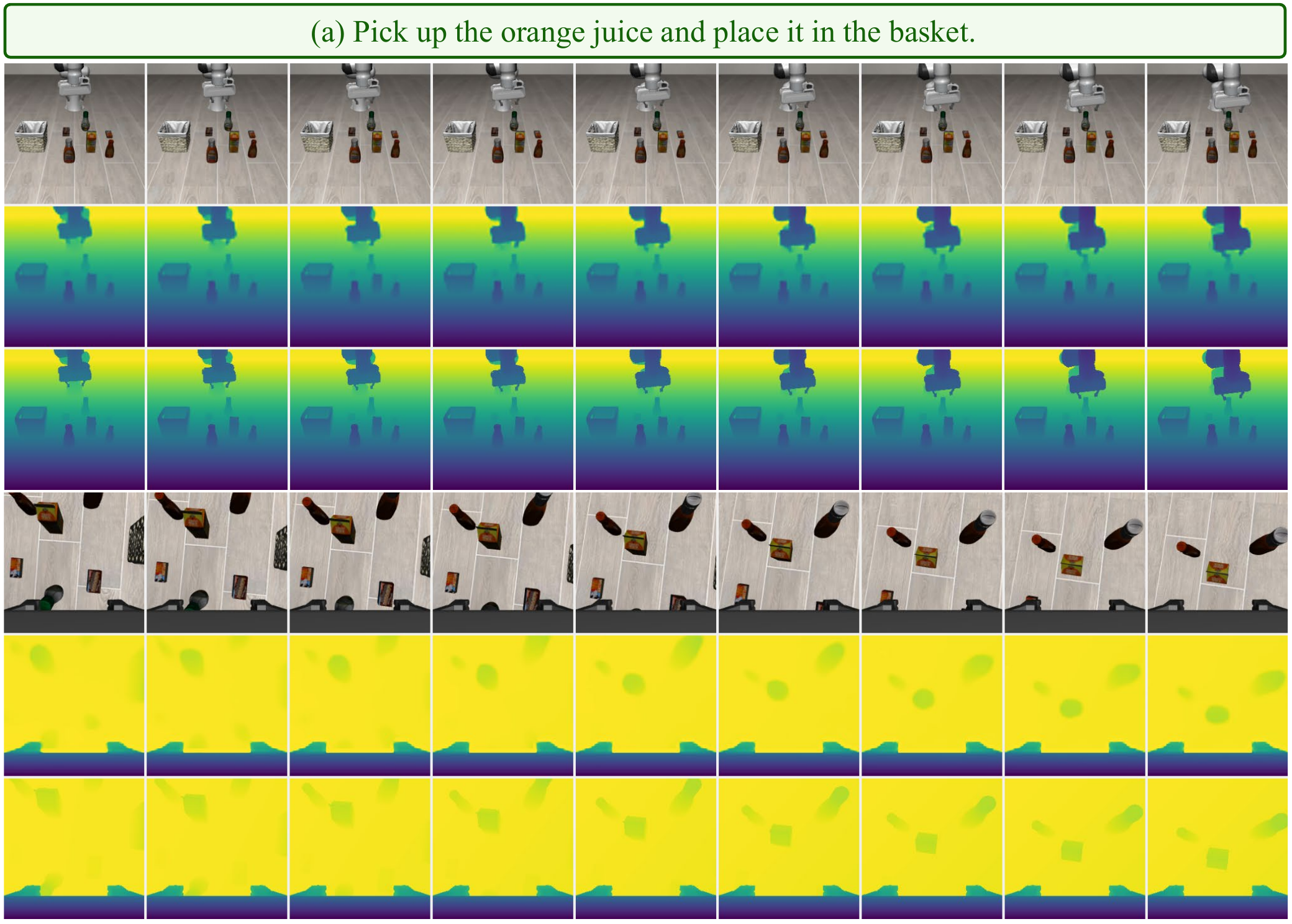}

    \captionsetup{labelformat=empty}
    \caption{}
\end{figure*}

\clearpage

\begin{figure*}[p]
    \ContinuedFloat
    \centering
    \includegraphics[width=\textwidth]{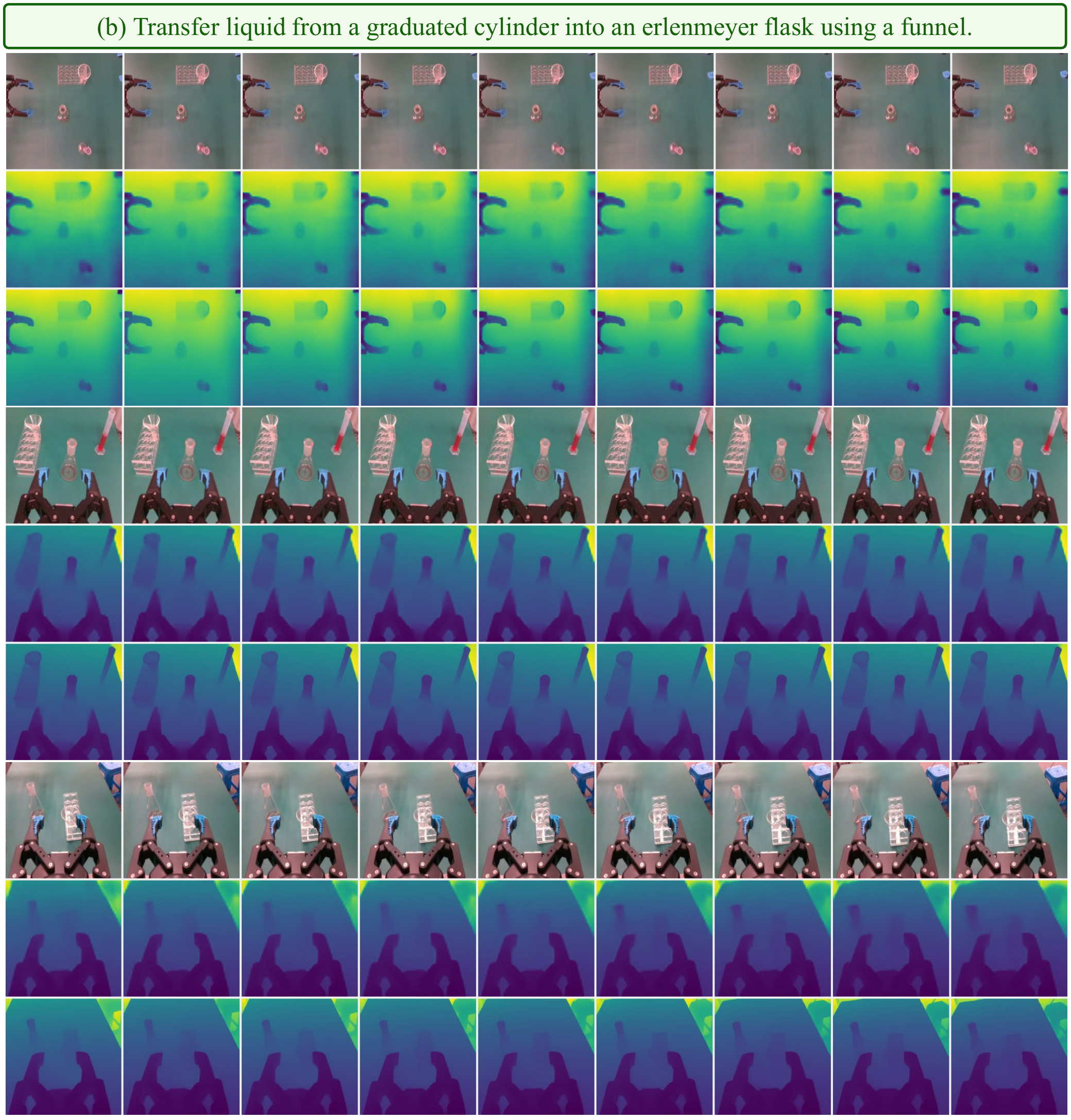}

    \caption{\small Qualitative visualization of predictive geometry transferred through the geometry-query pathway. The first panel shows examples from LIBERO, where each example contains a main camera view (upper half) and a wrist-camera view (lower half); within each view, the rows from top to bottom show the RGB observation, the depth prediction decoded from the predicted geometry-query features, and the pseudo-depth reference generated by Depth Anything. The second panel shows real-world manipulation examples with three camera views, corresponding to the main view, the left-arm wrist view, and the right-arm wrist view. For each view, the rows from top to bottom follow the same order: RGB observation, predicted depth, and pseudo-depth reference. The results illustrate that the geometry-query pathway can recover meaningful object geometry and relative spatial structure across both simulated and real-world settings.}

    \label{fig:real_depth_prob}
\end{figure*}

\clearpage

\begin{table}[!t]
\centering
\caption{\small Controlled comparison of different geometry-guidance paradigms under the same downstream VLA framework on the WidowX evaluation. The input-centric variant uses current-frame VGGT features as policy input, while the prediction-centric variant predicts future VGGT features as auxiliary guidance.}
\label{tab:guidance_paradigm_comparison}

\setlength{\tabcolsep}{3pt}
\renewcommand{\arraystretch}{0.95}

\resizebox{\columnwidth}{!}{
\begin{tabular}{cccccc}
\toprule
Guidance Paradigm
& \makecell{Put\\Spoon}
& \makecell{Put\\Carrot}
& \makecell{Stack\\Cube}
& \makecell{Put\\Eggplant}
& Avg. \\
\midrule
Input-centric
& 83.3 & 66.7 & 58.3 & 45.8 & 63.5 \\
Prediction-centric
& 87.5 & 62.5 & 62.5 & \textbf{50.0} & 65.6 \\
Ours
& \textbf{95.8} & \textbf{70.8} & \textbf{70.8} & \textbf{50.0} & \textbf{71.9} \\
\bottomrule
\end{tabular}
}
\end{table}

\noindent\textbf{Comparison of Guidance Paradigms.}
To further examine the design of CPR, we compare our approach with two representative guidance paradigms under the same downstream VLA framework and training setting. The \emph{input-centric} variant directly provides current-frame VGGT features to the policy, allowing it to access explicit geometric information from the observed scene. The \emph{prediction-centric} variant instead uses future VGGT features as a predictive target for guidance. In contrast, our method learns a predictive geometric representation from dense geometric features and further extracts action-aware information from geometric transitions. For a controlled comparison, all variants use the same pretraining data scale and number of optimization steps as CPR, followed by an identical downstream post-training procedure.

As shown in Tab.~\ref{tab:guidance_paradigm_comparison}, the input-centric and prediction-centric variants achieve average success rates of 63.5\% and 65.6\%, respectively, while CPR reaches 71.9\%. The consistent improvement suggests that explicitly incorporating current geometry or future geometric prediction alone does not fully capture the complementary information needed for action generation. By organizing predictive geometry together with action-aware information extracted from geometric transitions, CPR provides a more effective interface for downstream policy learning.

\noindent\textbf{Ablation on VGGT Encoding Strategies.}
We compare independent frame-view processing with a Joint-VGGT variant that jointly processes multiple views during geometric feature extraction. The WidowX results are as shown in Tab.\ref{tab:vggt}. Independent processing achieves 71.9\% average success, outperforming joint VGGT processing by 4.2 percentage points. It performs equally well or better across all four tasks, with improvements on three. This design is motivated by CPR's hierarchical representation organization. Rather than performing early joint aggregation within the geometric foundation model, we independently extract geometric features from each frame-view pair and subsequently organize them through our hierarchical view-then-time tokenizer. This explicitly separates view-level aggregation from temporal modeling, while still integrating multi-view geometric information during representation learning.

\begin{table}[!t]
    \centering
    \caption{\small Ablation of the style of VGGT encoding on WidowX benchmark.}
    \label{tab:vggt}

    \setlength{\tabcolsep}{3pt}
    \renewcommand{\arraystretch}{0.95}

    \resizebox{\columnwidth}{!}{
    \begin{tabular}{cccccc}
        \toprule
        Methods & \makecell{Put\\Spoon}
        & \makecell{Put\\Carrot}
        & \makecell{Stack\\Cube}
        & \makecell{Put\\Eggplant} & Avg. \\
        \midrule
        Joint-VGGT & 87.5 & 66.7 & \textbf{70.8} & 45.8 & 67.7 \\
        Ours & \textbf{95.8} & \textbf{70.8}
        & \textbf{70.8} & \textbf{50.0} & \textbf{71.9} \\
        \bottomrule
    \end{tabular}
    }
\end{table}

\noindent\textbf{Effect of Hierarchical View-then-Time Tokenization.}
We compare our hierarchical tokenizer with Flat Joint Tokenization, as shown in Tab.\ref{tab:hv}, which jointly compresses multi-view and temporal features. Our design achieves 71.9\% average WidowX success, compared with 69.8\% for flat tokenization. We further evaluate both variants through zero-shot generation using FastWAM on LIBERO-90, as shown in Tab.\ref{tab:libero-hv}. Our method improves PSNR from 21.08 to 22.91 and SSIM from 0.795 to 0.803, while reducing LPIPS from 0.193 to 0.186 and action $\ell_1$ error from 0.091 to 0.075. These results support explicitly separating view-level and temporal aggregation.
\begin{table}[!t]
    \centering
    \caption{Ablation on the hierarchical view-then-time tokenization on WidowX benchmark.}
    \label{tab:hv}
    \begin{tabular}{cccccc}
        \toprule
        Methods & \makecell{Put\\Spoon} 
            & \makecell{Put\\Carrot} 
            & \makecell{Stack\\Cube} 
            & \makecell{Put\\Eggplant} & Avg. \\
        \midrule
        Flat Joint & 87.5 & 66.7 & \textbf{70.8} & \textbf{54.3} & 69.8 \\
        Ours & \textbf{95.8} & \textbf{70.8} & \textbf{70.8} & 50.0 & \textbf{71.9} \\
        \bottomrule
    \end{tabular}
\end{table}

\begin{table}[!t]
    \centering
    \caption{Ablation on the hierarchical view-then-time tokenization on LIBERO-90 benchmark.}
    \label{tab:libero-hv}
    \begin{tabular}{ccccc}
        \toprule
        Methods & PSNR $\uparrow$ & SSIM $\uparrow$ & LPIPS $\downarrow$ & Action $\ell_1\downarrow$ \\
        \midrule
        Flat Joint & 21.08 & 0.795 & 0.193 & 0.091 \\
        Ours & \textbf{22.91} & \textbf{0.803} & \textbf{0.186} & \textbf{0.075} \\
        \bottomrule
    \end{tabular}
\end{table}

\noindent\textbf{Effect of Temporal Grouping.}
We investigate temporal grouping granularity by comparing group sizes of 2, 4, and 8 while keeping the anchor-frame configuration fixed, as shown in Tab.\ref{tab:group}. Although a group size of 2 achieves a slightly higher average success rate, we adopt the 4-frame configuration as it offers a favorable trade-off between performance and temporal grouping granularity, with per-step execution times of 0.36 s and 0.41 s for group sizes of 2 and 4, respectively. Larger groups (8) lead to noticeable performance degradation.
\begin{table}[!t]
    \centering
    \caption{\small Ablation on the temporal grouping on WidowX benchmark.}
    \label{tab:group}

    \setlength{\tabcolsep}{3pt}
    \renewcommand{\arraystretch}{0.95}

    \resizebox{\columnwidth}{!}{
    \begin{tabular}{cccccc}
        \toprule
        Group Size & \makecell{Put\\Spoon}
        & \makecell{Put\\Carrot}
        & \makecell{Stack\\Cube}
        & \makecell{Put\\Eggplant} & Avg. \\
        \midrule
        2 & 91.7 & \textbf{75.0} & \textbf{70.8} & \textbf{54.2} & \textbf{72.9} \\
        4 & \textbf{95.8} & 70.8 & \textbf{70.8} & 50.0 & 71.9 \\
        8 & 91.7 & 66.7 & \textbf{70.8} & 50.0 & 69.8 \\
        \bottomrule
    \end{tabular}
    }
\end{table}

\noindent\textbf{Effect of Causal Masking in Slot-Temporal Mixer.}
To examine causal masking, we replace causal attention with bidirectional attention while keeping the remaining architecture unchanged and conduct the experiments in WidowX and LBERO-90, as shown in Tab.\ref{tab:causal} and Tab.\ref{tab:causal_90}. Causal attention improves average WidowX success from 68.8\% to 71.9\%. Consistently, FastWAM's zero-shot generation improves PSNR from 20.95 to 22.91 and SSIM from 0.787 to 0.803, while reducing LPIPS from 0.201 to 0.186 and action $\ell_1$ error from 0.096 to 0.075. These results support causal temporal modeling for constructing history-conditioned predictive geometric states.

\begin{table}[!t]
    \centering
    \caption{Ablation of causal Slot-Temporal Mixer on WidowX benchmark.}
    \label{tab:causal}
    \resizebox{0.95\linewidth}{!}{
    \begin{tabular}{cccccc}
        \toprule
        Attention & \makecell{Put\\Spoon}
        & \makecell{Put\\Carrot}
        & \makecell{Stack\\Cube}
        & \makecell{Put\\Eggplant} & Avg. \\
        \midrule
        Bidirectional & 91.7 & \textbf{70.8} & 66.7 & 45.8 & 68.8 \\
        Causal & \textbf{95.8} & \textbf{70.8}
        & \textbf{70.8} & \textbf{50.0} & \textbf{71.9} \\
        \bottomrule
    \end{tabular}
    }
\end{table}

\begin{table}[!t]
    \centering
    \caption{\small Ablation of causal Slot-Temporal Mixer on LIBERO-90 benchmark.}
    \label{tab:causal_90}

    \setlength{\tabcolsep}{3pt}
    \renewcommand{\arraystretch}{0.9}

    \resizebox{0.95\columnwidth}{!}{
    \scriptsize
    \begin{tabular}{ccccc}
        \toprule
        Attention & PSNR $\uparrow$ & SSIM $\uparrow$
        & LPIPS $\downarrow$ & \makecell{Action\\$\ell_1\downarrow$} \\
        \midrule
        Bidirectional & 20.95 & 0.787 & 0.201 & 0.096 \\
        Causal & \textbf{22.91} & \textbf{0.803}
        & \textbf{0.186} & \textbf{0.075} \\
        \bottomrule
    \end{tabular}
    }
\end{table}

\noindent\textbf{Effect of Progressive Training Curriculum.}
We compare our four-stage curriculum with joint multi-objective training, where all pretraining objectives are introduced simultaneously. As shown in Tab.\ref{tab:progressive_90}, progressive training achieves 71.9\% average WidowX success, compared with 69.8\% for joint training. This supports progressively introducing geometric reconstruction, masked prediction, transition prediction, and motor supervision rather than optimizing all objectives jointly from the beginning.

\begin{table}[!t]
    \centering
    \caption{\small Ablation of progressive training curriculum on WidowX benchmark.}
    \label{tab:progressive_90}

    \setlength{\tabcolsep}{3pt}
    \renewcommand{\arraystretch}{0.95}

    \resizebox{\columnwidth}{!}{
    \begin{tabular}{cccccc}
        \toprule
        Training Strategy & \makecell{Put\\Spoon}
        & \makecell{Put\\Carrot}
        & \makecell{Stack\\Cube}
        & \makecell{Put\\Eggplant} & Avg. \\
        \midrule
        Joint Training & 91.7 & \textbf{70.8} & 66.7 & \textbf{50.0} & 69.8 \\
        Progressive & \textbf{95.8} & \textbf{70.8}
        & \textbf{70.8} & \textbf{50.0} & \textbf{71.9} \\
        \bottomrule
    \end{tabular}
    }
\end{table}

\appsection{Training Details}
\label{training_details}
\appsubsection{Datasets}
For predictive representation pretraining, we construct a mixed training corpus that combines robot manipulation datasets with additional general-purpose data. The robot manipulation data are designed to provide diverse interaction trajectories with explicit visuomotor structure and action supervision. We organize these data into two broad groups. The first group consists primarily of conventional robot manipulation datasets, including Open X-Embodiment (OXE)\citep{open_x_embodiment}, RoboTwin\citep{chen2025robotwin}, and LIBERO(exclude LIBERO-90)\citep{Libero}. These datasets provide diverse robotic interaction trajectories and action-labeled manipulation behaviors, which expose the representation model to a broad range of object interactions, robot motions, and action-conditioned scene changes. The second group consists of ego data, which provide substantially broader interaction diversity and complementary visuomotor observations beyond conventional fixed-base robot manipulation datasets\citep{jin2026umi,grauman2022ego4d}. In addition, we incorporate several general-purpose datasets\citep{goyal2017something} to further increase the diversity of visual scenes and improve the model's ability to capture geometric structure across a broader range of environments. We use the mixed corpus throughout representation pretraining rather than restricting individual training stages to a specific data source. This combination provides both manipulation-oriented action supervision and diverse visual-geometric observations, allowing the predictive representations to learn scene structure and its evolution from a broader distribution of interaction data.

\begin{table}[htbp]
\centering
\caption{Our training data mixture used during the pretraining phase.}
\begin{tabular}{cc}
\hline
Dataset & Ratio \\
\hline
UTAustin Mutex \citep{shah2023mutex} & 2.3 \\
DLR EDAN Shared Control \citep{quere2020shared} & 0.7 \\
Taco Play \citep{rosete2023latent,mees2022grounding} & 2.3 \\
Austin Sirius Dataset \citep{liu2025robot} & 1.5 \\
Roboturk \citep{mandlekar2019scaling} & 1.6 \\
Austin Sailor Dataset \citep{nasiriany2022learning} & 1.7 \\
Berkeley Fanuc Manipulation \citep{zhu2023fanuc} & 1.9 \\
UCSD Kitchen Dataset \citep{yan2023ucsd} & 0.6 \\
LIBERO \citep{Libero} & 3.4 \\
Viola \citep{zhu2023viola} & 0.8 \\
Austin Buds Dataset \citep{zhu2022bottom} & 0.2 \\
Bridge \citep{walke2023bridgedata,ebert2021bridge} & 11.6 \\
FMB Dataset \citep{luo2025fmb} & 1.3 \\
Furniture Bench Dataset \citep{heo2025furniturebench} & 2.3 \\
NYU Franka Play Dataset \citep{cui2022play}& 1.8 \\
Robotwin \citep{chen2025robotwin} & 7.3 \\
Berkeley Cable Routing \citep{luo2024multistage} & 1.0 \\
BC-Z \citep{jang2022bc} & 5.8 \\
Berkeley Autolab UR5 \citep{chen2023berkeley}& 2.3 \\
Stanford Hydra Dataset \citep{belkhale2023hydra} & 4.3 \\
Fractal \citep{brohan2022rt} & 10.7 \\
Jaco Play \citep{dassclvr} & 2.1 \\
DobbE \citep{shafiullah2023bringing} & 0.5 \\
Ego4D \citep{grauman2022ego4d} & 12.3 \\
UMI \citep{jin2026umi} & 10.8 \\
Something-Something V2 \citep{goyal2017something} & 8.9 \\
\hline
\end{tabular}
\label{tab:dataset_ratios}
\end{table}


\appsubsection{Training Protocol}
\noindent\textbf{Pretraining Configurations.}
We train the predictive representation model using a progressive multi-stage training protocol. Rather than introducing all objectives simultaneously, we gradually add the supervision objectives according to the role they play in shaping the complementary representations. This curriculum first establishes structurally informative geometric representations, then encourages temporal contextual modeling and transition prediction, and finally introduces action supervision to extract the motor representation.

\textbf{1) Stage 1: Geometry Reconstruction.}
We first train the representation model using only the geometry reconstruction objective $\mathcal{L}_{\mathrm{rec}}$. At this stage, the mixed robot manipulation and general-purpose data are used to establish an initial predictive geometric representation. Since the representation is obtained through hierarchical tokenization and temporal compression, the reconstruction objective provides a direct signal for preserving the geometric information contained in the pretrained foundation features.

\textbf{2) Stage 2: Masked Geometric Representation Prediction.}
After the representation acquires basic geometric structure, we introduce the masked geometric state prediction objective $\mathcal{L}_{\mathrm{mask}}$. Masked temporal slots are inferred from the available contextual states, encouraging the model to capture dependencies across the interaction sequence. This stage improves the ability of the predictive geometric representation to infer missing geometric information from temporal context.

\textbf{3) Stage 3: Geometric Transition Prediction.}
We next introduce the geometric transition prediction objective $\mathcal{L}_{\mathrm{trans}}$. In this stage, the representation model is explicitly trained to predict changes between consecutive predictive geometric states. This encourages the learned representation not only to describe scene structure but also to capture how the geometric configuration evolves over time.

\textbf{4) Stage 4: Motor Representation Learning.}
Finally, we introduce the action prediction objective $\mathcal{L}_{\mathrm{act}}^{\mathrm{pre}}$ using conventional robot manipulation data that provide temporally aligned action supervision. Motion queries attend to the geometric transitions $\Delta\mathbf{G}$, and the resulting motor representation is supervised with the corresponding action chunks. This final stage extracts action-aware information from the learned geometric transitions and completes the learning of the complementary predictive geometric and motor representations.

Unless otherwise specified, each stage is initialized from the parameters obtained in the preceding stage. All pretraining datasets have action dimensions $d_i\leq 32$. We therefore adopt a unified 32-dimensional action target, zero-padding dimensions when $d_i<32$. A dimension-wise mask excludes padded dimensions from the action regression loss. We use a four-stage progressive training schedule with a total of 120K optimization steps, which is consist of Geometry Reconstruction(0-25k),  Masked Geometric Representation Prediction(25k-50k),  Geometric Transition Prediction(50k-70k), Motor Representation Learning(70k-120k). We preserve each dataset's native action sampling frequency without resampling trajectories to a common frequency. For visual observations, we sample one frame every two native dataset frames. We set $\lambda_{\mathrm{mask}}=0.5$, $\lambda_{\mathrm{trans}}=0.5$, and $\lambda_{\mathrm{vc}}=0.06$. This progressive training strategy allows the model to first establish a structurally informative geometric representation before introducing increasingly task-oriented temporal and action supervision. We perform representation pretraining on 32 NVIDIA H200 GPUs with a global batch size of 512. The base learning rate is set to $1\times10^{-5}$ with learning-rate decay, and the model is trained for a total of 120K optimization steps. The pretrained 3D foundation encoder remains frozen throughout representation pretraining. Furthermore, for both the aforementioned experiments and the fairness comparisons with our method presented in the main paper and above, we adopt an identical pretraining setup to ensure fair comparisons, unless otherwise specified. This includes using the same pretraining data, number of training steps, and other relevant pretraining configurations.

\noindent\textbf{Post-training Configurations.}
For downstream post-training, the pretrained predictive representation model is kept frozen. For each training clip, it is used to construct the two privileged targets: the predictive geometric representation $\mathbf{G}$ and the motor representation $\mathbf{Z}$. We initialize the downstream VLA visual-language backbone from Qwen3-VL-4B-Instruct~\citep{yang2025qwen3}. The native action-generation objective of the corresponding VLA is retained throughout post-training, while the representation-guidance objectives are jointly optimized with it. We use a linear warm-up schedule for the first 5,000 optimization steps. To facilitate stable adaptation, we further apply a cosine decay schedule to the coefficient of the latent motor-alignment loss, $\lambda_{\mathrm{align}}$, gradually reducing the strength of latent alignment supervision as training progresses. Importantly, the pretrained representation model is not used by the downstream policy at inference time. After post-training, the policy operates solely on its standard current-time inputs, with the predictive and motor representations serving only as privileged supervision during training.


\appsection{Model Details}
\appsubsection{Hierarchical Geometric Tokenization}
We use a hierarchical tokenization process to convert dense multi-view video features extracted by VGGT \citep{wang2025vggt} into a compact and temporally organized geometric token sequence. Given a multi-frame, multi-view input clip, each RGB image is first resized and padded following the input preprocessing procedure of VGGT, resulting in a resolution of $518\times518$. The processed images are then independently fed into a frozen VGGT encoder to obtain dense geometric features for each frame-view pair. Since the number of available camera views varies across the pretraining datasets, we use a maximum of three input views at each time step. Samples with fewer than three views are zero-padded to the maximum view number, and the corresponding padded views are masked throughout the subsequent attention operations. After view-level compression, we further aggregate frame-level geometric states along the temporal dimension. Inspired by hierarchical temporal compression designs \citep{wan2025wan}, we retain the first frame $\mathbf{H}_0$ as an initial-observation anchor and group every four subsequent frames into a temporal slot. Specifically, for each temporal group, a set of learnable temporal queries performs cross-attention over the frame-level geometric features within the corresponding group. The frame features serve as keys and values, while the learnable temporal queries serve as queries. Learnable frame-position embeddings are added before temporal aggregation to preserve the relative temporal ordering of frames within each group. We further construct the temporal attention mask to match the grouping structure of the temporal slots. Frame features belonging to the same temporal group are mutually visible to the corresponding temporal queries, whereas features from different groups are prevented from attending to each other during this local aggregation stage. Within each temporal group, padded frame positions are additionally masked so that invalid frames do not affect the resulting representation. The first frame is treated as a dedicated temporal anchor and is retained without being merged with subsequent frames. 

Through this hierarchical view- and time-wise compression, dense multi-view VGGT features are progressively transformed into a compact tokenized geometric sequence $S$, where the first temporal slot corresponds to the initial observation and each subsequent slot summarizes a group of four consecutive frames. This hierarchical design substantially reduces the number of geometric tokens while preserving both view-specific information and temporal ordering. More importantly, the resulting sequence explicitly organizes geometric information at two levels: view aggregation captures a compact representation of the scene from multiple camera perspectives, while temporal aggregation provides a structured representation for modeling the evolution of scene geometry over the interaction sequence.


\appsubsection{Slot-Temporal Mixer}
Given geometric states $G\in\mathbb{R}^{B\times K\times S\times D}$, where $K=3$, $S=98$, and $D=512$, we reshape the input to $\mathbb{R}^{(B\cdot S)\times K\times D}$, treating each geometric slot as an independent temporal sequence. The mixer comprises four causal Transformer layers with 8 attention heads, a hidden dimension of 512, and an FFN dimension of 2048. Each layer adopts Pre-Norm, residual connections, and zero dropout. Causal self-attention operates exclusively along the temporal slot dimension, enabling each slot to aggregate information from current and preceding chunks without additional slot-wise attention. Learnable slot positional embeddings are added before temporal mixing.

\appsubsection{Masked Geometric Representation Prediction}
We adopt slot-level masking, replacing each masked slot with a learnable mask-query embedding. The target and context branches share the same Slot-Temporal Mixer parameters: the target branch encodes the complete sequence to produce detached target representations, while the context branch processes masked inputs. Before prediction, masked context features are explicitly overwritten with learnable mask-query embeddings to prevent information leakage from potentially reconstructed features.
The masked predictor comprises two bidirectional Pre-Norm Transformer layers with 8 attention heads, a hidden dimension of 512, an FFN dimension of 2048, and zero dropout. It uses independent learnable slot positional embeddings, separate from those of the Slot-Temporal Mixer. A prediction head (LayerNorm → Linear(512, 512)) projects the outputs into the target representation space. Smooth L1 loss is applied exclusively at masked slot positions against detached targets, establishing a slot-level predictive learning objective rather than individual-slot or pixel-level reconstruction.

\appsubsection{Geometric Transition Prediction}
An auxiliary geometric transition prediction module consisting of LayerNorm → Linear(512, 512) → GELU → Linear(512, 512). Given current geometric states $G_k$, it predicts $\Delta\mathbf{G_k}=\mathbf{G_{k+1}}-\mathbf{G_k}$, supervised by detached next-slot targets using Smooth L1 loss.

\appsubsection{Details of FastWAM}
\label{details_fast}
FastWAM \citep{yuan2026fastwam} differs from conventional VLA models \citep{bjorck2025gr00t,openvla-oft,black2024pi_0}, in that it jointly models future visual observations and robot actions through a generative world-action architecture. We therefore introduce additional interfaces to incorporate the two complementary predictive representations into the original FastWAM architecture while preserving its native world- and action-generation mechanisms. 

\begin{figure}[htbp]
    \centering
    \includegraphics[width=0.8\linewidth]{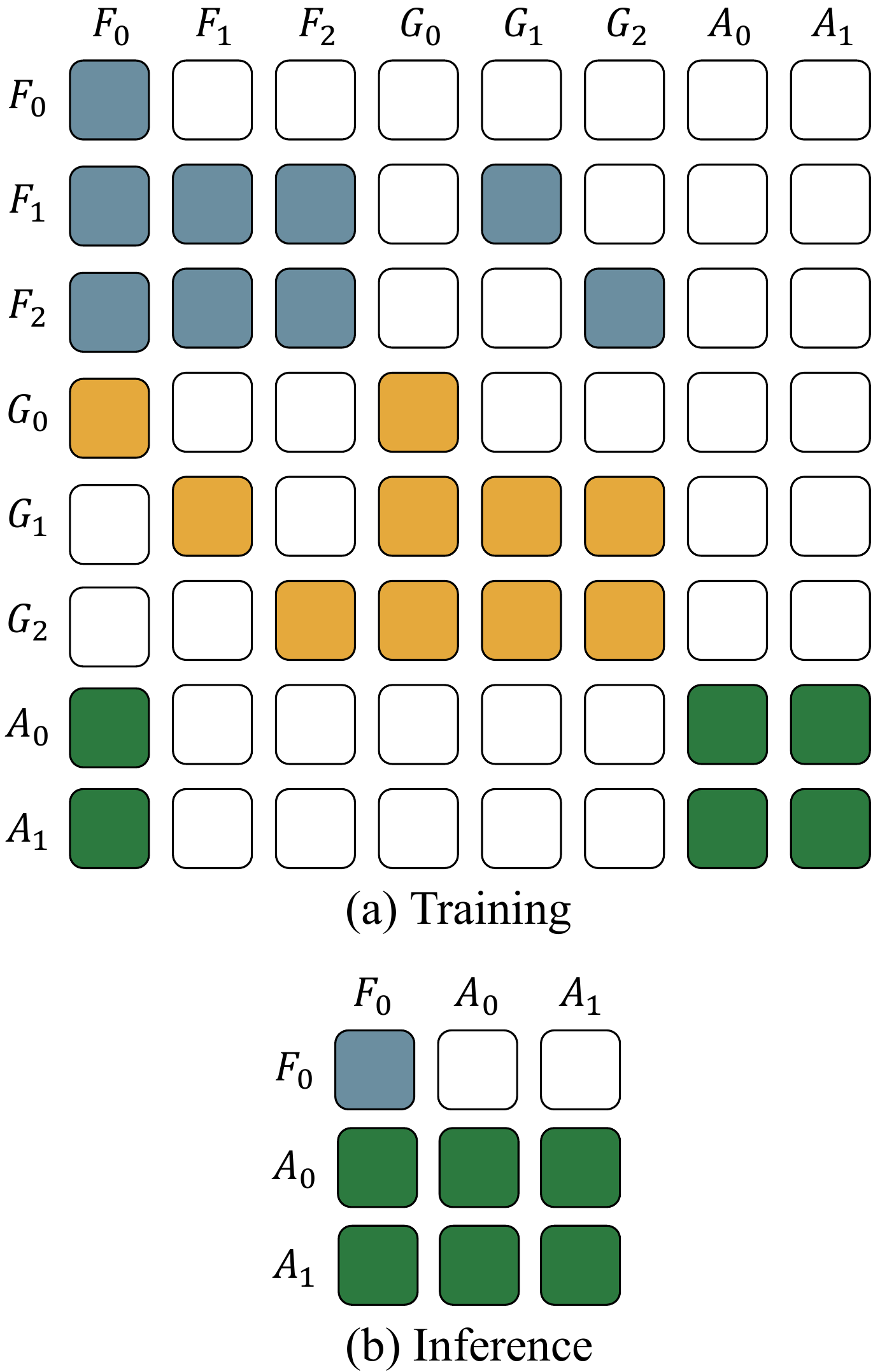}
    \caption{\small Attention mask of FastWAM. $F$ denotes video frames, $G$ indicates predictive geometric representation and $A$ represents actions to be denoised.}
    \label{fastwam_attention}
\end{figure}
Specifically, FastWAM consists of a visual world modeling branch implemented with a WAN\citep{wan2025wan} and an action-generation branch. To incorporate the predictive geometric representation, we introduce an additional geometry branch with an architecture and initialization strategy analogous to the original action branch. The geometry branch is trained to model the predictive geometric representations provided by the pretrained representation model, and operates in parallel with the original world and action modeling branches. This enables FastWAM to receive explicit supervision on the structured evolution of scene geometry in addition to its original visual world-prediction objective. The motor representation is incorporated through the action-generation pathway. Rather than introducing an additional generative branch for the motor representation, we use the pretrained motor representation as a latent guidance target for intermediate features within the action branch. Specifically, the corresponding intermediate action features are aligned with the motor representation using the same representation-level alignment objective as in the VLA setting. This provides action-aware geometric information directly to the branch responsible for generating robot actions, while leaving the native action modeling objective of FastWAM unchanged.

Because the predictive geometric representation is organized into temporal slots, an additional temporal alignment mechanism is required to establish correspondence between the geometric representation and the visual tokens in FastWAM. We therefore introduce a structured attention mask that explicitly couples temporally corresponding video and geometry slots. As illustrated in Fig.~\ref{fastwam_attention}, each video slot is allowed to attend to its corresponding geometry slot, enabling the visual representation at a given temporal position to access the geometric information associated with the same stage of the interaction. Conversely, each geometry slot is also allowed to attend to the corresponding video slot. This bidirectional local visibility establishes an explicit correspondence between visual and geometric representations while preventing a geometry slot from indiscriminately attending to temporally unrelated video tokens. The resulting attention pattern therefore encourages localized video-geometry interaction: Such slot-level correspondence allows the model to jointly reason about visual appearance and geometric structure at matched temporal positions, while preserving the temporal organization of the predictive representation. The mask is applied in addition to the native attention structure of FastWAM and does not require additional geometric observations at inference time. During inference, the additional geometry branch and the pretrained representation model are not required to receive externally provided geometric observations. The downstream FastWAM operates using its standard current-time visual input, from which the learned model generates the future world evolution and corresponding actions. Thus, as in the VLA setting, the predictive geometric and motor representations are used to provide privileged supervision during training, while no extra geometry input is required during deployment.


\appsection{Limitations and Future Work}
Our current framework primarily focuses on visual geometry and its temporal evolution to construct predictive and action-aware representations. An interesting direction for future work is to further incorporate tactile information into the same representation-learning framework. Tactile observations provide complementary cues about contact states, interaction forces, and fine-grained object configurations that may not be fully observable from vision alone. Integrating visual geometry with tactile feedback could therefore provide a richer representation of scene evolution, particularly for contact-rich manipulation and interactions involving occlusion or subtle physical state changes. More broadly, exploring multimodal predictive representations that jointly capture visual, geometric, and tactile evolution may further strengthen the connection between world modeling and action generation.

\appsection{Simulation and Real-World Evaluation Details}

\appsubsection{Simulation Evaluation Details}
We provide detailed descriptions of the simulation benchmarks employed in our experiments. 

\noindent\textbf{SimplerEnv Benchmark.}
To assess cross-embodiment generalization and investigate the correlation between simulation performance and real-world performance, we conduct experiments on the SimplerEnv benchmark~\citep{simpleenv} using the WidowX robot.

For the WidowX robot, we use the \textit{Visual Matching} setting to ensure consistency with the data distribution of the Bridge dataset\citep{Bridge_data}. We evaluate the robot on four manipulation tasks: \textit{Put Carrot on Plate} (\textit{Put Carrot}), \textit{Put Spoon on Towel} (\textit{Put Spoon}), \textit{Stack Green on Yellow} (\textit{Stack Cube}), and \textit{Put Eggplant in Basket} (\textit{Put Eggplant}). 

\noindent\textbf{LIBERO Benchmark.}
We additionally evaluate our method on the LIBERO benchmark~\citep{Libero} to examine different dimensions of temporal reasoning and knowledge transfer. In particular, we consider four task suites: \textit{Spatial}, which evaluates spatial reasoning under different object configurations; \textit{Object}, which focuses on object-centric manipulation involving diverse object categories; \textit{Goal}, which evaluates goal-conditioned instruction following in static environments; and \textit{Long}, which targets long-horizon planning over multi-stage manipulation tasks. Each suite consists of 10 tasks, and each task is provided with 50 human-teleoperated demonstrations.

\noindent\textbf{LIBERO-Plus Benchmark.}
LIBERO-Plus~\citep{fei2025libero} evaluates the robustness of policies trained on LIBERO under controlled perturbations at test time. It introduces variations across seven dimensions: \emph{Camera Viewpoints}, \emph{Robot Initial States}, \emph{Language Instructions}, \emph{Light Conditions}, \emph{Background Textures}, \emph{Sensor Noise}, and \emph{Object Layouts}. These perturbations are categorized into five difficulty levels, resulting in a total of 10{,}030 evaluation instances. We directly evaluate the checkpoint trained on LIBERO without performing any adaptation to the LIBERO-Plus perturbations. This evaluation protocol is designed to examine whether the learned policy representations remain effective when the observation distribution, robot initialization, or instruction formulation differs from those encountered during training.

\appsubsection{Real-World Robot Evaluation Details}
\begin{figure*}[!t]
\centering
\includegraphics[width=1.0\linewidth]{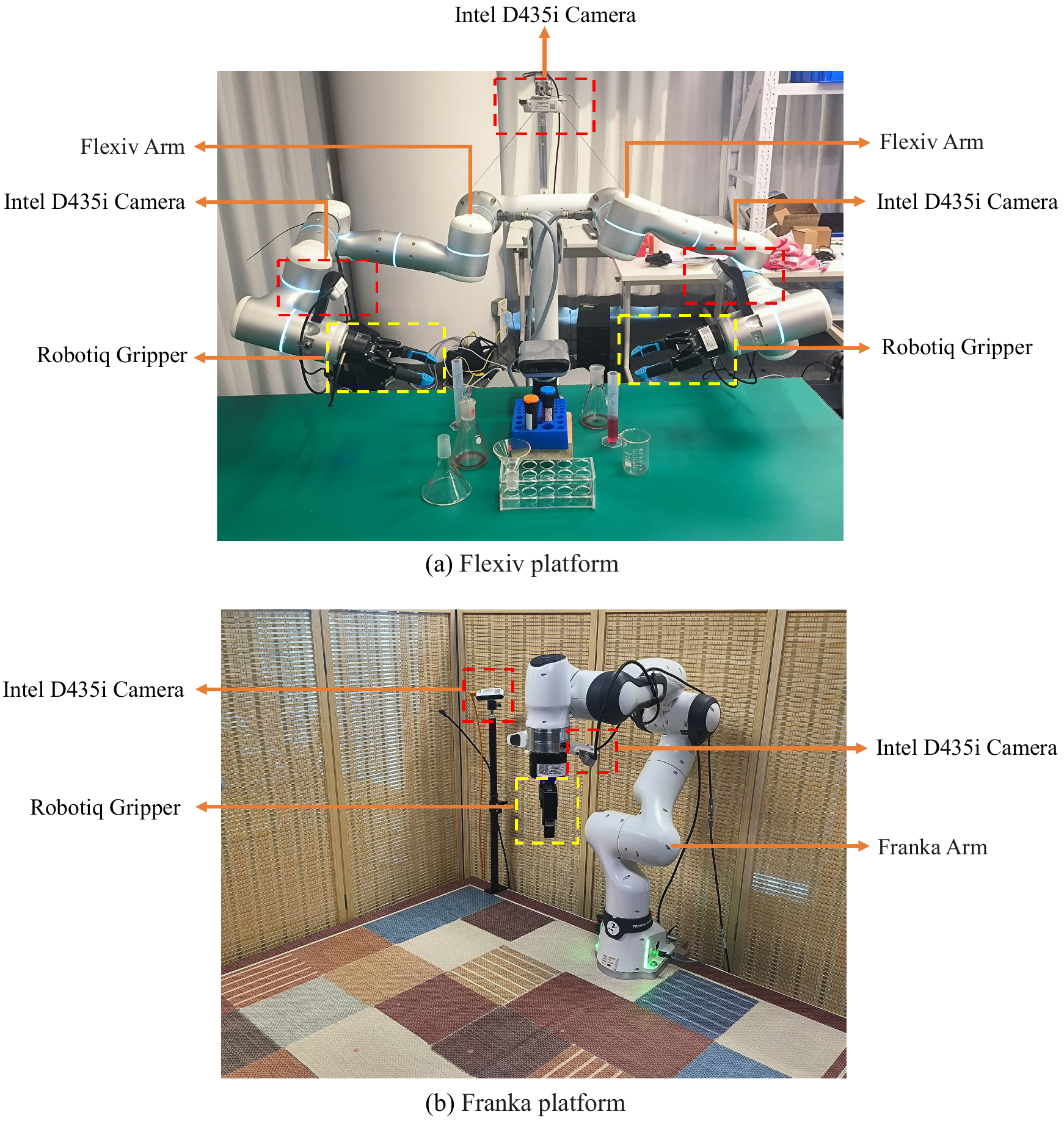}
\caption{\small Bimanual and single-arm manipulation platforms.}
\label{platform}
\end{figure*}

\begin{figure*}[!t]
\centering
\includegraphics[width=1.0\linewidth]{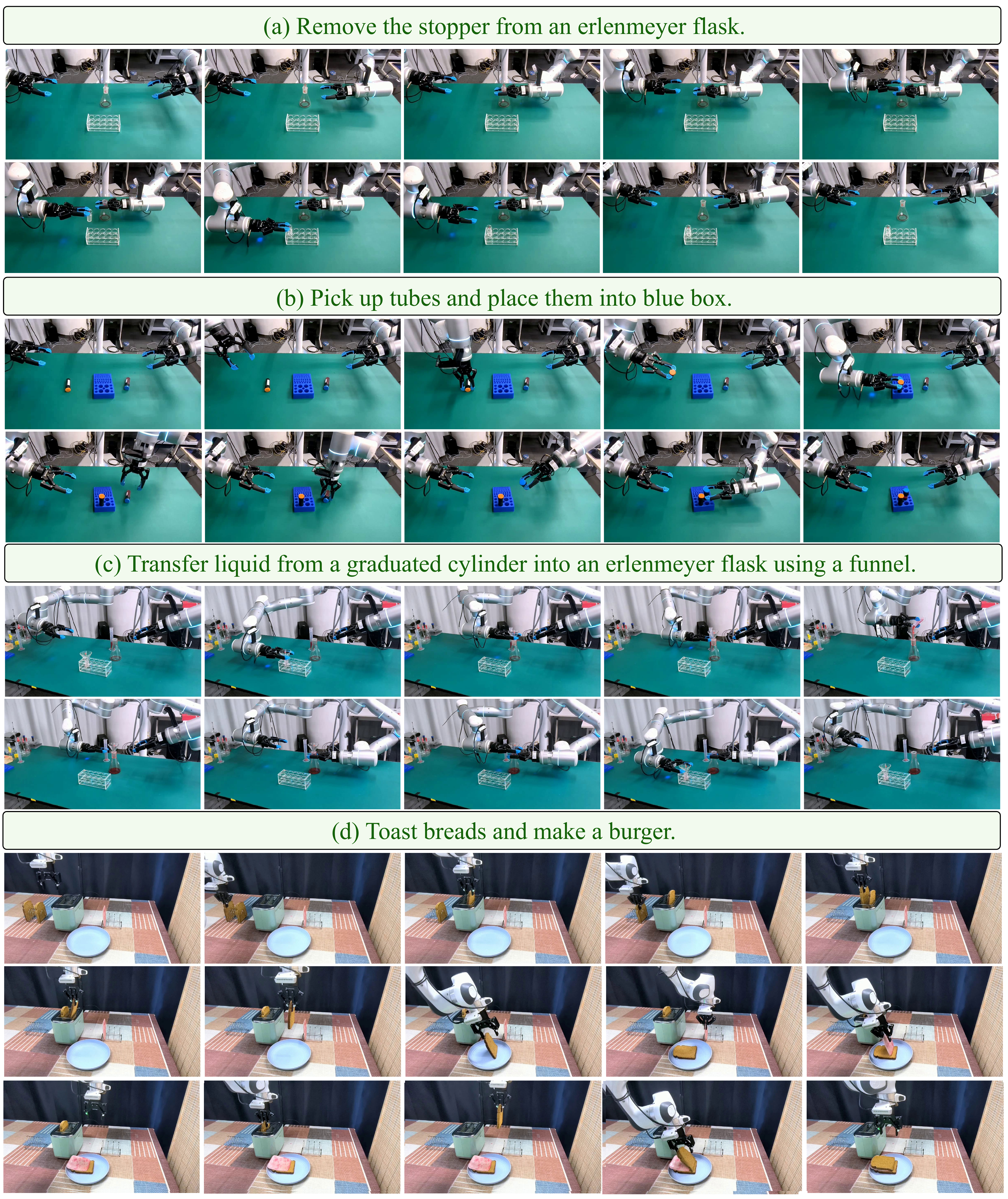}
\caption{\small Visualization of four real-robot manipulation tasks.}
\label{real_vis}
\end{figure*}

\noindent\textbf{Real-World Robot Setup.}
We conduct real-world experiments on both bimanual and single-arm manipulation platforms to evaluate the proposed CPR guidance under different embodiments and interaction requirements, as shown in Fig.\ref{platform}. For the bimanual setting, we use dual Flexiv robotic arms, each equipped with a Robotiq 2F-85 gripper. Visual observations are captured using three Intel RealSense D435i RGB-D cameras, including two wrist-mounted cameras and one head-mounted camera. The combination of third-person and wrist-view observations provides complementary global and local views of the manipulation scene, which is particularly important for bimanual tasks involving object coordination and fine-grained manipulation.

For the single-arm setting, we use a Franka Research 3 robotic platform, equipped with a 7-DoF manipulator and joint-level torque and force sensing. Visual observations are obtained from an Intel RealSense D435i RGB-D camera mounted in an eye-on-base configuration, providing a global view of the manipulation workspace. We additionally mount a second RealSense camera on the robot wrist to provide close-up observations of the manipulated objects and end-effector interaction. The robot is equipped with a Robotiq 2F-85 gripper for end-effector manipulation. These multi-view configurations provide both scene-level spatial context and fine-grained object-interaction cues for downstream action generation.

\noindent\textbf{Baseline Models.}
We compare our method with representative VLA baselines using different pretrained visual-language backbones. \textbf{OpenVLA}\citep{openvla} is initialized from the publicly released checkpoint pretrained on the OXE\citep{open_x_embodiment} dataset. For our real-world experiments, we extend the original input interface to jointly process third-person and wrist-view images. We also evaluate \textbf{$\pi_0$}\citep{black2024pi_0}, initialized from the official JAX checkpoint pretrained on the DROID\citep{khazatsky2024droid} dataset. Unless otherwise specified, the same downstream training protocol is used when incorporating CPR into the corresponding baseline model. 

\noindent\textbf{Real-World Task Suites.}
\label{task_suits}
As shown in Fig.\ref{real_vis}, we evaluate four real-world manipulation tasks covering bimanual coordination, geometric perception, contact-sensitive manipulation, and long-horizon task execution. The first three tasks are performed with the bimanual Flexiv setup, while the final task is conducted using the single-arm Franka platform.

\begin{itemize}
\item \textbf{Remove a Stopper.}
The robot removes a stopper from an erlenmeyer flask and places it at a designated holder. The task first requires the left arm to grasp and stabilize the flask, after which the right arm grasps the stopper and places it accurately into the corresponding circular opening on the storage rack. Successful execution therefore requires precise relative positioning between the two arms and the manipulated objects, as well as coordinated bimanual actions during the stopper removal and placement stages.

\item \textbf{Pick up Tubes into Box.}
The robot sequentially transfers two tubes into a blue box using both arms. The left arm first grasps the brown tube and places it into the box, followed by the right arm grasping the blue tube and placing it into the corresponding location. The task requires accurate perception of object locations and target geometry, together with appropriate end-effector positioning for grasping and insertion. Since the two manipulation stages involve different objects and hand-object configurations, the policy must maintain consistent spatial grounding throughout the sequential execution.

\item \textbf{Transfer Liquid into an Erlenmeyer.}
The robot transfers liquid from a graduated cylinder into an erlenmeyer flask using a funnel. The right arm first grasps the funnel and inserts it into the flask with sufficient spatial precision. It then grasps the liquid-filled graduated cylinder and carefully tilts it to pour the liquid through the funnel. After the transfer, the right arm stabilizes the flask while the left arm removes the funnel and returns it to a designated holder on the rack. This task combines fine-grained geometric alignment, coordinated bimanual manipulation, sequential object-role changes, and long-horizon execution, making the relative spatial configuration of the objects and tools important throughout the interaction.

\item \textbf{Toast Breads and Make a Burger.}
This task is performed on the single-arm Franka platform and involves a longer sequence of manipulation stages. The robot first places two slices of bread into the toaster, then transfers one slice onto a plate. It subsequently grasps a meat patty from the rack and stacks it onto the bread, before placing the remaining bread slice on top to complete the burger. The task requires maintaining the correct object order and intermediate states across multiple manipulation stages, while accurately positioning each object for grasping, placement, and stacking.

\end{itemize}

\noindent\textbf{Real-World Data Collection.}
For downstream post-training, we collect 100 demonstrations for each of the three bimanual tasks and 60 demonstrations for the single-arm task. The collected trajectories are used to train the downstream policies.

\appsection{Details of Compared Methods}
\label{sec:compared_methods}

We compare our approach with representative methods from three groups: standard policy baselines, input-centric methods, and prediction-centric methods. This categorization distinguishes how geometric or predictive information is introduced into policy learning: standard policies rely primarily on the model's native visual-language-action inputs, input-centric methods explicitly augment the current observation with geometric or spatial information, while prediction-centric methods introduce additional predictive or auxiliary supervision to encourage the policy to learn future, temporal, or geometric structure.

\noindent\textbf{Standard Policy Baselines.}
These methods generate robot actions from visual observations, language instructions, and, when applicable, proprioceptive states, without introducing an explicit future-representation or auxiliary geometric prediction target.

\begin{itemize}
\item \textbf{OpenVLA-OFT}\citep{openvla-oft}.
OpenVLA-OFT~\citep{openvla-oft} is an optimized post-training recipe for OpenVLA\citep{openvla} that studies action representations, decoding strategies, and fine-tuning objectives. Its policy uses continuous action representations together with action chunking, parallel decoding, and an $\ell_1$ regression objective for efficient multi-step action prediction. These design choices provide a strong and efficient baseline for adapting pretrained VLA models to downstream manipulation tasks.

\item \textbf{$\pi_0$}\citep{black2024pi_0}.
$\pi_0$ is a general-purpose VLA that augments a pretrained vision-language model with a dedicated action expert. The action expert generates continuous action chunks using conditional flow matching, allowing the model to produce smooth, high-frequency robot trajectories across multiple embodiments. The model is pretrained on diverse multi-robot manipulation data and subsequently adapted to downstream tasks through post-training.

\end{itemize}

\noindent\textbf{Input-Centric Methods.}
Input-centric methods enhance the policy primarily by providing additional geometric or spatial information derived from the current observation. Rather than constructing a predictive target from future observations, these approaches enrich the policy's current-time spatial representation through depth, point clouds, geometric embeddings, or spatially structured tokens.

\begin{itemize}
\item \textbf{GeoVLA}\citep{sun2025geovla}.
GeoVLA augments a VLA with explicit 3D geometric information derived from depth observations. Depth maps are converted into point clouds and processed by a dedicated Point Embedding Network to obtain 3D geometric embeddings. These embeddings are combined with the vision-language representation and processed by a 3D-enhanced action expert, providing the policy with explicit spatial information for action generation.

\item \textbf{3D-CAVLA}~\citep{bhat20253d}.
3D-CAVLA enhances VLA spatial awareness by incorporating depth perception and task-oriented visual context. The framework combines depth information with chain-of-thought reasoning and region-of-interest detection to focus the policy on task-relevant objects and spatial relationships. Its goal is to improve scene-context understanding and generalization to unseen manipulation tasks.

\item \textbf{SpatialVLA}~\citep{spatialvla}.
SpatialVLA explicitly introduces spatial representations into a generalist robot policy through Ego3D Position Encoding, which injects 3D positional information into the visual representation. It further uses Adaptive Action Grids to represent spatial robot movements in a discretized and robot-adaptive action space, facilitating spatially grounded manipulation and transfer across robot setups.

\item \textbf{FALCON}~\citep{zhang2025spatial}.
FALCON injects spatially rich tokens into the action-generation pathway of a VLA through an Embodied Spatial Model and a Spatial-Enhanced Action Head. The spatial branch can use RGB input alone and can additionally incorporate geometric modalities such as depth or camera pose when available. By consuming spatial tokens directly at the action head rather than concatenating them into the vision-language backbone, FALCON aims to strengthen 3D spatial understanding while preserving the semantic representation of the VLA.

\end{itemize}

\noindent\textbf{Prediction-Centric Methods.}
Prediction-centric methods introduce additional predictive or temporally structured supervision during policy learning. Rather than only enriching the current observation, these approaches encourage the policy to capture future scene evolution, depth structure, robot motion, or other predictive information that can support action generation.

\begin{itemize}
\item \textbf{4D-WAM}~\citep{yang20264d}.
4D-WAM introduces spatiotemporal supervision into World Action Models using 3D trajectory fields. It employs motion alignment to encourage consistency of feature changes across adjacent frames and destination alignment to guide the model toward long-horizon target information. These objectives provide local motion and long-range destination supervision through representation alignment, without requiring additional geometric inputs at inference time.

\item \textbf{ELAN4D}~\citep{he2026elan4d}.
ELAN4D provides embodiment-centric 4D supervision to VLA policies through future robot keypoint tracks. The tracks are derived from proprioceptive states using forward kinematics and describe the 3D displacement of robot joints and end-effectors over time. A lightweight auxiliary track decoder injects this predictive spatiotemporal supervision into the action-generation pathway during training, while the auxiliary branch can be removed at inference time.

\item \textbf{QDepth-VLA}~\citep{li2025qdepth}.
QDepth-VLA introduces depth prediction as an auxiliary supervision task for VLA training. A dedicated depth expert predicts quantized latent depth tokens obtained from a VQ-VAE representation of depth maps, encouraging the policy to learn depth-aware features that capture geometric structure relevant to manipulation. Unlike methods that directly provide depth as an additional input, QDepth-VLA uses depth prediction to shape the learned representation through auxiliary supervision.

\item \textbf{WoG}~\citep{su2026world}.
World Guidance (WoG) maps future observations into compact condition representations and uses them to guide action generation. The VLA is trained to jointly predict the compressed future condition and the corresponding action trajectory, introducing future-world modeling supervision in a compact condition space rather than requiring full future-image reconstruction. This design aims to retain predictive information that is useful for action generation while keeping the prediction target compact.

\item \textbf{WorldVLA}~\citep{worldvla}.
WorldVLA integrates visual world modeling and action generation within a unified autoregressive framework. The world model predicts future visual observations conditioned on the current scene and actions, while the action model predicts subsequent actions from visual observations. The two components are jointly trained so that future-world modeling and action generation provide mutual supervision. WorldVLA further introduces selective attention masking to reduce error propagation from previously generated actions during autoregressive action prediction.

\item \textbf{FLARE}~\citep{zheng2025flare}.
FLARE pretrains an action-aware observation embedding and introduces learnable future tokens into a diffusion policy, aligning their intermediate features with future observation embeddings. Our distinction is that CPR derives its motor representation specifically from action-supervised transitions between consecutive predictive geometric states, rather than directly aligning policy tokens with action-aware future observation embeddings. Meanwhile, CPR retains a separate predictive geometric representation to preserve structured scene information.

\item \textbf{GeoPredict}~\citep{qian2026geopredict}.
GeoPredict introduces predictive kinematic trajectories and future 3D Gaussian geometry as complementary training-time supervision for manipulation. This establishes an important precedent for using future 3D geometric information to guide action generation. CPR differs in its representation construction: rather than directly predicting future kinematic trajectories and 3D Gaussian geometry, it learns compact temporal geometric states through reconstruction, masked state prediction, and transition prediction, and subsequently derives an action-aware motor representation from their geometric transitions.

\item \textbf{DBIC-BM}~\citep{xu2025diffusion}.
\citep{xu2025diffusion}jointly predicts compressed future latents and actions, removing the future prediction branch at inference. CPR focuses on the complementary organization of geometric and motor representations and their distinct downstream transfer pathways.

\item \textbf{GAP}~\citep{xu2026action}.
GAP leverages pretrained 3D geometric priors and jointly predicts future action chunks and 3D geometric representations through conditional diffusion. In contrast, CPR decouples predictive geometry and action-aware motion into complementary representations, which are separately transferred to downstream policies through geometry-query supervision and motor-representation alignment.

\end{itemize}

{
    \small
    \bibliographystyle{ieeenat_fullname}
    \bibliography{main}

\begin{thebibliography}{100}
\providecommand{\url}[1]{\texttt{#1}}
\providecommand{\urlprefix}{URL }
\providecommand{\doi}[1]{https://doi.org/#1}

\bibitem{agarwal2026cosmos}
Agarwal, N., Ali, A., Allen, J., Antolini, M., Aubame, A., Azzolini, A., Bai, J., Bala, M., Balaji, Y., Bapst, J., et~al.: Cosmos 3: Omnimodal world models for physical ai. arXiv preprint arXiv:2606.02800  (2026)

\bibitem{agarwal2025cosmos}
Agarwal, N., Ali, A., Bala, M., Balaji, Y., Barker, E., Cai, T., Chattopadhyay, P., Chen, Y., Cui, Y., Ding, Y., et~al.: Cosmos world foundation model platform for physical ai. arXiv preprint arXiv:2501.03575  (2025)

\bibitem{Alpher02}
Alpher, F.: Frobnication. IEEE TPAMI  \textbf{12}(1),  234--778 (2002)

\bibitem{Alpher03}
Alpher, F., Fotheringham-Smythe, F.: Frobnication revisited. Journal of Foo  \textbf{13}(1),  234--778 (2003)

\bibitem{Alpher04}
Alpher, F., Fotheringham-Smythe, F., Gamow, F.: Can a machine frobnicate? Journal of Foo  \textbf{14}(1),  234--778 (2004)

\bibitem{Alpher05}
Alpher, F., Gamow, F.: Can a computer frobnicate? In: CVPR. pp. 234--778 (2005)

\bibitem{assran2023self}
Assran, M., Duval, Q., Misra, I., Bojanowski, P., Vincent, P., Rabbat, M., LeCun, Y., Ballas, N.: Self-supervised learning from images with a joint-embedding predictive architecture. In: Proceedings of the IEEE/CVF Conference on Computer Vision and Pattern Recognition. pp. 15619--15629 (2023)

\bibitem{assran2025v}
Assran, M., Bardes, A., Fan, D., Garrido, Q., Howes, R., Muckley, M., Rizvi, A., Roberts, C., Sinha, K., Zholus, A., et~al.: V-jepa 2: Self-supervised video models enable understanding, prediction and planning. arXiv preprint arXiv:2506.09985  (2025)

\bibitem{belkhale2023hydra}
Belkhale, S., Cui, Y., Sadigh, D.: Hydra: Hybrid robot actions for imitation learning. In: Conference on Robot Learning. pp. 2113--2133. PMLR (2023)

\bibitem{bhat20253d}
Bhat, V., Lan, Y.H., Krishnamurthy, P., Karri, R., Khorrami, F.: 3d cavla: Leveraging depth and 3d context to generalize vision language action models for unseen tasks. arXiv preprint arXiv:2505.05800  (2025)

\bibitem{bjorck2025gr00t}
Bjorck, J., Casta{\~n}eda, F., Cherniadev, N., Da, X., Ding, R., Fan, L., Fang, Y., Fox, D., Hu, F., Huang, S., et~al.: Gr00t n1: An open foundation model for generalist humanoid robots. arXiv preprint arXiv:2503.14734  (2025)

\bibitem{black2024pi_0}
Black, K., Brown, N., Driess, D., Esmail, A., Equi, M., Finn, C., Fusai, N., Groom, L., Hausman, K., Ichter, B., et~al.: $pi\_0$: A vision-language-action flow model for general robot control. arXiv preprint arXiv:2410.24164  (2024)

\bibitem{RT-2}
Brohan, A., Brown, N., Carbajal, J., Chebotar, Y., Chen, X., Choromanski, K., Ding, T., Driess, D., Dubey, A., Finn, C., et~al.: Rt-2: Vision-language-action models transfer web knowledge to robotic control. arXiv preprint arXiv:2307.15818  (2023)

\bibitem{RT-1}
Brohan, A., Brown, N., Carbajal, J., Chebotar, Y., Dabis, J., Finn, C., Gopalakrishnan, K., Hausman, K., Herzog, A., Hsu, J., et~al.: Rt-1: Robotics transformer for real-world control at scale. arXiv preprint arXiv:2212.06817  (2022)

\bibitem{brohan2022rt}
Brohan, A., Brown, N., Carbajal, J., Chebotar, Y., Dabis, J., Finn, C., Gopalakrishnan, K., Hausman, K., Herzog, A., Hsu, J., et~al.: Rt-1: Robotics transformer for real-world control at scale. arXiv preprint arXiv:2212.06817  (2022)

\bibitem{bruce2024genie}
Bruce, J., Dennis, M.D., Edwards, A., Parker-Holder, J., Shi, Y., Hughes, E., Lai, M., Mavalankar, A., Steigerwald, R., Apps, C., et~al.: Genie: Generative interactive environments. In: Forty-first International Conference on Machine Learning (2024)

\bibitem{bu2025agibot}
Bu, Q., Cai, J., Chen, L., Cui, X., Ding, Y., Feng, S., Gao, S., He, X., Hu, X., Huang, X., et~al.: Agibot world colosseo: A large-scale manipulation platform for scalable and intelligent embodied systems. arXiv preprint arXiv:2503.06669  (2025)

\bibitem{bu2025univla}
Bu, Q., Yang, Y., Cai, J., Gao, S., Ren, G., Yao, M., Luo, P., Li, H.: Univla: Learning to act anywhere with task-centric latent actions. arXiv preprint arXiv:2505.06111  (2025)

\bibitem{cai2026internvla}
Cai, J., Cai, Z., Cao, J., Chen, Y., He, Z., Jiang, L., Li, H., Li, H., Li, Y., Liu, Y., et~al.: Internvla-a1: Unifying understanding, generation and action for robotic manipulation. arXiv preprint arXiv:2601.02456  (2026)

\bibitem{cai2026internvla_a1}
Cai, J., Cai, Z., Cao, J., Chen, Y., He, Z., Jiang, L., Li, H., Li, H., Li, Y., Liu, Y., et~al.: Internvla-a1: Unifying understanding, generation and action for robotic manipulation. arXiv preprint arXiv:2601.02456  (2026)

\bibitem{worldvla}
Cen, J., Yu, C., Yuan, H., Jiang, Y., Huang, S., Guo, J., Li, X., Song, Y., Luo, H., Wang, F., et~al.: Worldvla: Towards autoregressive action world model. arXiv preprint arXiv:2506.21539  (2025)

\bibitem{chen2023berkeley}
Chen, L.Y., Adebola, S., Goldberg, K.: Berkeley ur5 demonstration dataset. URL https://sites. google. com/view/berkeley-ur5/home  (2023)

\bibitem{chen2025robotwin}
Chen, T., Chen, Z., Chen, B., Cai, Z., Liu, Y., Li, Z., Liang, Q., Lin, X., Ge, Y., Gu, Z., et~al.: Robotwin 2.0: A scalable data generator and benchmark with strong domain randomization for robust bimanual robotic manipulation. arXiv preprint arXiv:2506.18088  (2025)

\bibitem{chen2025villa}
Chen, X., Wei, H., Zhang, P., Zhang, C., Wang, K., Guo, Y., Yang, R., Wang, Y., Xiao, X., Zhao, L., et~al.: Villa-x: enhancing latent action modeling in vision-language-action models. arXiv preprint arXiv:2507.23682  (2025)

\bibitem{chen2025internvla}
Chen, X., Chen, Y., Fu, Y., Gao, N., Jia, J., Jin, W., Li, H., Mu, Y., Pang, J., Qiao, Y., et~al.: Internvla-m1: A spatially guided vision-language-action framework for generalist robot policy. arXiv preprint arXiv:2510.13778  (2025)

\bibitem{open_x_embodiment}
Collaboration, O.X.E.: Open {X-E}mbodiment: Robotic learning datasets and {RT-X} models. \url{https://arxiv.org/abs/2310.08864} (2023)

\bibitem{cui2022play}
Cui, Z.J., Wang, Y., Shafiullah, N.M.M., Pinto, L.: From play to policy: Conditional behavior generation from uncurated robot data. arXiv preprint arXiv:2210.10047  (2022)

\bibitem{dassclvr}
Dass, S., Yapeter, J., Zhang, J., Zhang, J., Pertsch, K., Nikolaidis, S., Lim, J.J.: Clvr jaco play dataset, 2023. URL https://github. com/clvrai/clvr\_jaco\_play\_dataset

\bibitem{ding2026skillplug}
Ding, Z.H., Wang, Z.: Skillplug: Unsupervised skill mining for few-shot adaptation in robotic manipulation. IEEE Robotics and Automation Letters  (2026)

\bibitem{dong2025interleaved}
Dong, S., Wang, S., Liu, X., Li, C., Hou, H., Wei, Z.: Interleaved latent visual reasoning with selective perceptual modeling. arXiv preprint arXiv:2512.05665  (2025)

\bibitem{pi_05KI}
Driess, D., Springenberg, J.T., Ichter, B., Yu, L., Li-Bell, A., Pertsch, K., Ren, A.Z., Walke, H., Vuong, Q., Shi, L.X., et~al.: Knowledge insulating vision-language-action models: Train fast, run fast, generalize better. arXiv preprint arXiv:2505.23705  (2025)

\bibitem{Bridge_data}
Ebert, F., Yang, Y., Schmeckpeper, K., Bucher, B., Georgakis, G., Daniilidis, K., Finn, C., Levine, S.: Bridge data: Boosting generalization of robotic skills with cross-domain datasets. arXiv preprint arXiv:2109.13396  (2021)

\bibitem{ebert2021bridge}
Ebert, F., Yang, Y., Schmeckpeper, K., Bucher, B., Georgakis, G., Daniilidis, K., Finn, C., Levine, S.: Bridge data: Boosting generalization of robotic skills with cross-domain datasets. arXiv preprint arXiv:2109.13396  (2021)

\bibitem{elman1990finding}
Elman, J.L.: Finding structure in time. Cognitive science  \textbf{14}(2),  179--211 (1990)

\bibitem{esser2021taming}
Esser, P., Rombach, R., Ommer, B.: Taming transformers for high-resolution image synthesis. In: Proceedings of the IEEE/CVF conference on computer vision and pattern recognition. pp. 12873--12883 (2021)

\bibitem{fan2025interleave}
Fan, C., Jia, X., Sun, Y., Wang, Y., Wei, J., Gong, Z., Zhao, X., Tomizuka, M., Yang, X., Yan, J., et~al.: Interleave-vla: Enhancing robot manipulation with interleaved image-text instructions. arXiv preprint arXiv:2505.02152  (2025)

\bibitem{fei2025libero}
Fei, S., Wang, S., Shi, J., Dai, Z., Cai, J., Qian, P., Ji, L., He, X., Zhang, S., Fei, Z., et~al.: Libero-plus: In-depth robustness analysis of vision-language-action models. arXiv preprint arXiv:2510.13626  (2025)

\bibitem{goyal2017something}
Goyal, R., Kahou, S.E., Michalski, V., Materzynska, J., Westphal, S., Kim, H., Haenel, V., Fruend, I., Yianilos, P., Mueller-Freitag, M., et~al.: The “something something” video database for learning and evaluating visual common sense. In: 2017 IEEE international conference on computer vision (ICCV). pp. 5843--5851. IEEE (2017)

\bibitem{grauman2022ego4d}
Grauman, K., Westbury, A., Byrne, E., Chavis, Z., Furnari, A., Girdhar, R., Hamburger, J., Jiang, H., Liu, M., Liu, X., et~al.: Ego4d: Around the world in 3,000 hours of egocentric video. In: Proceedings of the IEEE/CVF conference on computer vision and pattern recognition. pp. 18995--19012 (2022)

\bibitem{guo2026unified}
Guo, J., Li, Q., Li, P., Chen, Z., Sun, N., Su, Y., Wang, H., Zhang, Y., Li, X., Liu, H.: Unified 4d world action modeling from video priors with asynchronous denoising. arXiv preprint arXiv:2604.26694  (2026)

\bibitem{he2016identity}
He, K., Zhang, X., Ren, S., Sun, J.: Identity mappings in deep residual networks. In: European conference on computer vision. pp. 630--645. Springer (2016)

\bibitem{he2026elan4d}
He, Z., Yang, B., Fang, Z., Zhou, K., Jiang, L., Qian, J., Mo, F., Yan, J., Torr, P., Li, X., et~al.: Elan4d: Embodiment-centric 4d supervision for vision-language-action models via plug-and-play adaptation. arXiv preprint arXiv:2605.30484  (2026)

\bibitem{heo2025furniturebench}
Heo, M., Lee, Y., Lee, D., Lim, J.J.: Furniturebench: Reproducible real-world benchmark for long-horizon complex manipulation. The International Journal of Robotics Research  \textbf{44}(10-11),  1863--1891 (2025)

\bibitem{hochreiter1997long}
Hochreiter, S., Schmidhuber, J.: Long short-term memory. Neural computation  \textbf{9}(8),  1735--1780 (1997)

\bibitem{huang2025roboground}
Huang, H., Chen, X., Chen, Y., Li, H., Han, X., Wang, Z., Wang, T., Pang, J., Zhao, Z.: Roboground: Robotic manipulation with grounded vision-language priors. In: Proceedings of the Computer Vision and Pattern Recognition Conference. pp. 22540--22550 (2025)

\bibitem{intelligence2025pi05}
Intelligence, P., Black, K., Brown, N., Darpinian, J., Dhabalia, K., Driess, D., Esmail, A., Equi, M., Finn, C., Fusai, N., et~al.: Pi0.5: a vision-language-action model with open-world generalization. arXiv preprint arXiv:2504.16054  (2025)

\bibitem{pi_05}
Intelligence, P., Black, K., Brown, N., Darpinian, J., Dhabalia, K., Driess, D., Esmail, A., Equi, M., Finn, C., Fusai, N., et~al.: $pi_{0.5}$: a vision-language-action model with open-world generalization. arXiv preprint arXiv:2504.16054  (2025)

\bibitem{jang2022bc}
Jang, E., Irpan, A., Khansari, M., Kappler, D., Ebert, F., Lynch, C., Levine, S., Finn, C.: Bc-z: Zero-shot task generalization with robotic imitation learning. In: conference on Robot Learning. pp. 991--1002. PMLR (2022)

\bibitem{jin2026umi}
Jin, S., Wang, Y., Duan, Y., Wu, D., Dong, G., Liu, X., Li, X., Jia, H., Zhang, Z., Wang, T., et~al.: Umi-bench 1.0: An open and reproducible real-world benchmark for tabletop robotic manipulation with umi data. arXiv preprint arXiv:2606.10382  (2026)

\bibitem{kalman1960new}
Kalman, R.E.: A new approach to linear filtering and prediction problems  (1960)

\bibitem{khazatsky2024droid}
Khazatsky, A., Pertsch, K., Nair, S., Balakrishna, A., Dasari, S., Karamcheti, S., Nasiriany, S., Srirama, M.K., Chen, L.Y., Ellis, K., et~al.: Droid: A large-scale in-the-wild robot manipulation dataset. arXiv preprint arXiv:2403.12945  (2024)

\bibitem{openvla-oft}
Kim, M.J., Finn, C., Liang, P.: Fine-tuning vision-language-action models: Optimizing speed and success. arXiv preprint arXiv:2502.19645  (2025)

\bibitem{openvla}
Kim, M.J., Pertsch, K., Karamcheti, S., Xiao, T., Balakrishna, A., Nair, S., Rafailov, R., Foster, E., Lam, G., Sanketi, P., et~al.: Openvla: An open-source vision-language-action model. arXiv preprint arXiv:2406.09246  (2024)

\bibitem{Authors14}
LastName, F.: The frobnicatable foo filter (2014), face and Gesture submission ID 324. Supplied as supplemental material {\tt fg324.pdf}

\bibitem{Authors14b}
LastName, F.: Frobnication tutorial (2014), supplied as supplemental material {\tt tr.pdf}

\bibitem{lecun2022path}
LeCun, Y.: A path towards autonomous machine intelligence version 0.9. 2, 2022-06-27. Open Review  \textbf{62}(1),  1--62 (2022)

\bibitem{lee2026act}
Lee, S., Mo, S., Han, W.S.: Act with intent: Distilling behavior intent for vision-language-action models. arXiv preprint arXiv:2608.23478  (2026)

\bibitem{lee2026mu_0}
Lee, S., Jung, Y., Lee, J., Shin, J., Shahidzadeh, A.H., Lee, Y.C., Kim, H.J., Huang, J.B., Huang, F.: $\backslash\mu_0$: A scalable 3d interaction-trace world model. arXiv preprint arXiv:2606.13769  (2026)

\bibitem{li2025spatial}
Li, F., Song, W., Zhao, H., Wang, J., Ding, P., Wang, D., Zeng, L., Li, H.: Spatial forcing: Implicit spatial representation alignment for vision-language-action model. arXiv preprint arXiv:2510.12276  (2025)

\bibitem{li2026spatial}
Li, F., Song, W., Zhao, H., Wang, J., Ding, P., Wang, D., Zeng, L., Li, H.: Spatial forcing: Implicit spatial representation alignment for vision-language-action model. In: International Conference on Learning Representations. vol.~2026, pp. 132324--132345 (2026)

\bibitem{li2026robointer}
Li, H., Wang, Z., Ding, Z.h., Yang, S., Chen, Y., Tian, Y., Hu, X., Wang, T., Lin, D., Zhao, F., et~al.: Robointer: A holistic intermediate representation suite towards robotic manipulation. arXiv preprint arXiv:2602.09973  (2026)

\bibitem{li2025cronusvla}
Li, H., Yang, S., Chen, Y., Tian, Y., Yang, X., Chen, X., Wang, H., Wang, T., Zhao, F., Lin, D., et~al.: Cronusvla: Transferring latent motion across time for multi-frame prediction in manipulation. arXiv preprint arXiv:2506.19816  (2025)

\bibitem{li2026causal}
Li, L., Zhang, Q., Luo, Y., Yang, S., Wang, R., Han, F., Yu, M., Gao, Z., Xue, N., Zhu, X., et~al.: Causal world modeling for robot control. arXiv preprint arXiv:2601.21998  (2026)

\bibitem{li2025switchvla}
Li, M., Zhao, Z., Che, Z., Liao, F., Wu, K., Xu, Z., Ren, P., Jin, Z., Liu, N., Tang, J.: Switchvla: Execution-aware task switching for vision-language-action models. arXiv preprint arXiv:2506.03574  (2025)

\bibitem{cogact}
Li, Q., Liang, Y., Wang, Z., Luo, L., Chen, X., Liao, M., Wei, F., Deng, Y., Xu, S., Zhang, Y., et~al.: Cogact: A foundational vision-language-action model for synergizing cognition and action in robotic manipulation. arXiv preprint arXiv:2411.19650  (2024)

\bibitem{UVA}
Li, S., Gao, Y., Sadigh, D., Song, S.: Unified video action model. arXiv preprint arXiv:2503.00200  (2025)

\bibitem{roboflamingo}
Li, X., Liu, M., Zhang, H., Yu, C., Xu, J., Wu, H., Cheang, C., Jing, Y., Zhang, W., Liu, H., Li, H., Kong, T.: Vision-language foundation models as effective robot imitators. arXiv preprint arXiv:2311.01378  (2023)

\bibitem{simpleenv}
Li, X., Hsu, K., Gu, J., Pertsch, K., Mees, O., Walke, H.R., Fu, C., Lunawat, I., Sieh, I., Kirmani, S., et~al.: Evaluating real-world robot manipulation policies in simulation. arXiv preprint arXiv:2405.05941  (2024)

\bibitem{li2025robotic}
Li, Y., Gong, Z., Li, H., Huang, X., Kang, H., Bai, G., Ma, X.: Robotic visual instruction. In: Proceedings of the Computer Vision and Pattern Recognition Conference. pp. 12155--12165 (2025)

\bibitem{li2026wam4d}
Li, Y., Wei, X., Cao, J., Wang, H., Chi, X., Bai, C., Sun, Q., Li, J., Zhang, X., Jia, P., et~al.: Wam4d: Fast 4d world action model via spatial register tokens. arXiv preprint arXiv:2606.14048  (2026)

\bibitem{li2025qdepth}
Li, Y., Chen, Y., Zhou, M., Li, H., Zhang, Z., Zhao, D.: Qdepth-vla: Quantized depth prediction as auxiliary supervision for vision-language-action models. arXiv preprint arXiv:2510.14836  (2025)

\bibitem{liang2025mm}
Liang, H., Chen, X., Wang, B., Chen, M., Liu, Y., Zhang, Y., Chen, Z., Yang, T., Chen, Y., Pang, J., et~al.: Mm-act: Learn from multimodal parallel generation to act. arXiv preprint arXiv:2512.00975  (2025)

\bibitem{liao2025genie}
Liao, Y., Zhou, P., Huang, S., Yang, D., Chen, S., Jiang, Y., Hu, Y., Cai, J., Liu, S., Luo, J., et~al.: Genie envisioner: A unified world foundation platform for robotic manipulation. arXiv preprint arXiv:2508.05635  (2025)

\bibitem{lin2025depth}
Lin, H., Chen, S., Liew, J., Chen, D.Y., Li, Z., Shi, G., Feng, J., Kang, B.: Depth anything 3: Recovering the visual space from any views. arXiv preprint arXiv:2511.10647  (2025)

\bibitem{Libero}
Liu, B., Zhu, Y., Gao, C., Feng, Y., Liu, Q., Zhu, Y., Stone, P.: Libero: Benchmarking knowledge transfer for lifelong robot learning. Advances in Neural Information Processing Systems  \textbf{36} (2024)

\bibitem{liu2025robot}
Liu, H., Nasiriany, S., Zhang, L., Bao, Z., Zhu, Y.: Robot learning on the job: Human-in-the-loop autonomy and learning during deployment. The International Journal of Robotics Research  \textbf{44}(10-11),  1727--1742 (2025)

\bibitem{luo2026being}
Luo, H., Zhang, W., Feng, Y., Zheng, S., Xu, H., Xu, C., Xi, Z., Fu, Y., Lu, Z.: Being-h0. 7: A latent world-action model from egocentric videos. arXiv preprint arXiv:2605.00078  (2026)

\bibitem{luo2024multistage}
Luo, J., Xu, C., Geng, X., Feng, G., Fang, K., Tan, L., Schaal, S., Levine, S.: Multistage cable routing through hierarchical imitation learning. IEEE Transactions on Robotics  \textbf{40},  1476--1491 (2024)

\bibitem{luo2025fmb}
Luo, J., Xu, C., Liu, F., Tan, L., Lin, Z., Wu, J., Abbeel, P., Levine, S.: Fmb: a functional manipulation benchmark for generalizable robotic learning. The International Journal of Robotics Research  \textbf{44}(4),  592--606 (2025)

\bibitem{lv2025f1}
Lv, Q., Kong, W., Li, H., Zeng, J., Qiu, Z., Qu, D., Song, H., Chen, Q., Deng, X., Pang, J.: F1: A vision-language-action model bridging understanding and generation to actions. arXiv preprint arXiv:2509.06951  (2025)

\bibitem{ma2026internvla}
Ma, H., Cai, J., Xu, X., Li, H., Yang, Y., Tian, Y., Cao, J., Zhu, H., Qiu, Z., Yang, Y., et~al.: Internvla-a1. 5: Unifying understanding, latent foresight, and action for compositional generalization. arXiv preprint arXiv:2607.04988  (2026)

\bibitem{mandlekar2019scaling}
Mandlekar, A., Booher, J., Spero, M., Tung, A., Gupta, A., Zhu, Y., Garg, A., Savarese, S., Fei-Fei, L.: Scaling robot supervision to hundreds of hours with roboturk: Robotic manipulation dataset through human reasoning and dexterity. In: 2019 IEEE/RSJ International Conference on Intelligent Robots and Systems (IROS). pp. 1048--1055. IEEE (2019)

\bibitem{mees2022grounding}
Mees, O., Borja-Diaz, J., Burgard, W.: Grounding language with visual affordances over unstructured data. arXiv preprint arXiv:2210.01911  (2022)

\bibitem{nasiriany2022learning}
Nasiriany, S., Gao, T., Mandlekar, A., Zhu, Y.: Learning and retrieval from prior data for skill-based imitation learning. arXiv preprint arXiv:2210.11435  (2022)

\bibitem{niu2024llarva}
Niu, D., Sharma, Y., Biamby, G., Quenum, J., Bai, Y., Shi, B., Darrell, T., Herzig, R.: Llarva: Vision-action instruction tuning enhances robot learning. arXiv preprint arXiv:2406.11815  (2024)

\bibitem{oquab2023dinov2}
Oquab, M., Darcet, T., Moutakanni, T., Vo, H., Szafraniec, M., Khalidov, V., Fernandez, P., Haziza, D., Massa, F., El-Nouby, A., et~al.: Dinov2: Learning robust visual features without supervision. arXiv preprint arXiv:2304.07193  (2023)

\bibitem{qian2026geopredict}
Qian, J., Han, B., Shi, C., Xiao, L., Yang, L., Shi, S., Jiang, L.: Geopredict: Leveraging predictive kinematics and 3d gaussian geometry for precise vla manipulation. In: Proceedings of the IEEE/CVF Conference on Computer Vision and Pattern Recognition. pp. 13529--13539 (2026)

\bibitem{spatialvla}
Qu, D., Song, H., Chen, Q., Yao, Y., Ye, X., Ding, Y., Wang, Z., Gu, J., Zhao, B., Wang, D., et~al.: Spatialvla: Exploring spatial representations for visual-language-action model. arXiv preprint arXiv:2501.15830  (2025)

\bibitem{quere2020shared}
Quere, G., Hagengruber, A., Iskandar, M., Bustamante, S., Leidner, D., Stulp, F., Vogel, J.: Shared control templates for assistive robotics. In: 2020 IEEE international conference on robotics and automation (ICRA). pp. 1956--1962. IEEE (2020)

\bibitem{rabiner2002tutorial}
Rabiner, L.R.: A tutorial on hidden markov models and selected applications in speech recognition. Proceedings of the IEEE  \textbf{77}(2),  257--286 (2002)

\bibitem{ranftl2021vision}
Ranftl, R., Bochkovskiy, A., Koltun, V.: Vision transformers for dense prediction. In: 2021 IEEE/CVF International Conference on Computer Vision (ICCV). pp. 12159--12168. IEEE (2021)

\bibitem{rosete2023latent}
Rosete-Beas, E., Mees, O., Kalweit, G., Boedecker, J., Burgard, W.: Latent plans for task-agnostic offline reinforcement learning. In: Conference on Robot Learning. pp. 1838--1849. PMLR (2023)

\bibitem{shafiullah2023bringing}
Shafiullah, N.M.M., Rai, A., Etukuru, H., Liu, Y., Misra, I., Chintala, S., Pinto, L.: On bringing robots home. arXiv preprint arXiv:2311.16098  (2023)

\bibitem{shah2023mutex}
Shah, R., Mart{\'\i}n-Mart{\'\i}n, R., Zhu, Y.: Mutex: Learning unified policies from multimodal task specifications. arXiv preprint arXiv:2309.14320  (2023)

\bibitem{smolvla}
Shukor, M., Aubakirova, D., Capuano, F., Kooijmans, P., Palma, S., Zouitine, A., Aractingi, M., Pascal, C., Russi, M., Marafioti, A., et~al.: Smolvla: A vision-language-action model for affordable and efficient robotics. arXiv preprint arXiv:2506.01844  (2025)

\bibitem{reconvla}
Song, W., Zhou, Z., Zhao, H., Chen, J., Ding, P., Yan, H., Huang, Y., Tang, F., Wang, D., Li, H.: Reconvla: Reconstructive vision-language-action model as effective robot perceiver. arXiv preprint arXiv:2508.10333  (2025)

\bibitem{su2026world}
Su, Y., Chen, S., Shi, H., Liu, M., Zhang, Z., Huang, N., Zhong, W., Zhu, Z., Liu, Y., Liu, X.: World guidance: World modeling in condition space for action generation. arXiv preprint arXiv:2602.22010  (2026)

\bibitem{sun2026vla}
Sun, J., Zhang, W., Qi, Z., Ren, S., Liu, Z., Zhu, H., Sun, G., Jin, X., Chen, Z.: Vla-jepa: Enhancing vision-language-action model with latent world model. arXiv preprint arXiv:2602.10098  (2026)

\bibitem{sun2025geovla}
Sun, L., Xie, B., Liu, Y., Shi, H., Wang, T., Cao, J.: Geovla: Empowering 3d representations in vision-language-action models. arXiv preprint arXiv:2508.09071  (2025)

\bibitem{team2025evaluating}
Team, G.R., Choromanski, K., Devin, C., Du, Y., Dwibedi, D., Gao, R., Jindal, A., Kipf, T., Kirmani, S., Leal, I., et~al.: Evaluating gemini robotics policies in a veo world simulator. arXiv preprint arXiv:2512.10675  (2025)

\bibitem{seer}
Tian, Y., Yang, S., Zeng, J., Wang, P., Lin, D., Dong, H., Pang, J.: Predictive inverse dynamics models are scalable learners for robotic manipulation. arXiv preprint arXiv:2412.15109  (2024)

\bibitem{tian2025interndata}
Tian, Y., Yang, Y., Xie, Y., Cai, Z., Shi, X., Gao, N., Liu, H., Jiang, X., Qiu, Z., Yuan, F., et~al.: Interndata-a1: Pioneering high-fidelity synthetic data for pre-training generalist policy. arXiv preprint arXiv:2511.16651  (2025)

\bibitem{vaswani2017attention}
Vaswani, A., Shazeer, N., Parmar, N., Uszkoreit, J., Jones, L., Gomez, A.N., Kaiser, {\L}., Polosukhin, I.: Attention is all you need. Advances in neural information processing systems  \textbf{30} (2017)

\bibitem{walke2023bridgedata}
Walke, H.R., Black, K., Zhao, T.Z., Vuong, Q., Zheng, C., Hansen-Estruch, P., He, A.W., Myers, V., Kim, M.J., Du, M., et~al.: Bridgedata v2: A dataset for robot learning at scale. In: Conference on robot learning. pp. 1723--1736. PMLR (2023)

\bibitem{wan2025wan}
Wan, T., Wang, A., Ai, B., Wen, B., Mao, C., Xie, C.W., Chen, D., Yu, F., Zhao, H., Yang, J., et~al.: Wan: Open and advanced large-scale video generative models. arXiv preprint arXiv:2503.20314  (2025)

\bibitem{wang2025vggt}
Wang, J., Chen, M., Karaev, N., Vedaldi, A., Rupprecht, C., Novotny, D.: Vggt: Visual geometry grounded transformer. In: 2025 IEEE/CVF Conference on Computer Vision and Pattern Recognition (CVPR). pp. 5294--5306. IEEE (2025)

\bibitem{wang2026repwam}
Wang, J., Zhang, Q., Yang, S., Luo, Y., Shen, Y., Wu, Z., Jiang, Y.G., Xu, Y.: Repwam: World action modeling with representation visual-action tokenizers. arXiv preprint arXiv:2606.13674  (2026)

\bibitem{wang2026robointer1}
Wang, Z., Li, H., Wang, W., Cai, J., Zeng, J., Chen, Y., Pang, J., Liu, S.: Robointer1. 5: A holistic intermediate representation suite for embodied world modeling and robotic manipulation. arXiv preprint arXiv:2607.18709  (2026)

\bibitem{wei2022chain}
Wei, J., Wang, X., Schuurmans, D., Bosma, M., Xia, F., Chi, E., Le, Q.V., Zhou, D., et~al.: Chain-of-thought prompting elicits reasoning in large language models. Advances in neural information processing systems  \textbf{35},  24824--24837 (2022)

\bibitem{xu2026action}
Xu, C., Li, H., Cheng, S., Fan, H., Feng, Z., Liu, S.: Action-geometry prediction with 3d geometric prior for bimanual manipulation. In: Proceedings of the IEEE/CVF Conference on Computer Vision and Pattern Recognition. pp. 35036--35046 (2026)

\bibitem{xu2025diffusion}
Xu, H., Ding, J., Xu, J., Wang, R., Chen, J., Mai, J., Fu, Y., Ghanem, B., Xu, F., Elhoseiny, M.: Diffusion-based imaginative coordination for bimanual manipulation. In: 2025 IEEE/CVF International Conference on Computer Vision (ICCV). pp. 11469--11479. IEEE (2025)

\bibitem{xu2026futurevla}
Xu, X., Li, H., Ye, J., Chen, Y., Zeng, J., Chen, X., Xu, L., Lin, D., Li, W., Pang, J.: Futurevla: Joint visuomotor prediction for vision-language-action model. arXiv preprint arXiv:2603.10712  (2026)

\bibitem{yan2023ucsd}
Yan, G., Wu, K., Wang, X.: ucsd kitchens dataset (2023)

\bibitem{yan2026robust}
Yan, H., Li, J., He, J., Zhong, Z., Yu, M., Song, W., Zhu, J., Zheng, Y., Du, Y., You, J., et~al.: Robust-wam: Bridging generative pretraining and semantic foresight in world-action models. arXiv preprint arXiv:2608.05903  (2026)

\bibitem{yang2025qwen3}
Yang, A., Li, A., Yang, B., Zhang, B., Hui, B., Zheng, B., Yu, B., Gao, C., Huang, C., Lv, C., et~al.: Qwen3 technical report. arXiv preprint arXiv:2505.09388  (2025)

\bibitem{magma}
Yang, J., Tan, R., Wu, Q., Zheng, R., Peng, B., Liang, Y., Gu, Y., Cai, M., Ye, S., Jang, J., et~al.: Magma: A foundation model for multimodal ai agents. arXiv preprint arXiv:2502.13130  (2025)

\bibitem{yang20264d}
Yang, L., Song, W., Wang, X., Sheng, P., Fang, Z., Zhou, Z., He, J., Yan, H., Chen, J., Sun, N., et~al.: 4d-wam: Infusing spatiotemporal awareness into world action models through trajectory fields. arXiv preprint arXiv:2608.08023  (2026)

\bibitem{instructvla}
Yang, S., Li, H., Chen, Y., Wang, B., Tian, Y., Wang, T., Wang, H., Zhao, F., Liao, Y., Pang, J.: Instructvla: Vision-language-action instruction tuning from understanding to manipulation. arXiv preprint arXiv:2507.17520  (2025)

\bibitem{yang2026robo3r}
Yang, S., Xu, L., Li, H., Mu, J., Zeng, J., Lin, D., Pang, J.: Robo3r: Enhancing robotic manipulation with accurate feed-forward 3d reconstruction. arXiv preprint arXiv:2602.10101  (2026)

\bibitem{yang2025novel}
Yang, S., Yu, W., Zeng, J., Lv, J., Ren, K., Lu, C., Lin, D., Pang, J.: Novel demonstration generation with gaussian splatting enables robust one-shot manipulation. arXiv preprint arXiv:2504.13175  (2025)

\bibitem{yang2025gripper}
Yang, Y., Cai, Z., Tian, Y., Zeng, J., Pang, J.: Gripper keypose and object pointflow as interfaces for bimanual robotic manipulation. arXiv preprint arXiv:2504.17784  (2025)

\bibitem{ye2026world}
Ye, S., Ge, Y., Zheng, K., Gao, S., Yu, S., Kurian, G., Indupuru, S., Tan, Y.L., Zhu, C., Xiang, J., et~al.: World action models are zero-shot policies. arXiv preprint arXiv:2602.15922  (2026)

\bibitem{ye2025lapa}
Ye, S., Jang, J., Jeon, B., Joo, S., Yang, J., Peng, B., Mandlekar, A., Tan, R., Chao, Y.W., Lin, B.Y., Liden, L., Lee, K., Gao, J., Zettlemoyer, L., Fox, D., Seo, M.: Latent action pretraining from videos. In: The Thirteenth International Conference on Learning Representations (ICLR) (2025)

\bibitem{yuan2026fastwam}
Yuan, T., Dong, Z., Liu, Y., Zhao, H.: Fast-wam: Do world action models need test-time future imagination? arXiv preprint arXiv:2603.16666  (2026), \url{https://arxiv.org/abs/2603.16666}

\bibitem{yuan2025depthvla}
Yuan, T., Liu, Y., Lu, C., Chen, Z., Jiang, T., Zhao, H.: Depthvla: Enhancing vision-language-action models with depth-aware spatial reasoning. arXiv preprint arXiv:2510.13375  (2025)

\bibitem{zhang2025dreamvla}
Zhang, W., Liu, H., Qi, Z., Wang, Y., Yu, X., Zhang, J., Dong, R., He, J., Lu, F., Wang, H., et~al.: Dreamvla: a vision-language-action model dreamed with comprehensive world knowledge. arXiv preprint arXiv:2507.04447  (2025)

\bibitem{zhang2025spatial}
Zhang, Z., Li, H., Dai, Y., Zhu, Z., Zhou, L., Liu, C., Wang, D., Tay, F.E., Chen, S., Liu, Z., et~al.: From spatial to actions: Grounding vision-language-action model in spatial foundation priors. arXiv preprint arXiv:2510.17439  (2025)

\bibitem{zhen20243d}
Zhen, H., Qiu, X., Chen, P., Yang, J., Yan, X., Du, Y., Hong, Y., Gan, C.: 3d-vla: A 3d vision-language-action generative world model. arXiv preprint arXiv:2403.09631  (2024)

\bibitem{zheng2025x}
Zheng, J., Li, J., Wang, Z., Liu, D., Kang, X., Feng, Y., Zheng, Y., Zou, J., Chen, Y., Zeng, J., et~al.: X-vla: Soft-prompted transformer as scalable cross-embodiment vision-language-action model. arXiv preprint arXiv:2510.10274  (2025)

\bibitem{zheng2024tracevla}
Zheng, R., Liang, Y., Huang, S., Gao, J., Daum{\'e}~III, H., Kolobov, A., Huang, F., Yang, J.: Tracevla: Visual trace prompting enhances spatial-temporal awareness for generalist robotic policies. arXiv preprint arXiv:2412.10345  (2024)

\bibitem{zheng2025flare}
Zheng, R., Wang, J., Reed, S., Bjorck, J., Fang, Y., Hu, F., Jang, J., Kundalia, K., Lin, Z., Magne, L., et~al.: Flare: Robot learning with implicit world modeling. arXiv preprint arXiv:2505.15659  (2025)

\bibitem{zhu2023fanuc}
Zhu, X., Tian, R., Xu, C., Huo, M., Zhan, W., Tomizuka, M., Ding, M.: Fanuc manipulation: A dataset for learning-based manipulation with fanuc mate 200id robot (2023)

\bibitem{zhu2023viola}
Zhu, Y., Joshi, A., Stone, P., Zhu, Y.: Viola: Imitation learning for vision-based manipulation with object proposal priors. In: Conference on Robot Learning. pp. 1199--1210. PMLR (2023)

\bibitem{zhu2022bottom}
Zhu, Y., Stone, P., Zhu, Y.: Bottom-up skill discovery from unsegmented demonstrations for long-horizon robot manipulation. IEEE Robotics and Automation Letters  \textbf{7}(2),  4126--4133 (2022)

\end{thebibliography}
}



\end{document}